%% file: main.tex
\documentclass[10pt,twocolumn,letterpaper]{article}

\usepackage[pagenumbers]{cvpr} 

\input{preamble}

\definecolor{cvprblue}{rgb}{0.21,0.49,0.74}
\usepackage[pagebackref,breaklinks,colorlinks,citecolor=cvprblue]{hyperref}

\def\paperID{225} \def\confName{3DV\xspace}
\def\confYear{2026\xspace}

\title{DINO in the Room: Leveraging 2D Foundation Models for 3D Segmentation}

\author{Karim Knaebel$^{1,\star}$
\and
Kadir Yilmaz$^{1,\star}$
\and
Daan de Geus$^{1,2}$
\and
Alexander Hermans$^{1}$
\and
David Adrian$^{3}$
\and
Timm Linder$^{3}$
\and
Bastian Leibe$^{1}$
\and
{\normalsize $^1$RWTH Aachen University} \quad
{\normalsize $^2$Eindhoven University of Technology} \quad
{\normalsize $^3$Bosch Center for AI}
\and
{\normalsize \httpsurl{vision.rwth-aachen.de/ditr}}
}

\begin{document}
\maketitle

\begingroup
\def\thefootnote{$\star$}
\footnotetext{Equal contribution. The order is determined by a last-minute coin flip.}
\endgroup

\input{sec/0_abstract}
\input{sec/1_introduction}

\input{sec/2_related}

\input{sec/3_method}
\input{sec/4_experiments}

\input{sec/5_conclusion}

{
\small
\PAR{Acknowledgements.}
Karim Knaebel, Kadir Yilmaz and Alexander Hermans are funded by the project \enquote{Context Understanding for Autonomous Systems} by Robert Bosch GmbH.
Computations were performed with computing resources granted by RWTH Aachen under projects \texttt{rwth1604} and \texttt{rwth1730}.

}

{
    \small
    \bibliographystyle{ieeenat_fullname}
    \bibliography{main}
}

\clearpage
\appendix

\input{sec/X_suppl}

\input{sec/X_suppl_floats}

\end{document}

%% file: sec/0_abstract.tex
\begin{abstract}
Vision foundation models (VFMs) trained on large-scale image datasets provide high-quality features that have significantly advanced 2D visual recognition.
However, their potential in 3D scene segmentation remains largely untapped, despite the common availability of 2D images alongside 3D point cloud datasets.
While significant research has been dedicated to 2D--3D fusion, recent state-of-the-art 3D methods predominantly focus on 3D data, leaving the integration of VFMs into 3D models underexplored.
In this work, we challenge this trend by introducing \ours{}, a generally applicable approach that extracts 2D foundation model features, projects them to 3D, and finally injects them into a 3D point cloud segmentation model.
\ours{} achieves state-of-the-art results on both indoor and outdoor 3D semantic segmentation benchmarks.
To enable the use of VFMs even when images are unavailable during inference, we additionally propose to pretrain 3D models by distilling 2D foundation models.
By initializing the 3D backbone with knowledge distilled from 2D VFMs, we create a strong basis for downstream 3D segmentation tasks, ultimately boosting performance across various datasets.
\end{abstract}

%% file: sec/1_introduction.tex
\section{Introduction}
\label{sec:intro}

The idiom \enquote{elephant in the room} refers to a situation where something important is being ignored, while it should be discussed.
Currently, 2D foundation models, such as DINOv2~\cite{oquab2023dinov2}, remain largely ignored for 3D segmentation, despite their ability to provide strong semantic priors.
Therefore, we believe that we need to talk about the \sout{elephant} DINO in the room.

The current paradigm for state-of-the-art 3D segmentation, both in literature and on benchmarks, is to use models with specialized 3D backbones that are trained from scratch~\cite{wu2022ptv2,wu2024ptv3,choy20194d}.
In contrast, for the highly related task of image segmentation, the current paradigm is to use 2D backbones initialized with pretrained weights of strong vision foundation models (VFMs)~\cite{caron2021dino,oquab2023dinov2,fang2024eva02,radford2021clip,zhai2023siglip,kirillov2023sam,Rombach2022stablediffusion,kerssies2024benchmarking,kerssies2025eomt}.
These VFMs are predominantly trained in a self-supervised manner on large-scale image datasets, enabling strong generalization capabilities.
However, current 3D datasets~\cite{dai2017scannet,rozenberszki2022scannet200,armeni2016s3dis,zheng2020structured3d,behley2019semantickitti,caesar2020nuscenes,sun2020waymo} are orders of magnitude smaller than their 2D counterparts~\cite{schuhmann2021laion,schuhmann2022laion}.
As a result, while progress has been made~\cite{wu2025sonata}, generalist 3D foundation models have yet to emerge.
Interestingly, though, we observe that 3D point cloud data is often accompanied by corresponding 2D images~\cite{dai2017scannet,rozenberszki2022scannet200,caesar2020nuscenes,sun2020waymo}.
This raises the question: how can we leverage the power of 2D foundation models for 3D segmentation?

\input{fig/0_teaser}

For most 3D point clouds, corresponding images are typically available due to the process with which this data is captured.
In indoor scenarios, a scene is captured with RGB or RGB-D cameras as a video sequence and a colored point cloud is formed via 3D reconstruction~\cite{dai2017scannet,rozenberszki2022scannet200}.
As a result, the data inherently includes 2D images, and correspondences between images and point clouds are obtained as part of data preprocessing.
In outdoor scenarios, 3D street scenes are typically captured with LiDAR scanners mounted on vehicles that are often also equipped with cameras, providing corresponding 2D images~\cite{caesar2020nuscenes,sun2020waymo,behley2019semantickitti}.
Despite the availability of these images, current state-of-the-art methods for both indoor and outdoor 3D semantic segmentation typically only use 3D data~\cite{wu2024ptv3, yilmaz24mask4former,wu2024ppt,kolodiazhnyi2024oneformer3d}.
While 2D--3D fusion has been investigated with moderate success~\cite{athar20234dformer, jain2024odin, zhang2023lcps, robert2022DeepViewAgg,tan2024epmf}, existing works have thus far not capitalized on the existence of semantically rich VFMs.

In this paper, we challenge the current paradigm and claim that VFMs can, as they did in 2D, enrich 3D models with generalist features that are not easily learned from the 3D data.
To verify this, we take a state-of-the-art 3D segmentation model~\cite{wu2024ptv3} and augment it with 2D VFM features.
Concretely, we take a frozen DINOv2~\cite{oquab2023dinov2} and extract 2D image features that correspond to the points in the associated 3D point cloud.
Next, for the 3D points that have been matched to 2D features, we inject these 2D features into the 3D model at different decoder stages.
Then, we train this model for the regular segmentation objective, resulting in a 2D-to-3D injection approach (see \cref{fig:teaser} (a)).
With this injection approach, which only uses unlabeled images, the semantic segmentation performance improves significantly, achieving new state-of-the-art results.
Especially for indoor datasets with many semantic classes and for outdoor datasets that provide $360^{\circ}$ camera coverage, this setup outperforms the 3D-only baseline on public leaderboards by large margins, \ie, \plusmiou{7.1} on ScanNet200~\cite{rozenberszki2022scannet200} and \plusmiou{2.4} on nuScenes~\cite{caesar2020nuscenes}.
Furthermore, it achieves the top score on the recent ScanNet++ benchmark~\cite{yeshwanthliu2023scannetpp}, outperforming the next best approach by \plusmiou{3.0}.
We call this approach \textit{DINO In The Room} (\ours{}).

Moreover, we show that even if images are not available during inference, it is still possible to utilize features from the same frozen DINOv2 model to enhance segmentation performance by pretraining 3D models through \textit{distillation}~\cite{hinton2015distilling}.
Specifically, we take 3D point cloud datasets that have corresponding images and teach a 3D student model to output features aligned with those extracted from a DINOv2 teacher model, using a distillation objective (see \cref{fig:teaser} (b)).
Subsequently, the pretrained, distilled student model can simply be fine-tuned for 3D segmentation in a regular fashion, without requiring corresponding images, and thus without causing any overhead during inference.
This distillation pretraining enables the 3D model to capture the semantic richness of 2D foundation models without requiring any labeled data, effectively using them as distillation targets instead of semantic labels.
Furthermore, it allows pretraining across multiple datasets without adjusting for dataset-specific semantic label sets, as the target feature space remains consistent.
With this setup, we observe consistent improvements on all datasets compared to random initialization.
On SemanticKITTI~\cite{behley2019semantickitti}, where there is only a single image per 3D LiDAR scan, and on S3DIS~\cite{armeni2016s3dis}, where camera views are sparser, this distillation setup even outperforms \ours{}.
We call this approach \textit{Distill DITR} (\oursdist{}).

Altogether, these results highlight the significant yet underexplored potential of VFMs for 3D segmentation tasks.
Therefore, we recommend that whenever corresponding images are available, they should be used to complement 3D segmentation models, preferably using injection and otherwise using distillation.
Additionally, we note that while we focus on DINOv2 in this paper, as it is one of the strongest existing VFMs, we show that the general setup can also leverage even stronger models, like the very recently introduced DINOv3~\cite{simeoni2025dinov3}.

In summary, our contributions are as follows:
\begin{itemize}
    \item We show that injection of DINOv2 features from 2D images into a 3D model can significantly improve the segmentation performance, achieving new state-of-the-art results on multiple indoor and outdoor datasets.
    \item We demonstrate that DINOv2 can also act as a teacher to pretrain 3D models through distillation, improving performance without requiring 2D data during inference.
\end{itemize}

%% file: fig/0_teaser.tex
\begin{figure}[t]
	\centering
\includegraphics[width=\columnwidth]{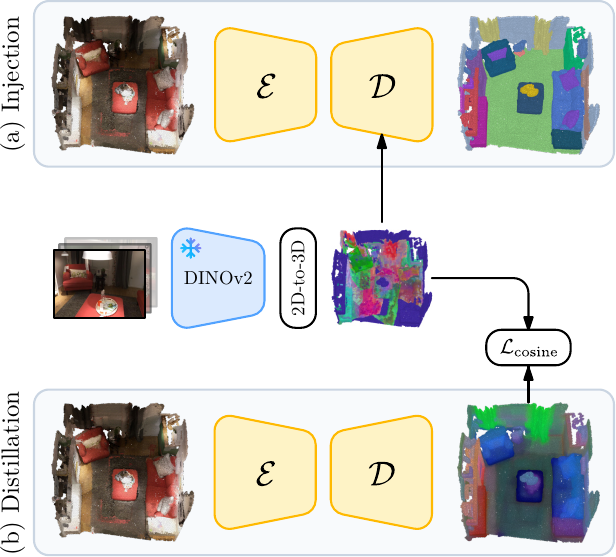}
	\caption{
            \textbf{DINO in the room (\ours{})}.
            We present an approach to (a) inject or (b) distill DINOv2 features into 3D semantic segmentation models that yields state-of-the-art results across indoor and outdoor 3D benchmarks.
}
	\label{fig:teaser}
\end{figure}

%% file: sec/2_related.tex
\section{Related Work}
\label{sec:related}

\PARbegin{3D Semantic Segmentation.}
Despite architectural differences, most 3D backbones for dense tasks such as segmentation follow a U-Net-like hierarchical encoder--decoder design with skip connections.
Early methods apply point-wise fully-connected layers~\cite{qi2017pointnet,qi2017pointnet++}, or continuous 3D convolutions~\cite{wu2019pointconv,thomas2019kpconv} to capture local geometric patterns.
Later, MinkUNet~\cite{choy20194d} significantly improves efficiency and accuracy by voxelizing point clouds and utilizing sparse 3D convolutions~\cite{graham2018submanifold}.
Following the success of Vision Transformers~\cite{dosovitskiy2021vit}, attention mechanisms have been increasingly adopted for 3D segmentation, although their quadratic computational complexity poses challenges for large-scale scenes.
To address this, several methods~\cite{lai2022stratified,Wang2023OctFormer,yang2023swin3d,zhao2021ptv1,wu2022ptv2} restrict attention to local neighborhoods, where points are subsampled and aggregated hierarchically across multiple layers.
Following this, Point Transformer V3 (PTv3)~\cite{wu2024ptv3} employs space-filling curves to map 3D point clouds into 1D sequences, effectively preserving the local neighborhood structure while significantly simplifying and accelerating local attention computations.
Most recently, Sonata~\cite{wu2025sonata} scales up PTv3, tripling the number of parameters, and adopts DINOv2-style self-supervised learning on point clouds, enabling the use of a substantially larger 3D training corpus~\cite{avetisyan2403scenescript,zheng2020structured3d,ramakrishnan2021hm3d,baruch2021arkitscenes,armeni2016s3dis,dai2017scannet,yeshwanthliu2023scannetpp}, albeit still orders of magnitude smaller than its 2D counterparts.
In contrast, our approach enhances PTv3 by directly incorporating features from powerful 2D VFMs such as DINOv2, thereby leveraging the vast amount of 2D training data on which these models are pretrained.

\PAR{2D--3D Fusion.}
Significant research has been dedicated to fusing 2D and 3D information for 3D segmentation.
Existing fusion methods are often designed with a domain-specific focus, addressing either indoor or outdoor scenes.

Indoor fusion methods exploit multi-view information from RGB-D videos, as points are visible from multiple angles.
Kundu \etal \cite{kundu2020VMVfusion} render 2D images and their corresponding ground-truth labels from 3D mesh reconstructions to train a 2D segmentation model.
During inference, resulting multi-view 2D predictions are aggregated in 3D space.
Other approaches leverage 2D segmentation annotations provided by the 3D datasets~\cite{dai2017scannet}, either by pretraining 2D networks~\cite{jaritz2019mvpnet} or by jointly training 2D and 3D networks with interactions between them~\cite{hu2021bpnet}.
Recent methods use dedicated architectures for 2D--3D fusion, introducing either an aggregation module that selectively fuses 2D features into a 3D backbone~\cite{robert2022DeepViewAgg}, or a unified 2D--3D model with alternating 2D--3D layers enabling multi-stage fusion of multi-modal information~\cite{jain2024odin}.

Outdoor fusion methods combine appearance-based 2D features with geometric LiDAR features for segmentation.
Some methods \cite{zhuang2021pmf,tan2024epmf} project LiDAR points to the image plane and jointly train a camera and a LiDAR network, while simultaneously aligning their features.
Other approaches introduce more sophisticated fusion mechanisms, such as selecting semantically relevant 2D regions for each point via local attention modules~\cite{zhang2023lcps}, or establishing 2D--3D correspondences through joint spatial and semantic reasoning~\cite{li2023MSeg3D}.
Finally, 4D-Former \cite{athar20234dformer} explores 2D--3D fusion for LiDAR segmentation and tracking, integrating 2D features into a 3D backbone at multiple stages.

While these works effectively use 2D information for 3D segmentation and achieve modest improvements, we observe that the community has not leveraged 2D VFMs for 3D segmentation, leaving a powerful information source untapped.
In this work, to the best of our knowledge, we are the first to explore how the power of VFMs can be leveraged for \textit{state-of-the-art} 3D segmentation, and demonstrate how 2D-to-3D injection with DITR yields new state-of-the-art results.
Moreover, \ours{} is the first fusion approach shown to be effective for both indoor and outdoor scenes, highlighting its general applicability.

\PAR{2D--3D Distillation.}
Recent methods explore various strategies to distill 2D representations into 3D backbones for enhanced representation learning in LiDAR point clouds.
Yan \etal \cite{yan20222dpass} propose a joint training framework, where a 3D model is trained for both 3D segmentation and alignment with a 2D model.
Several approaches apply contrastive learning either on super-points and super-pixels generated by pretrained 2D backbones~\cite{sautier2022slidr, liu2023seal}, or on prototypes created in both the 2D and 3D domains~\cite{chen2024pretraining3d}.
Others~\cite{mahmoud2023selfsupervised, mahmoud2024imagetolidar} propose new contrastive loss functions, and conduct experiments with linear probing or limited-data setups.
Recently, ScaLR~\cite{puy2024threepillars} studies the effects of the size of the backbones and datasets in an image-to-LiDAR distillation setting.
These methods focus on---and succeed in---image-to-LiDAR distillation for zero-shot, limited-data, or linear probing 3D segmentation settings.
However, they often yield little or no improvement when used as pretrained models that are subsequently fine-tuned on 100\,\% of the data, and fall short of state-of-the-art methods in that setting.
In contrast, we demonstrate that our \oursdist{} distillation \emph{is} an effective pretraining step that significantly improves the fine-tuning performance of a state-of-the-art model (PTv3). 
Moreover, unlike previous methods, it also works especially well for dense indoor scenes.

Orthogonal to other distillation methods, there is also a line of work that distills specific capabilities from VFMs.
Peng \etal~\cite{Peng2023OpenScene} distill language-aligned CLIP~\cite{radford2021clip} features from multiple views for 3D open-vocabulary segmentation.
Other approaches~\cite{osep2024bettercallsal,Huang2023Segment3D} use projected SAM~\cite{kirillov2023sam} segmentation masks as pseudo ground-truth labels to train class-agnostic 3D segmentation models.
Although these approaches also distill from 2D foundation models, they focus on obtaining task-specific capabilities in a zero-shot setting to avoid the reliance on 3D segmentation annotations.
In contrast, we use distillation to obtain a pretrained 3D model that extracts semantically rich point features, which can then be fine-tuned for 3D segmentation and, in principle, other 3D tasks as well.

%% file: sec/3_method.tex
\section{Method}
\label{sec:method}

\input{fig/1_method}

We propose two variants of \ours{} for 3D semantic segmentation: an \emph{injection} approach that uses 2D features from DINOv2~\cite{oquab2023dinov2} during training and inference, and \oursdist{}, a \emph{distillation} approach that aligns 3D features with DINOv2 features during a pretraining phase, which can be followed by image-free fine-tuning on a specific dataset.
In the following, we describe both variants in detail.

\subsection{Injection}\label{sec:method:injection}
In many modern 3D datasets, images from calibrated cameras are provided alongside the point clouds~\cite{dai2017scannet,armeni2016s3dis,caesar2020nuscenes,behley2019semantickitti,sun2020waymo}.
We leverage these images to inject semantically rich 2D features into the 3D backbone.
An overview of this process is shown in \cref{fig:method}: we first map 3D points to their corresponding pixels to extract associated DINOv2 features (2D-to-3D Mapping), and then inject these features into the skip connections of the 3D backbone's decoder (3D Feature Fusion).

\PAR{2D-to-3D Mapping.}
Let \(\mathcal{P} = \{\vb{p}_i \in \mathbb{R}^3\}_{i=1}^N\) be a 3D point cloud of \(N\) points, and assume a collection of \(K\) calibrated cameras, \(\{(\vb{I}_k, \vb{K}_k, \vb{T}_k)\}^{K}_{k=1}\).
Each camera has intrinsics \(\vb{K}_k\) and world-to-camera extrinsics \(\vb{T}_k\), with \(\vb{I}_k\) denoting its captured image of resolution \(H \times W\).
We feed each 2D image \(\vb{I}_k\) into a frozen DINOv2 ViT~\cite{dosovitskiy2021vit}, obtaining patch-level embeddings \(\vb{F}_k \in \mathbb{R}^{\frac{H}{P} \times \frac{W}{P} \times D_{\text{2D}}}\), where \(P\) is the patch size and \(D_{\text{2D}}\) is the feature dimension.
For each point \(\vb{p}_i\), we transform it into the \(k\)-th camera space as \(\vb{q}_i^k = \vb{T}_k \, (\vb{p}_i, 1)\) and multiply by the intrinsics \(\vb{K}_k\) to obtain homogeneous pixel coordinates \(\left(x_i^k,\, y_i^k,\, z_i^k\right) = \vb{K}_k \vb{q}_i^k\).
The pixel coordinates are then given by \(
\left(u_i^k,\, v_i^k\right) = \left(x_i^k / z_i^k,\, y_i^k / z_i^k\right)\).
We consider point \(\vb{p}_i\) visible in the \(k\)-th image if \(u_i^k \in [0,W)\), \(v_i^k \in [0,H)\) and the depth \(z_i^k\) is positive (view-frustum culling).
If \(\vb{p}_i\) is visible in the \(k\)-th image, we determine the corresponding patch index as
\begin{equation*}
    \left(\hat{u}_i^k,\, \hat{v}_i^k\right) = \left(\left\lfloor \tfrac{u_i^k}{P} \right\rfloor, \left\lfloor \tfrac{v_i^k}{P} \right\rfloor\right),
\end{equation*}
and assign the corresponding feature from the 2D feature map \(\vb{F}_k\) to \(\vb{p}_i\).
If \(\vb{p}_i\) is visible in multiple images, we randomly select one of them to provide the feature.
Otherwise, if \(\vb{p}_i\) is not visible in any image, we assign an all-zero feature vector.
Empirically, we find that directly assigning the feature from the frozen 2D feature map \(\vb{F}_k\) at patch index \((\hat{u}_i^k, \hat{v}_i^k)\) yields better results than bilinearly interpolating features from \(\vb{F}_k\) at the pixel coordinates \((u_i^k, v_i^k)\).
Also, aggregating from multiple 2D feature maps resulted in inferior performance compared to random selection.

\PAR{3D Feature Fusion.}
Because point clouds are unstructured, common 3D backbones~\cite{wu2024ptv3,choy20194d,zhu2021cylindrical} voxelize the input, keeping only one point per voxel, thereby reducing \(\mathcal{P}\) to \(\mathcal{P}'\) with \(M \le N\) points.
These backbones typically follow a U-Net-like encoder--decoder architecture, processing \(\mathcal{P}'\) at multiple spatial resolution levels \(l \in \{1, 2, \dots, L\}\).
At each encoder level, a pooling layer downsamples the point features to a coarser spatial resolution.
Formally, let \(\vb{X}_l^\mathcal{E} \in \mathbb{R}^{M_l \times D_l}\) denote the output of the \(l\)-th level encoder block \(\mathcal{E}_l\), where \(l=1\) is the finest (original) resolution with \(M_1 = M\).
After the encoder and the bottleneck \(\mathcal{B}\), the decoder upsamples and refines these features back toward the original resolution, yielding \(\vb{X}_l^\mathcal{D}\) as the output of each decoder block \(\mathcal{D}_l\).
To simplify notation, we define \(\vb{X}_{L + 1}^\mathcal{D}\) to be the output of the bottleneck \(\mathcal{B}\).

To incorporate DINOv2 features, we first use the previously described assignment to gather point-wise features \(\vb{X}^\text{2D} \in \mathbb{R}^{M \times D_\text{2D}}\) for the points in \(\mathcal{P}'\) from the 2D feature maps \(\vb{F}_k\).
Then, to mirror the spatial structure of the decoder features \(\vb{X}_l^\mathcal{D}\) across different levels, we repeatedly apply max pooling to \(\vb{X}^\text{2D}\), obtaining features \(\vb{X}_l^\text{2D}\) for each level \(l\).
In the decoder, features from the encoder (\(\vb{X}_l^\mathcal{E}\)) and unpooled features from the previous decoder level (\(\uparrow\negthickspace\vb{X}_{l+1}^\mathcal{D}\)) are combined via skip connections.
In \ours{}, we additionally inject \(\vb{X}_l^\text{2D}\) into these skip connections. Specifically, the fused input to the \(l\)-th decoder block \(\mathcal{D}_l\) becomes
\begin{equation*}
    \underbrace{f_l^{\mathcal{D}}\bigl(\uparrow\negthickspace\vb{X}_{l+1}^\mathcal{D}\bigr)}_{\text{\makebox[0pt]{prev.\ dec.\ block}}}
    \;+\;
    \underbrace{f_l^{\mathcal{E}}\bigl(\vb{X}_l^\mathcal{E}\bigr)}_{\text{\makebox[0pt]{skip connection}}}
    \;+\;
    \underbrace{f_l^{\text{2D}}\bigl(\vb{X}_l^\text{2D}\bigr)}_{\text{\makebox[0pt]{DINOv2}}},
\end{equation*}
where \(+\) denotes element-wise addition, and each \(f_l^{\cdot}\) is a linear projection into the decoder's feature dimension followed by batch normalization~\cite{ioffe2015batchnorm} and GELU~\cite{hendrycks2016gelu}.
Finally, the output \(\vb{X}^\mathcal{D}_1\) of the last decoder block yields per-point features for \(\mathcal{P}'\), which are passed through a linear segmentation head to produce class logits. With our injection approach, we ensure that rich 2D VFM features are available at all 3D decoder feature resolutions, enabling access to 2D information where needed and at different levels of granularity.

\subsection{Distillation}\label{sec:method:distillation}
While injecting DINOv2 features directly into the 3D backbone can significantly boost segmentation performance, relying on calibrated images at inference time may be restrictive in certain real-world scenarios.
To address situations where only 3D data can be used at test time, we propose a distillation scheme (\oursdist{}) to transfer 2D knowledge into a pure 3D model as a pretraining step.

\PAR{Pretraining via 2D-to-3D Alignment.}
During pretraining, we first match 3D points to 2D patches and assign per-point DINOv2 features $\vb{X}^\text{2D}$ as described in \cref{sec:method:injection}.
We then feed only the point cloud into the 3D backbone, but instead of predicting segmentation logits, the network's final linear layer regresses the DINOv2 features.
For each point \(\vb{p}_i\), we denote the predicted feature by \(\vb{x}_i^\text{pred}\) and the corresponding DINOv2 target feature by \(\vb{x}_i^\text{2D}\).
We minimize the following cosine similarity loss~\cite{Peng2023OpenScene}, averaged over all visible points:
\begin{equation}
    \mathcal{L}_{\text{cosine}} = \frac{1}{|\mathcal{V}|}
    \sum_{i \in \mathcal{V}}
    \left[
        1 \;-\;
        \frac{\vb{x}_i^\text{pred} \cdot \vb{x}_i^\text{2D}}
        {\|\vb{x}_i^\text{pred}\|\;\|\vb{x}_i^\text{2D}\|}
        \right],
\end{equation}
where $\mathcal{V}$ is the set of indices of points visible in at least one camera.
This distillation pretraining encourages the 3D backbone to replicate the semantically rich representations of DINOv2, which capture fine-grained details without being constrained by the coarse granularity of traditional semantic segmentation annotations.
An overview of the distillation setup is shown in \cref{fig:teaser} (b).

\PAR{Image-Free Inference.}
After pretraining, we discard the linear regression head and replace it with the standard segmentation head.
The pretrained model can then be fine-tuned on any 3D segmentation dataset without requiring image features.
The final model thus makes predictions based on 3D data only, yet benefits from the semantic knowledge transferred from strong 2D models.

\PAR{Multi-Dataset Training.}
A key advantage of our 2D-to-3D distillation approach is that it does not require any annotated data.
It only requires aligned 2D and 3D inputs.
This enables us to combine multiple unannotated point cloud datasets under a single distillation objective, where each point simply regresses to its corresponding DINOv2 feature.
Following prior multi-dataset work~\cite{wu2024ppt}, we maintain separate batch-normalization layers per dataset, which empirically leads to more stable training.

%% file: fig/1_method.tex
\begin{figure*}[t]
    \centering
    \includegraphics[width=1\textwidth]{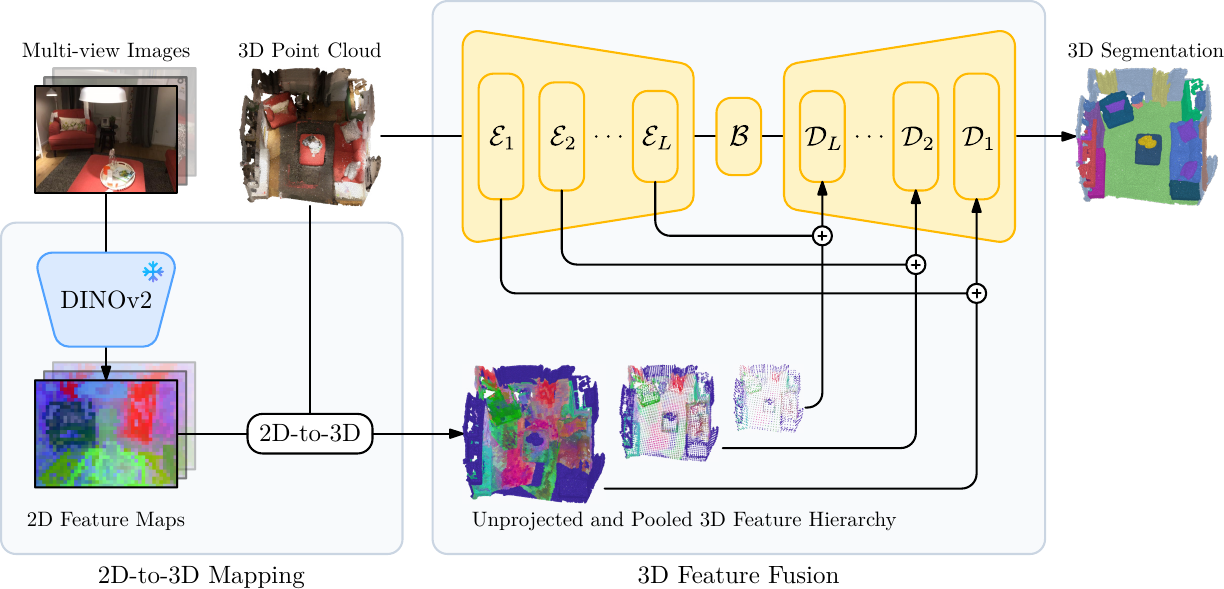}
\caption{
        \textbf{\ours{} architecture overview.}
        We extract 2D image features from a frozen DINOv2~\cite{oquab2023dinov2} model~\boxdino{} and unproject them (2D-to-3D) onto the 3D point cloud.
        The unprojected features are subsequently max-pooled to create a multi-scale feature hierarchy.
        The raw point cloud is fed through a 3D backbone~\boxbackbone{} and the unprojected image features are added to the skip connection between the encoder \(\mathcal{E}_l\) and decoder \(\mathcal{D}_l\) blocks on each level $l \in \{1, 2, \dots, L\}$.
        The model is then trained with the regular segmentation loss.
    }
    \label{fig:method}
\end{figure*}

%% file: sec/4_experiments.tex
\section{Experiments}
\label{sec:experiments}

\subsection{Datasets}
We evaluate our method on well-established 3D indoor and outdoor segmentation benchmarks.
For indoor segmentation, we select ScanNet~\cite{dai2017scannet}, ScanNet200~\cite{rozenberszki2022scannet200}, ScanNet++~v2 (SN++)~\cite{yeshwanthliu2023scannetpp}, and S3DIS~\cite{armeni2016s3dis}.
All four datasets consist of colored point clouds and corresponding RGB-D frames.
For outdoor segmentation, we select nuScenes~\cite{caesar2020nuscenes}, SemanticKITTI (Sem.KITTI)~\cite{behley2019semantickitti}, and Waymo~\cite{sun2020waymo}.

\subsection{Implementation Details}
We use PTv3~\cite{wu2024ptv3} as our 3D backbone, as it is the state-of-the-art model for 3D semantic segmentation across both indoor and outdoor datasets.
Unless stated otherwise, we retain the default architecture and hyperparameters of PTv3 to ensure fair comparison. Additional details are provided in the supplementary material.
For all datasets, we adhere to standard evaluation protocols and metrics.
Reproducing PTv3 results on the S3DIS\footnote{https://github.com/Pointcept/Pointcept/issues/154} and SemanticKITTI\footnote{https://github.com/Pointcept/Pointcept/issues/186} datasets has been notoriously hard for the community.
Therefore, we also report our reproduced results as a reference.

For outdoor scenarios, we observe large performance improvements with larger DINOv2 variants and thus we opt for the largest ViT-g model.
For indoor datasets, the ViT-L and ViT-g variants yield comparable results (as shown in \cref{tab:backbone}).
Consequently, we choose the ViT-L model to optimize resource efficiency, enabling the use of more images while remaining within GPU memory constraints.

During both training and inference on outdoor scenes, we use all available camera views: one for SemanticKITTI, five for Waymo, and six for nuScenes.
These camera views appear in a consistent orientation with respect to the LiDAR sensor and can be assumed always to be present.
For the indoor datasets, the number of RGB-D frames per scene can vary tremendously and always surpasses the number of frames we can effectively process with the available GPU memory.
Therefore, unless specified otherwise, we select 10 uniformly sampled, temporally equidistant frames from each RGB-D video during inference, while randomly sampling 10 views per scene during training to enhance data diversity.
Since DINOv2 is pretrained at a maximum resolution of \(518 \times 518\), we resize input images to maintain a similar number of patches while preserving aspect ratio~\cite{tschannen2025siglip2}.

\subsection{Main Results}

\input{tab/0_indoor_sem_seg}
\input{tab/1_outdoor_sem_seg}

\PARbegin{Injection.}
\cref{tab:performance_metrics_indoor,tab:performance_metrics_outdoor} show that \ours{} significantly outperforms the reproduced PTv3 baseline~\cite{wu2024ptv3} by injecting DINOv2 image features into the PTv3 backbone.
These improvements are consistent across both indoor and outdoor benchmarks.
We find that the performance improvement is particularly large on the hidden test set of ScanNet200, where we observe a gain of \plusmiou{7.1}.
This shows that image feature injection is especially beneficial for more complex segmentation tasks with many categories.
The only dataset where we observe only moderate gains is the Sem.KITTI dataset.
However, this is to be expected because this dataset provides images from only a single front-facing camera, covering only a fraction of the 3D points, limiting the number of points for which 2D injection can be applied.

Compared to existing state-of-the-art methods, we note that \ours{} not only outperforms all previous 2D--3D fusion methods, but also the very recent Sonata~\cite{wu2025sonata} on most datasets.
The improved performance compared to existing fusion methods is especially noteworthy because these methods only focus on either indoor or outdoor segmentation, while \ours{} is effective in both settings, and because some of them require 2D segmentation labels, which \ours{} does not.
The fact that \ours{} surpasses Sonata, on the other hand, is striking because Sonata uses a PTv3 backbone that is three times larger than \ours{}'s, and because it uses an expanded 3D training corpus.
Interestingly, we observe in \cref{tab:backbone} that \ours{} still outperforms Sonata even when using the small DINOv2-S backbone, with \ours{}'s model size---including a frozen DINOv2---amounting to only half of that of Sonata.
This reinforces our main message: given the relatively limited availability of 3D data, 2D VFMs are crucial for advancing 3D segmentation, and even the inexpensive small variants can be highly effective.

\input{tab/2_indoor_distillation}
\input{tab/3_outdoor_distillation}

\PAR{Distillation.}
For distillation pretraining, we explore two settings: (1) pretraining on individual datasets and fine-tuning on the same dataset to show the effectiveness of VFM features as distillation targets, and (2) joint pretraining on multiple datasets.
In the indoor multi-dataset distillation case, we jointly pretrain on ScanNet and Structured3D~\cite{zheng2020structured3d} and for the outdoor case we use nuScenes, SemanticKITTI, and Waymo.
We exclude S3DIS from pretraining due to its smaller size and instead use it to assess generalization on an unseen dataset.
The results are presented in \cref{tab:indoor_distillation} and \cref{tab:outdoor_distillation} as \oursdist{}.
We find that single-dataset distillation yields consistent improvements over the PTv3 baseline, demonstrating the effectiveness of DINOv2 distillation. The use of multi-dataset distillation further improves the results.
In indoor datasets, distillation pretraining significantly improves the segmentation performance, with gains of \plusmiou{2.4} on ScanNet, \plusmiou{2.3} on ScanNet200 and \plusmiou{2.9} on S3DIS over the state-of-the-art 3D baseline.
The improvement on S3DIS is particularly interesting, indicating that the distilled features can generalize to unseen datasets.
For outdoor datasets, \oursdist{} also produces consistent improvements over the baseline.
In particular, the distillation approach yields notable gains on SemanticKITTI, and even surpasses the \emph{injected} \ours{} model (see \cref{tab:performance_metrics_outdoor}).
This indicates that even when images are not available, or compute constraints do not allow for the computation of DINOv2 features during inference, \oursdist{} can utilize the strong 2D features that it sees during pretraining for enhanced segmentation during inference.

\input{fig/2_qualitative}
\cref{fig:distill_pca} visualizes feature predictions by \oursdist{} prior to fine-tuning.
When compared to the ground-truth segmentation labels, objects are clearly separable by color, even when visualizing only the first three principal components of the high-dimensional feature space.
This further indicates the value of 2D VFM features for 3D semantic segmentation.

\subsection{Ablation Study}

\input{tab/X_ablation_injection_point}

\PARbegin{2D--3D Injection Method.}
As described in \cref{sec:method:injection}, \ours{} injects 2D image features into all blocks of the 3D decoder. 
In \cref{tab:injection_place}, we compare this approach to several alternatives, including the 2D--3D injection methods that are employed by existing state-of-the-art 2D--3D fusion methods for indoor 3D segmentation that do not require 2D labels~\cite{jaritz2019mvpnet,robert2022DeepViewAgg,yang2023DMFnet}.
The results show that \textit{early} injection---\ie, before the 3D encoder---as employed by MVPNet~\cite{jaritz2019mvpnet} and DVA~\cite{robert2022DeepViewAgg}, is better than \textit{no} injection, but that it underperforms \ours{}'s \textit{intermediate} injection in all decoder blocks. 
The same applies to DMF-Net's approach~\cite{yang2023DMFnet} of having separate 3D encoder--decoders for feature extraction and 3D segmentation, and injecting features between these two encoder--decoders. 
Finally, we also evaluate \textit{late} injection methods, which either inject 2D features only in the last decoder block or after the last decoder block, but we find that these also perform worse than \ours{}'s default injection.
These results highlight the importance of informing the 3D model about 2D features at multiple stages of the network, as done by \ours{}'s 2D--3D injection approach.

In addition to the injection of 2D image features, we also experiment with injecting the \texttt{[cls]} tokens, generated by the 2D VFM, as additional tokens into the self-attention operations of the 3D decoder. 
This is inspired by the observation in \cref{tab:visible_invisible_split} that \ours{} even achieves significant improvements on invisible points, suggesting that it might leverage additional features as a kind of global context, which could be contained by the \texttt{[cls]} token as well.
However, we find that injecting the \texttt{[cls]} tokens only yields a small boost on indoor scenarios and does not impact outdoor performance. 
Therefore, we did not further pursue this direction.

\input{tab/X_ablation_vit_backbone}

\PAR{Image Backbone Ablation.}
To assess the impact of the image backbone used to extract 2D features, we experiment with various commonly used image backbones with frozen weights as shown in \cref{tab:backbone}.
When comparing different DINOv2-pretrained ViT models, there is a consistent trend that larger models perform better.
Comparing DINOv2 with other pretraining approach, we find that a standard ViT-L model pretrained on IN21k~\cite{deng2009imagenet,steiner2022vitaugreg} shows a modest performance improvement in indoor scenarios compared to the no-injection baseline, but reduces performance on the outdoor nuScenes dataset.
SigLIP~2~\cite{tschannen2025siglip2} and AIMv2~\cite{fini2024aimv2}, on the other hand, consistently outperform the baseline in both indoor and outdoor settings, but perform worse than DINOv2.
Finally, DINOv3~\cite{simeoni2025dinov3}, which was released only days ago, performs even better than DINOv2 with ViT-L, and achieves the absolute best ScanNet200 performance. 
These results show that strong pretraining, as used in foundation models, is key to achieving consistent gains across diverse environments.
They also suggest that our conclusions are not specific to DINOv2 but hold with very recent, more powerful VFMs like DINOv3 as well, and that these models are simply drop-in replacements in our framework.

\subsection{Additional Analyses}

\input{tab/4_indoor_efficient}
\PARbegin{ScanNet Data Efficient Benchmark.}
To demonstrate that \oursdist{} is also effective in label-scarce settings, we use the ScanNet \enquote{limited reconstructions} benchmark and show results for different percentages of available data during fine-tuning in \cref{tab:performance_limited_data_finetuning}.
For this setting, we use MinkUNet~\cite{choy20194d} as the student model to compare with previous unsupervised approaches in a consistent setting.
Similar to CSC~\cite{hou2021contrastive}, we use additional unlabeled images.
\oursdist{} outperforms all previous unsupervised pretraining approaches in all settings.
The gap is even larger when a small percentage of data is available during fine-tuning.
Overall, this experiment shows that our distillation approach can be applied to other commonly used 3D backbones and that it provides a strong supervision signal compared to raw 3D data.

\input{tab/distribution_images}
\PAR{Visibility vs.\ Performance.}
To better understand the source of the performance improvement, in \cref{tab:visible_invisible_split}, we compare the performance of \ours{} and the baseline on visible and invisible points separately.
\ours{} shows a notable improvement on visible points for both the ScanNet200 and SemanticKITTI datasets.
Moreover, it still outperforms the PTv3 baseline even on invisible points.
The large improvement on ScanNet200 suggests that the model captures global context from the 2D features, aiding the segmentation of invisible regions.

\PAR{Resource Usage.}
We assess the runtime impact of \ours{}, compared to the PTv3 baseline without additional injection of 2D features.
On ScanNet200, using a ViT-L DINOv2 backbone and 10 camera views, training time increases from 12 to 15 hours using two H100 GPUs.
The average per-scene inference latency increases from \num{41} to \SI{76}{\milli\second}, while the required GPU memory increases from \num{1.4} to \SI{5.6}{\gibi\byte} (using a single H100 GPU for inference).
Although this is a limitation of \ours{}, we note that these inference penalties do not apply to \oursdist{} as the architecture remains unchanged in that setting.

%% file: tab/0_indoor_sem_seg.tex
\begin{table}[t]
	\centering
	\small
	\renewcommand{\tabcolsep}{1.3pt}
	\begin{tabularx}{\linewidth}{lcYYcYYcYcY}
		\toprule
\multirow{2}[2]{*}{Method}         &  & \multicolumn{2}{c}{ScanNet}   &               & \multicolumn{2}{c}{ScanNet200} &                               & S3DIS         &  & SN++                                                              \\
		\cmidrule{3-4} \cmidrule{6-7} \cmidrule{9-9} \cmidrule{11-11}
		                                   &  & Val                           & Test          &                                & Val                           & Test          &  & Area5                                          &  & Test          \\\midrule
		ST~\cite{lai2022stratified}        &  & 74.3                          & 73.7          &                                & ---                           & ---           &  & 72.0                                           &  & ---           \\PTv1~\cite{zhao2021ptv1}           &  & 70.6                          & ---           &                                & 27.8                          & ---           &  & 70.4                                           &  & ---           \\PointNeXt~\cite{qian2022pointnext} &  & 71.5                          & 71.2          &                                & ---                           & ---           &  & 70.5                                           &  & ---           \\MinkUNet~\cite{choy20194d}         &  & 72.2                          & 73.6          &                                & 25.0                          & 25.3          &  & 65.4                                           &  & 45.6          \\OctFormer~\cite{Wang2023OctFormer} &  & 75.7                          & 76.6          &                                & 32.6                          & 32.6          &  & ---                                            &  & 46.0          \\Swin3D~\cite{yang2023swin3d}       &  & 76.4                          & ---           &                                & ---                           & ---           &  & 72.5                                           &  & ---           \\PTv2~\cite{wu2022ptv2}             &  & 75.4                          & 74.2          &                                & 30.2                          & ---           &  & 71.6                                           &  & 44.5          \\PTv3~\cite{wu2024ptv3}             &  & \textcolor{noreproduce}{77.5} & 77.9          &                                & \textcolor{noreproduce}{35.2} & 37.8          &  & \textcolor{noreproduce}{73.4}\rlap{$^\dagger$} &  & 48.8          \\\downrightarrow reproduced         &  & 76.8                          & ---           &                                & 35.4                          & ---           &  & 72.1                                           &  & ---           \\Sonata~\cite{wu2025sonata}         &  & 79.4                          & ---           &                                & 36.8                          & ---           &  & \textbf{76.0}                                           &  & 49.5\rlap{$^\ddagger$}           \\\midrule
		DVA~\cite{robert2022DeepViewAgg}
		                                   &  & 71.0                          & ---           &                                & ---                           & ---           &  & 67.2                                           &  & ---
		\\
		BPNet$^\ddagger$~\cite{hu2021bpnet}
		                                   &  & 73.9                          & 74.9          &                                & ---                           & ---           &  & ---                                            &  & ---
		\\
		DMF-Net~\cite{yang2023DMFnet}
		                                   &  & 75.6                          & 75.2          &                                & ---                           & ---           &  & ---                                            &  & ---
		\\
		VMVF$^*$~\cite{kundu2020VMVfusion}
		                                   &  & 76.4                          & 74.6          &                                & ---                           & ---           &  & ---                                            &  & ---
		\\
		ODIN$^*$~\cite{jain2024odin}
		                                   &  & 77.8                          & 74.4          &                                & 40.5                          & 36.8          &  & 68.6                                           &  & ---
		\\
		\midrule
		\ours{}                            &  & \textbf{80.5}                 & \textbf{79.7} &                                & \textbf{41.2}                 & \textbf{44.9} &  & 74.1                                  &  & \textbf{52.5} \\\bottomrule
	\end{tabularx}
	\caption{
		\textbf{Indoor semantic segmentation results (mIoU).}
		\ours{} significantly outperforms PTv3 on all datasets and obtains state-of-the-art results on three of them. We compare against both 3D-only (top) and 2D--3D fusion methods (bottom).
$^\dagger$ Uses a smaller point patch size and relative positional encoding.
        $^\ddagger$ Latest results from the challenge in combination with multi-dataset fine-tuning.
        $^*$ Requires 2D segmentation labels.
	}
	\label{tab:performance_metrics_indoor}
\end{table}

%% file: tab/1_outdoor_sem_seg.tex
\begin{table}[t]
	\centering
	\renewcommand{\tabcolsep}{1.5pt}
	\small
	\begin{tabularx}{\linewidth}{lcYYcYYcY}
		\toprule
\multirow{2}[2]{*}{Method}           &  & \multicolumn{2}{c}{nuScenes}  &               & \multicolumn{2}{c}{Sem.KITTI} &                               & Waymo                                            \\
		\cmidrule{3-4} \cmidrule{6-7} \cmidrule{9-9}
		                                     &  & Val                           & Test          &                               & Val                           & Test          &  & Val                           \\\midrule
		MinkUNet~\cite{choy20194d}           &  & 73.3                          & ---             &                               & 63.8                          & ---             &  & 65.9                          \\SPVNAS~\cite{tang2020spvnas}         &  & 77.4                          & ---             &                               & 64.7                          & 66.4          &  & ---                             \\Cylinder3D~\cite{zhu2021cylindrical} &  & 76.1                          & 77.2          &                               & 64.3                          & 67.8          &  & ---                             \\SphereFormer~\cite{lai2023spherical} &  & 78.4                          & 81.9          &                               & 67.8                          & \textbf{74.8} &  & 69.9                          \\PTv2~\cite{wu2022ptv2}               &  & 80.2                          & 82.6          &                               & 70.3                 & 72.6          &  & 70.6                          \\PTv3~\cite{wu2024ptv3}               &  & \textcolor{noreproduce}{80.4} & 82.7          &                               & \textcolor{noreproduce}{70.8} & 74.2          &  & \textcolor{noreproduce}{71.3} \\\downrightarrow reproduced           &  & 79.9                          & ---           &                               & 68.3                          & ---           &  & 71.5                          \\Sonata~\cite{wu2025sonata}           &  & 81.7                          & ---           &                               & \textbf{72.6}                          & ---           &  & 72.9                          \\\midrule
		4D-Former~\cite{athar20234dformer}

		                                     &  & 78.9                          & 80.4          &                               & 66.3                          & ---           &  & ---                           \\
		2DPASS~\cite{yan20222dpass}
		                                     &  & 79.4                          & 80.8          &                               & 69.3                 & 72.9          &  & ---                           \\
		MSeg3D~\cite{li2023MSeg3D}
		                                     &  & 80.0                          & 81.1          &                               & 66.7                          & ---           &  & 69.6                          \\
		LCPS~\cite{zhang2023lcps}
		                                     &  & 80.5                          & 78.9          &                               & 67.5                          & 62.8          &  & ---                           \\
		\midrule
		\ours{}                              &  & \textbf{84.2}                 & \textbf{85.1} &                               & 69.0                          & 74.4 &  & \textbf{73.3}                 \\
		\bottomrule
	\end{tabularx}
	\caption{
            \textbf{Outdoor semantic segmentation results (mIoU).}
            \ours{} obtains state-of-the-art results on datasets with full camera coverage: nuScenes and Waymo.
            We compare against both 3D-only (top) and 2D--3D fusion methods (bottom).
        }
	\label{tab:performance_metrics_outdoor}
\end{table}

%% file: tab/2_indoor_distillation.tex
\begin{table}[t]
	\centering
	\small
\begin{tabularx}{\linewidth}{lYYY}
		\toprule
		Method                      & ScanNet                       & ScanNet200                    & S3DIS                                          \\
		\midrule
		PTv3~\cite{wu2024ptv3}      & \textcolor{noreproduce}{77.5} & \textcolor{noreproduce}{35.2} & \textcolor{noreproduce}{73.4}\rlap{$^\dagger$} \\
		\downrightarrow reproduced  & 76.8                          & 35.4                          & 72.1                                           \\

        \midrule
		\oursdist{}                 & 78.6                          & 37.2                          & ---                                            \\
		\oursdist{} (multi-dataset) & \textbf{79.2}                 & \textbf{37.7}                 & \textbf{75.0}                                  \\
		\bottomrule
	\end{tabularx}
        \caption{
            \textbf{Indoor distillation results (mIoU).}
            We evaluate models pretrained and fine-tuned on the same datasets, as well as one pretrained on ScanNet and Structured3D jointly.
            All models outperform the randomly initialized PTv3 baseline.
            $^\dagger$ Uses a smaller point patch size and relative positional encoding.
        }
	\label{tab:indoor_distillation}
\end{table}

%% file: tab/3_outdoor_distillation.tex
\begin{table}[t]
	\centering
	\small
\begin{tabularx}{\linewidth}{lYYY}
		\toprule
		Method                      & nuScenes                      & Sem.KITTI                     & Waymo                         \\
		\midrule
		PTv3~\cite{wu2024ptv3}      & \textcolor{noreproduce}{80.4} & \textcolor{noreproduce}{70.8} & \textcolor{noreproduce}{71.3} \\
		\downrightarrow reproduced  & 79.9                          & 68.3                          & 71.5                          \\
\midrule
		\oursdist{}                 & \textbf{80.9}                 & ---                           & 71.6                          \\
		\oursdist{} (multi-dataset) & 80.7                          & \textbf{69.8}                 & \textbf{72.1}                 \\
		\bottomrule
	\end{tabularx}
	\caption{
            \textbf{Outdoor distillation results (mIoU).}
            We evaluate models pretrained and fine-tuned on the same datasets, as well as one pretrained on all three.
            In all cases, our models outperform the randomly initialized PTv3 baseline.
            Sem.KITTI is excluded from single-dataset pretraining due to its single-camera setup.
        }
	\label{tab:outdoor_distillation}
\end{table}

%% file: fig/2_qualitative.tex
\begin{figure}
    \renewcommand{\tabcolsep}{1pt}
    \begin{tabularx}{\linewidth}{cYY}
        \raisebox{20pt}{\rotatebox{90}{\scriptsize S3DIS}}&\includegraphics[width=\linewidth]{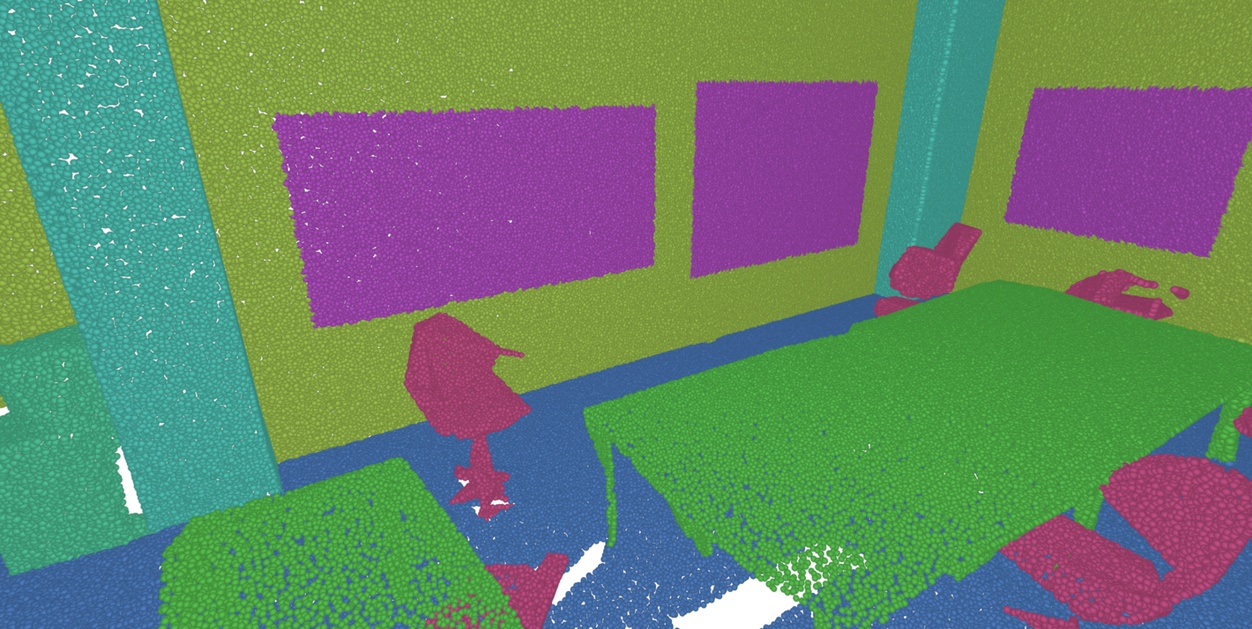}&\includegraphics[width=\linewidth]{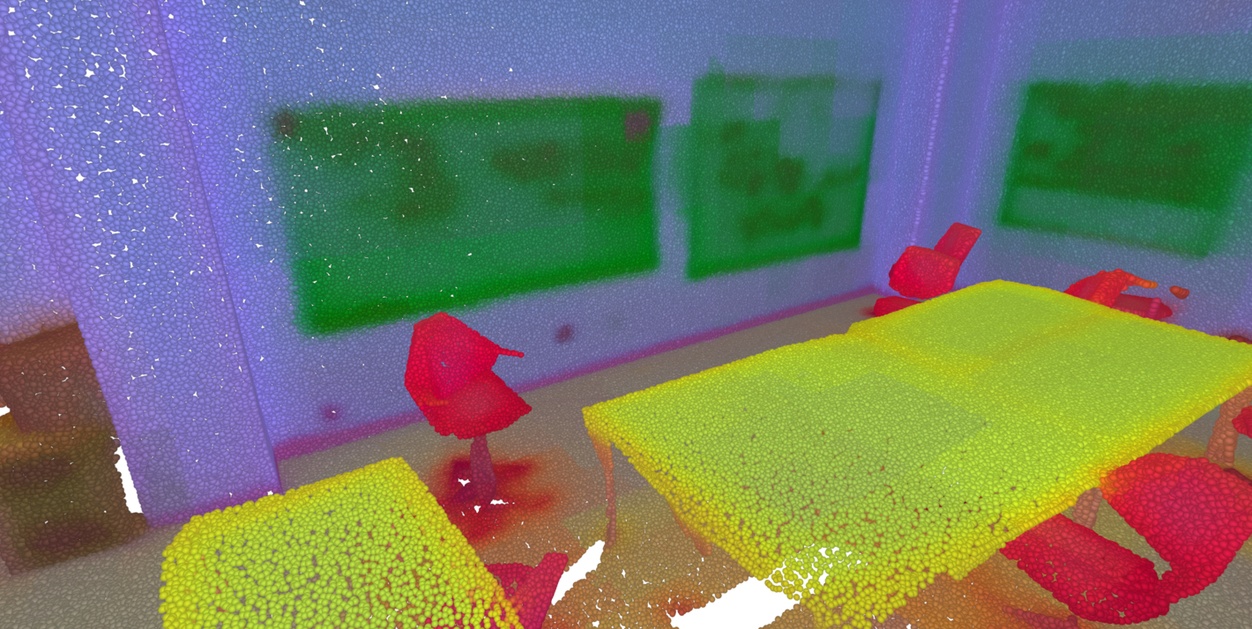}\\[-3pt]

        \raisebox{15pt}{\rotatebox{90}{\scriptsize ScanNet}}&\adjincludegraphics[width=\linewidth, trim={{.0000\width} {.0\height} {.1\width} {.1\height}}, clip]{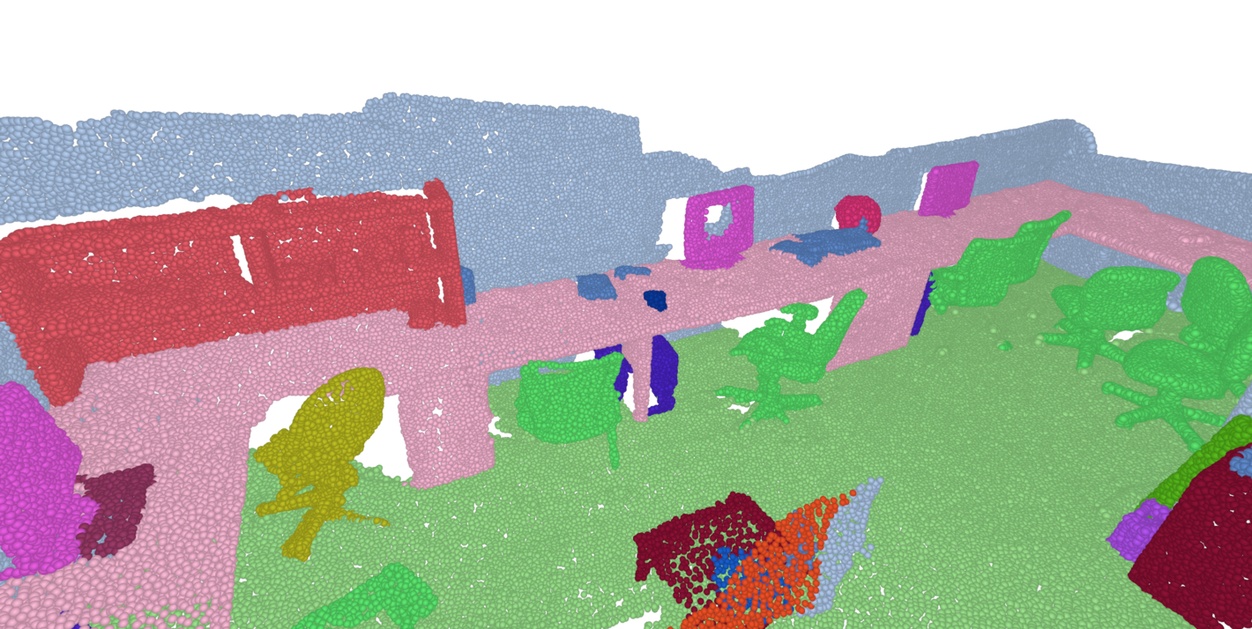}&\adjincludegraphics[width=\linewidth, trim={{.0000\width} {.0\height} {.1\width} {.1\height}}, clip]{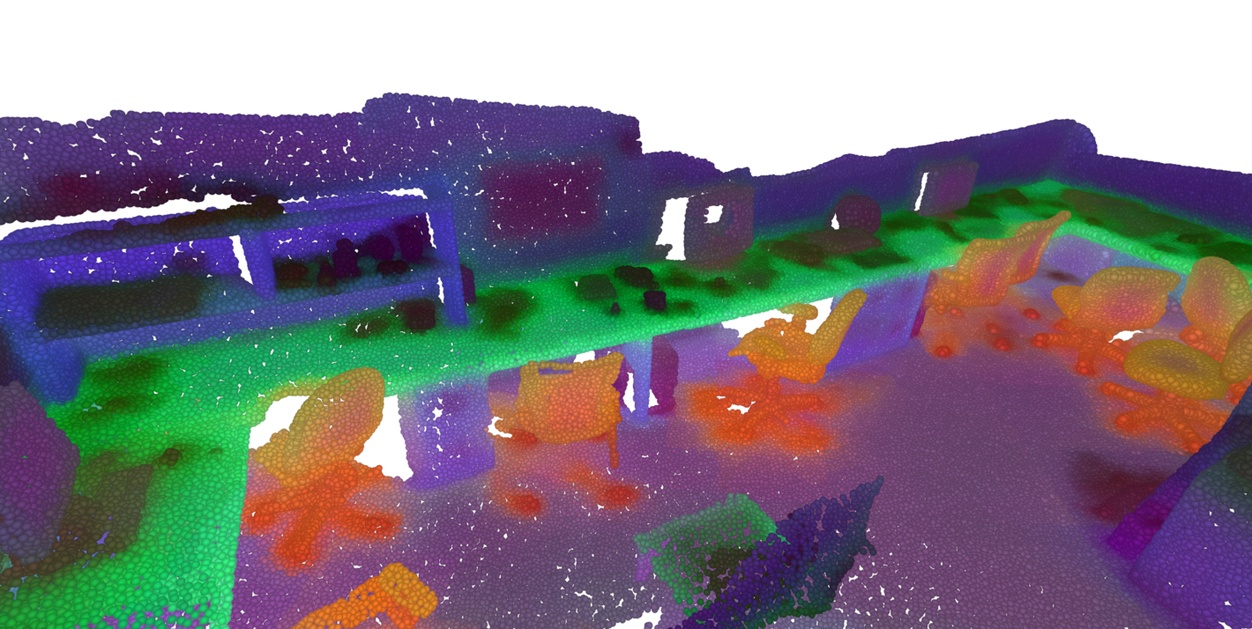}\\[-3pt]\raisebox{5pt}{\rotatebox{90}{\scriptsize SemanticKITTI}}&\adjincludegraphics[width=\linewidth, trim={{.0000\width} {.0\height} {.2\width} {.2\height}}, clip]{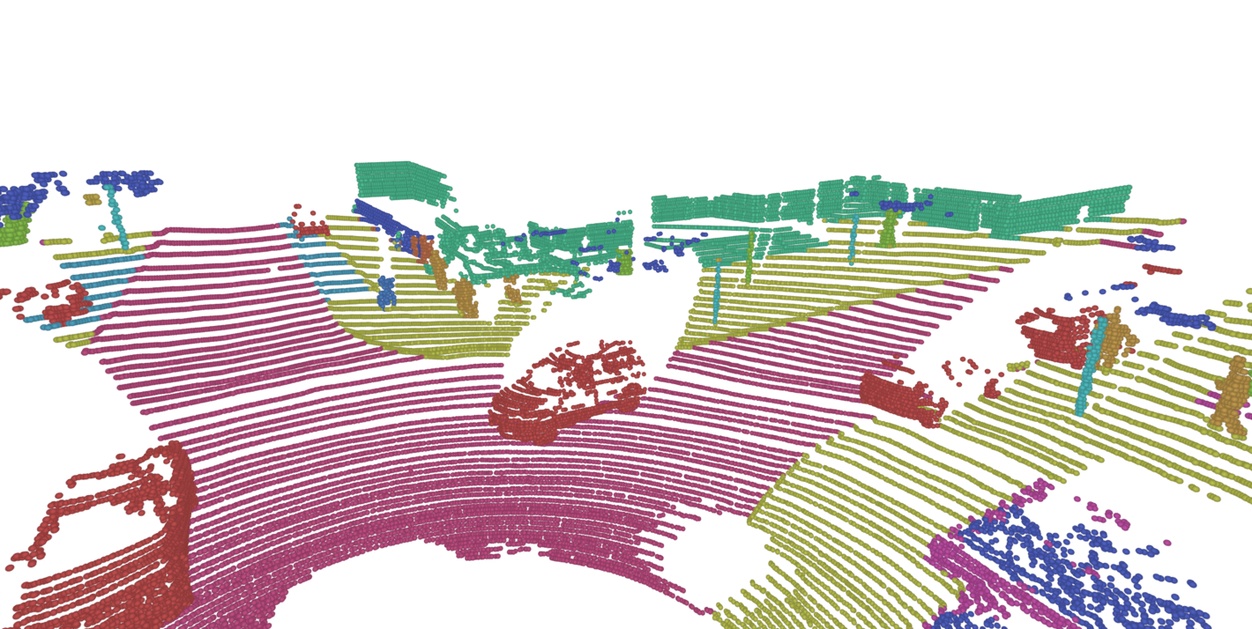}&\adjincludegraphics[width=\linewidth, trim={{.0000\width} {.0\height} {.2\width} {.2\height}}, clip]{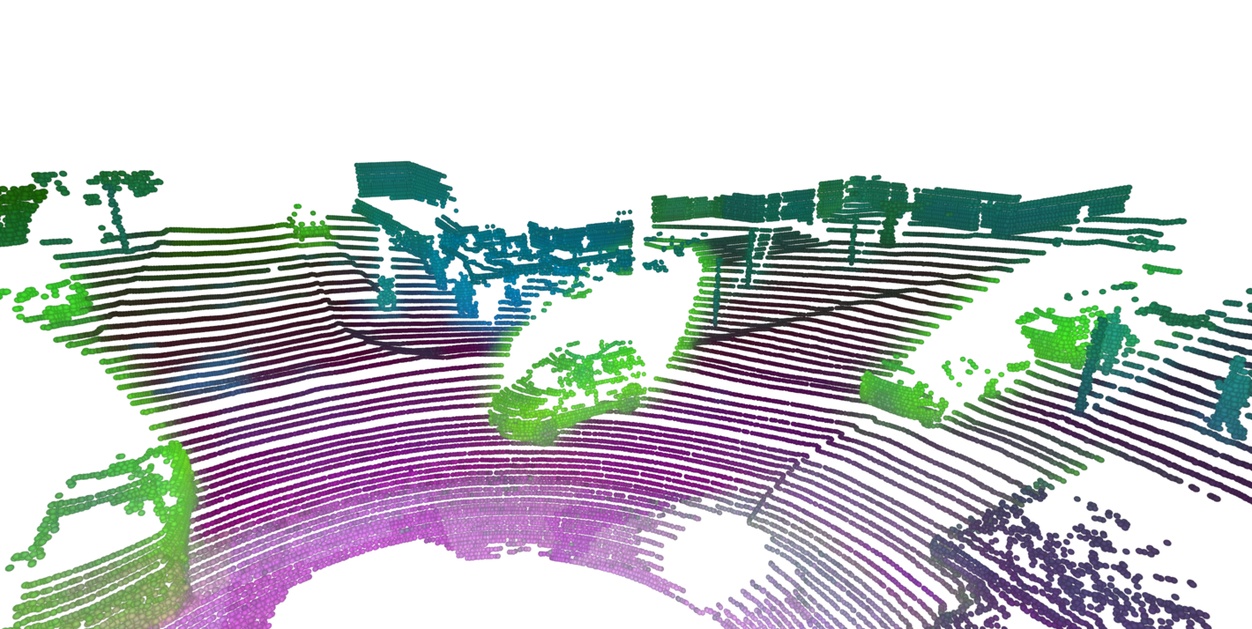}\\[-4pt]&\scriptsize Ground-Truth Semantic Segmentation&\scriptsize Distilled DINOv2 Features\\\end{tabularx}
    \vspace{-5pt}
    \caption{\textbf{\oursdist{} distillation.} We show PCA projected features from \oursdist{} after distillation. Many objects in the ground-truth segmentation are clearly separable in the predicted features, indicating the value of DINOv2 features for 3D segmentation.
    The colors are not expected to match, but color \enquote{clusters} should align.}
    \label{fig:distill_pca}
\end{figure}

%% file: tab/X_ablation_injection_point.tex
\begin{table}[t]
    \small
    \centering
    \renewcommand{\tabcolsep}{1pt}
    \begin{tabularx}{\linewidth}{lYY}
        \toprule
        Injection method  & ScanNet200  & nuScenes \\
        \midrule
        \textit{None}\; (\ie, PTv3~\cite{wu2024ptv3})  & 35.2 & 79.9 \\
        \midrule
        \multicolumn{3}{c}{\textit{2D image feature injection}} \\
        Before 3D encoder~\cite{jaritz2019mvpnet,robert2022DeepViewAgg} & 40.1 & 82.3 \\
        Between two 3D encoder--decoders~\cite{yang2023DMFnet} & 40.4 & 82.2 \\
        In 3D decoder (all blocks) (\ie, \textbf{\ours{}}) & \textbf{41.2} & \textbf{83.1} \\
        In 3D decoder (last block only) & 40.1 & 82.5 \\
        After 3D decoder & 37.6 & 82.5 \\ 
        \midrule
        \multicolumn{3}{c}{\textit{Other}} \\
        \texttt{[cls]} tokens in 3D decoder (all blocks) & 37.4 & 79.4    \\
        \bottomrule
    \end{tabularx}
    \caption{\textbf{Comparison of different 2D--3D injection methods.} We compare different methods to inject 2D features, ordered from \textit{early} to \textit{late} injection. All experiments use DINOv2-L.}
    \label{tab:injection_place}
\end{table}

%% file: tab/X_ablation_vit_backbone.tex
\begin{table}[t]
    \renewcommand{\tabcolsep}{5pt}
    \centering
    \small
    \begin{tabularx}{\linewidth}{XlYY}
        \toprule
\multicolumn{2}{l}{Image backbone}   & \multirow{2}[2]{*}{ScanNet200} & \multirow{2}[2]{*}{nuScenes}                 \\
        \cmidrule{1-2}
        Pretraining                          & Model                                                                         \\

        \midrule
        ---                                  & ---                    & 35.2                         & 80.4          \\

        IN21k~\cite{steiner2022vitaugreg}    & ViT-L                          & 38.2                         & 80.2          \\
        AIMv2~\cite{fini2024aimv2}    & ViT-L                          & 39.1                         & 82.8          \\

        SigLIP 2~\cite{tschannen2025siglip2} & ViT-g                          & 38.1                         & 83.6          \\

        DINOv2~\cite{oquab2023dinov2}        & ViT-S                          & 38.2                         & 82.8          \\

        DINOv2~\cite{oquab2023dinov2}        & ViT-B                          & 40.7                         & 83.0          \\

        DINOv2~\cite{oquab2023dinov2}        & ViT-L                          & 41.2                & 83.1          \\

        DINOv2~\cite{oquab2023dinov2}        & ViT-g                          & 40.8                         & \textbf{84.2} \\

        DINOv3~\cite{simeoni2025dinov3}        & ViT-L                          & \textbf{42.3}                         & 83.9 \\

        \bottomrule
    \end{tabularx}
    \caption{\textbf{Injection with different image backbones.} We compare different image backbones and pretraining schemes for \ours{}.}
    \label{tab:backbone}
\end{table}

%% file: tab/4_indoor_efficient.tex
\begin{table}[t]
    \centering
    \renewcommand{\tabcolsep}{3.5pt}
    \small
    \begin{tabularx}{\linewidth}{cYYYYY}
        \toprule
        \% Data              & Scratch & CSC~\cite{hou2021contrastive}    & MSC~\cite{wu2023masked}     & PPT~\cite{wu2024ppt}     & \oursdist{} \\\midrule
        \progressbar{1}   & 26.0 & 28.9\qinc{2.9}                    & 29.2\qinc{3.2}              & 31.3\qinc{5.3}           & \textbf{34.1}\qinc{8.1}   
        \\
        \progressbar{5}   & 47.8 & 49.8\qinc{2.0}                    & 50.7\qinc{2.9}              & 52.2\qinc{4.4}           & \textbf{56.6}\qinc{8.8}          
        \\
        \progressbar{10}   & 56.7 & 59.4\qinc{2.7}                    & 61.0\qinc{4.3}              & 62.8\qinc{6.1}           & \textbf{65.2}\qinc{8.5}
        \\
        \progressbar{20}   & 62.9 & 64.6\qinc{1.7}                    & 64.9\qinc{2.0}              & 66.4\qinc{3.5}           & \textbf{68.3}\qinc{5.4} \\
        \progressbar{100}             & 72.2 & 73.8\qinc{1.6}                    & 75.3\qinc{3.1}              & 75.8\qinc{3.6}           & \textbf{76.2}\qinc{4.0}
        \\
        \bottomrule
    \end{tabularx}
\caption{
        \textbf{ScanNet \enquote{limited reconstructions} benchmark with the MinkUNet~\cite{choy20194d} 3D backbone.}
        All methods perform pretraining on unannotated raw data, followed by segmentation fine-tuning on fixed labeled subsets of ScanNet.
        For fair comparison, all methods use a MinkUNet backbone (not PTv3).
    }
    \label{tab:performance_limited_data_finetuning}
\end{table}

%% file: tab/distribution_images.tex
\begin{table}[t]
    \small
    \centering
    \renewcommand{\tabcolsep}{2pt}
    \begin{tabularx}{\linewidth}{lcYYcYY}
        \toprule
        \multirow{2}[2]{*}{Method} &&
\multicolumn{2}{c}{ScanNet200} &&
        \multicolumn{2}{c}{SemanticKITTI} \\
        \cmidrule{3-4} \cmidrule{6-7}
        &&
        Visible & Invisible &&
        Visible & Invisible \\
        \midrule

        PTv3~\cite{wu2024ptv3} && 35.2 & 34.5 && 68.5 & 68.1 \\
        \ours{} && \textbf{41.5} & \textbf{39.7} && \textbf{71.6} & \textbf{68.3}\\

        \bottomrule
    \end{tabularx}
    \caption{\textbf{Performance on visible and invisible points.} We compare the segmentation performance of points visible in at least one image to those that are never visible. For SemanticKITTI, only 18.9\,\% of labeled points are visible.}
    \label{tab:visible_invisible_split}
\end{table}

%% file: sec/5_conclusion.tex
\section{Conclusion}
\label{sec:conclusion}
With this work, to the best of our knowledge, we are the first to demonstrate that the rich semantic features of 2D VFMs, like DINOv2, can be leveraged to advance 3D segmentation performance by large margins.
First, with \ours{}, we demonstrate that injecting frozen 2D VFM features into a 3D model's decoder yields significant performance gains, achieving new state-of-the-art results.
Second, we show that 2D VFMs can enable substantial improvements even when images are unavailable during inference, by using our \oursdist{} distillation strategy to pretrain a 3D backbone.
Notably, both \ours{} and \oursdist{} require only unlabeled images and are not bound by the choice of VFM, making them readily and generally applicable.
In conclusion, given our promising results, we strongly advocate the use of 2D VFMs for 3D scene understanding whenever possible.

%% file: sec/X_suppl.tex
\section{Implementation Details}
\label{sec:supp:impl_details}

Our implementation builds upon PTv3~\cite{wu2024ptv3}. 
As such, we use the same hyperparameter settings wherever possible. 
However, unlike PTv3, \ours{} also utilizes images, for which we add image-specific augmentations.
Below, we provide detailed implementation settings for each dataset.

\paragraph{Indoor Semantic Segmentation.}

Following PTv3, we train the model with a batch size of 12 for 800 epochs on ScanNet~\cite{dai2017scannet} and ScanNet200~\cite{rozenberszki2022scannet200}, and for 3000 epochs on the smaller S3DIS~\cite{armeni2016s3dis} dataset.
For ScanNet++~\cite{yeshwanthliu2023scannetpp}, we again follow PTv3, dividing the large point clouds into overlapping \SI{6}{\meter} by \SI{6}{\meter} tiles with a stride of \SI{3}{\meter}, and training for 400 epochs.
Optimization is performed using the AdamW~\cite{loshchilov2019adamw} optimizer, with a 1cycle~\cite{smith2019onecyclelr} learning rate schedule.
The maximum learning rate is set to \num{6e-3} with the learning rate peaking at 5\,\% of the training process.
Point cloud augmentations include random rotation, random dropout of points, random scaling, random flipping, elastic distortion, and color jittering.
Grid sampling is applied with a grid size of \SI{2}{\centi\meter}.

Before being fed into DINOv2~\cite{oquab2023dinov2}, ScanNet images are resized to $420 \times 560$ pixels while S3DIS images are resized to $518 \times 518$ pixels, preserving the aspect ratio and ensuring a comparable number of image tokens to those used during DINOv2 training.
Images undergo augmentation with random horizontal flipping and color jittering.
During training, 10 images are randomly selected per 3D scene, while during validation, 10 temporally equidistant images are chosen for consistent evaluation.

We only inject features for points that are visible in the selected images.
To prevent injecting features for occluded points, we compare the depth of each projected point with the raw depth value from the RGB-D sensor and exclude points where the depth difference exceeds a small margin of error (\SI{20}{\centi\meter}). 
Additionally, to ensure that image features are consistently assigned to points at similar distances, points that are closer than \SI{1}{\meter} or farther than \SI{4}{\meter} are filtered out.
Given that the 3D scene is captured from multiple camera angles with overlapping regions, most points fall in this range in at least one image.

When distilling DINOv2~\cite{oquab2023dinov2} features into the 3D model, the ScanNet~\cite{dai2017scannet} and Structured3D~\cite{zheng2020structured3d} datasets are used for joint training.
Each batch consists of samples from only one dataset, with half of the batches derived from ScanNet and the other half from Structured3D.
We train the model for a total of 160k steps with a batch size of 12.
After distillation, the pretrained model is fine-tuned separately on the ScanNet, ScanNet200, and S3DIS datasets.
For ScanNet and S3DIS, the model is fine-tuned for 10\,\% of the default number of epochs, and, except for the segmentation head, the learning rate is set to 10\,\% of its original value.
For ScanNet200, fine-tuning is performed using the default training setup described earlier.
This approach reflects differences in dataset complexity: ScanNet and S3DIS, with fewer classes, achieve strong performance with minimal fine-tuning, while ScanNet200, which has a larger and more challenging class set, benefits from longer training to fully utilize the model's capacity.

\paragraph{Outdoor Semantic Segmentation.}
Following PTv3, the model is trained with a batch size of 12 for 50 epochs on nuScenes~\cite{caesar2020nuscenes}, SemanticKITTI~\cite{behley2019semantickitti} and Waymo~\cite{sun2020waymo}.
Again, optimization is performed using the AdamW optimizer, with a 1cycle learning rate schedule.
The maximum learning rate is set to \num{2e-3} with the learning rate peaking at 4\,\% of the training process.
Point cloud augmentations include random rotation, random scaling, and random flipping.
Grid sampling is performed with a grid size of \SI{5}{\centi\meter}.

To obtain DINOv2 features, nuScenes~\cite{caesar2020nuscenes} images are resized to $378 \times 672$ pixels, SemanticKITTI~\cite{behley2019semantickitti} to $378 \times 1246$, and Waymo~\cite{sun2020waymo} to $308 \times 672$.
This keeps the original aspect ratios while also yielding a number of image tokens similar to what was used during DINOv2's training.
We make an exception for the wide-aspect-ratio images in SemanticKITTI, using a higher resolution to avoid excessively reducing the vertical resolution.
As with the indoor datasets, images undergo random horizontal flipping and color jittering augmentations.

For distillation, we jointly train the model on the nuScenes, SemanticKITTI, and Waymo datasets.
Each batch consists of samples of one dataset and batches are equally distributed across the datasets.
We train the model for a total of 350k steps with a batch size of 12.
After distillation, the pretrained model is fine-tuned separately on nuScenes, SemanticKITTI, and Waymo.
For the SemanticKITTI dataset, the model is fine-tuned for 10\,\% of the default number of epochs and, except for the segmentation head, the learning rate is set to 10\,\% of its original value.
For nuScenes and Waymo, fine-tuning is performed using the default training setup described earlier.

\section{Impact of Image Selection Strategy}
\input{tab/number_visible_images}
In \cref{tab:number_used_images}, we analyze the impact of varying the number of images and the image selection strategy during inference while using the same trained model.
For ScanNet200, we evaluate two models, one trained with 6 images per scene (mIoU\textsubscript{6}) and another with 10 (mIoU\textsubscript{10}) to evaluate the impact of the number of training images.
For nuScenes, we only evaluate a model trained on all available images.
We find that, on ScanNet200, the model trained with 6 images benefits from seeing more images during inference than during training, and performs better than the one trained with 10 images when fewer images are used during inference.
Furthermore, both models demonstrate robustness to seeing fewer images during inference, with performance degrading gracefully as they are trained with randomly selected images.
However, on nuScenes, the model always receives the full surrounding visual context during training.
As a result, the performance degrades significantly when frames are missing during inference.
Finally, on ScanNet200, we also show the effectiveness of using temporally-equidistant images during inference, resulting in higher point coverage and consistently outperforming the alternative of selecting frames randomly.

\section{Qualitative Results}

\paragraph{Distillation.}
We apply Principal Component Analysis (PCA) to the point features generated by the distilled \oursdist{} model and visualize the first three principal components as RGB colors.
This provides a qualitative view of the semantic information captured during distillation pretraining.
Note that this model operates without image input during inference.

\cref{fig:supp:pca_scannet_1,fig:supp:pca_scannet_2,fig:supp:pca_kitti,fig:supp:pca_nuscenes_1,fig:supp:pca_nuscenes_2} highlight the rich semantic features distilled from DINOv2.
Points belonging to the same semantic class exhibit similar colors, reflecting alignment along the first three principal axes.
For instance, in \cref{fig:supp:pca_scannet_1,fig:supp:pca_scannet_2}, objects such as tables, chairs, books, and walls consistently map to distinct colors across the scene.
These results show that the model captures fine-grained semantic representations of 3D scenes without requiring semantic labels during training.
Furthermore, \cref{fig:supp:pca_scannet_1,fig:supp:pca_scannet_2} reveal that objects like chairs and tables are segmented into finer subparts in feature space, surpassing the granularity of the datasets' annotations, which this model does not use.
This semantically rich feature representation significantly enhances the model's performance when used as initialization before fine-tuning for semantic segmentation, as is evident by the quantitative results in the main paper.

\paragraph{Semantic Segmentation.}
\cref{fig:supp:seg_scannet_1,fig:supp:seg_scannet_2,fig:supp:seg_nuscenes} compare the predictions of \ours{} and PTv3, with a focus on points where \ours{} benefits from DINOv2 features.

In \cref{fig:supp:seg_scannet_1}, we observe that PTv3 misclassifies a `cart' as a `shelf' due to the structural similarities between these classes in the 3D point cloud. 
In contrast, \ours{} accurately identifies the carts by incorporating visual cues from 2D images via DINOv2 feature injection.

In \cref{fig:supp:seg_scannet_2} we show another perspective of the same scene. 
Here, PTv3 misclassifies a `cabinet' as a `wall', as the cabinet appears as a flat surface that is flush with the wall and has a similar color, making it almost impossible to identify it as a cabinet based on the point cloud only.
However, \ours{} correctly classifies the cabinet, as it is clearly visible in one of the image views. 
This highlights how leveraging strong image features can complement and enhance geometric features, enabling the model to differentiate between structurally similar classes.

Finally, in \cref{fig:supp:seg_nuscenes}, we present an example from the nuScenes dataset.
PTv3 misclassifies a pedestrian crossing the street as a car, overlooks several pedestrians on the sidewalk, and confuses a bus with a building.
These errors occur in areas distant from the ego vehicle, where LiDAR data is sparse, and objects are represented by only a few points.
In contrast, \ours{} accurately predicts these instances, leveraging contextual information from the respective image for disambiguation.

%% file: tab/number_visible_images.tex
\begin{table}[t]
    \footnotesize
    \centering
    \renewcommand{\tabcolsep}{2pt}
    \begin{tabularx}{\linewidth}{YY c YYY c YY}
        \toprule
       \multicolumn{2}{c}{Image selection}&&
       \multicolumn{3}{c}{ScanNet200}&&
       \multicolumn{2}{c}{nuScenes} \\
       \cmidrule{1-2} \cmidrule{4-6} \cmidrule{8-9}
        \#~Images & Method && mIoU\textsubscript{6} & mIoU\textsubscript{10} & \%~Vis. && mIoU & \%~Vis.\\  
        \midrule
        0 & eq.dist. && 32.8 & 31.8& \phantom{0}0.0 && 37.5 & \phantom{0}0.0 \\
        1 & eq.dist. && 36.5 & 35.7 & \phantom{0}9.8 && 50.9 & 11.2  \\
        3 & eq.dist. && 38.2 & 38.0 & 24.6 && 72.7 & 40.5 \\
        6 & eq.dist. && 39.2 & 40.1 & 41.7 && 83.1 & 77.0 \\
        10 & eq.dist. && 39.9 & 41.2 & 56.7 && --- & --- \\
        \midrule
        6 & random && 38.6 & 38.4 & 37.0 && --- & --- \\
        10 & random && 39.2 & 40.5 & 49.5 && --- & --- \\

        \bottomrule
    \end{tabularx}

\caption{\textbf{Impact of image selection strategy during inference.} We compare two inference-time image selection methods: random selection and temporally-equidistant frame selection (eq.dist.). For ScanNet200, we compare models trained on 6 (mIoU\textsubscript{6}) or 10 random images (mIoU\textsubscript{10}). For nuScenes, the model is trained using all six camera views. During inference, we use different subsets of cameras: no camera; the front camera alone; the front, rear-left, and rear-right cameras; all six cameras.}
    \label{tab:number_used_images}
\end{table}

%% file: sec/X_suppl_floats.tex
\input{fig/supplementary/pca_scannet.tex}

\input{fig/supplementary/pca_kitti.tex}
\input{fig/supplementary/pca_nuscenes.tex}

\input{fig/supplementary/seg_scannet.tex}
\input{fig/supplementary/seg_nuscenes.tex}

%% file: fig/supplementary/pca_scannet.tex
\begin{figure*}[ht]
    \centering
    \begin{tabular}{cc}
        \includegraphics[width=0.4\textwidth]{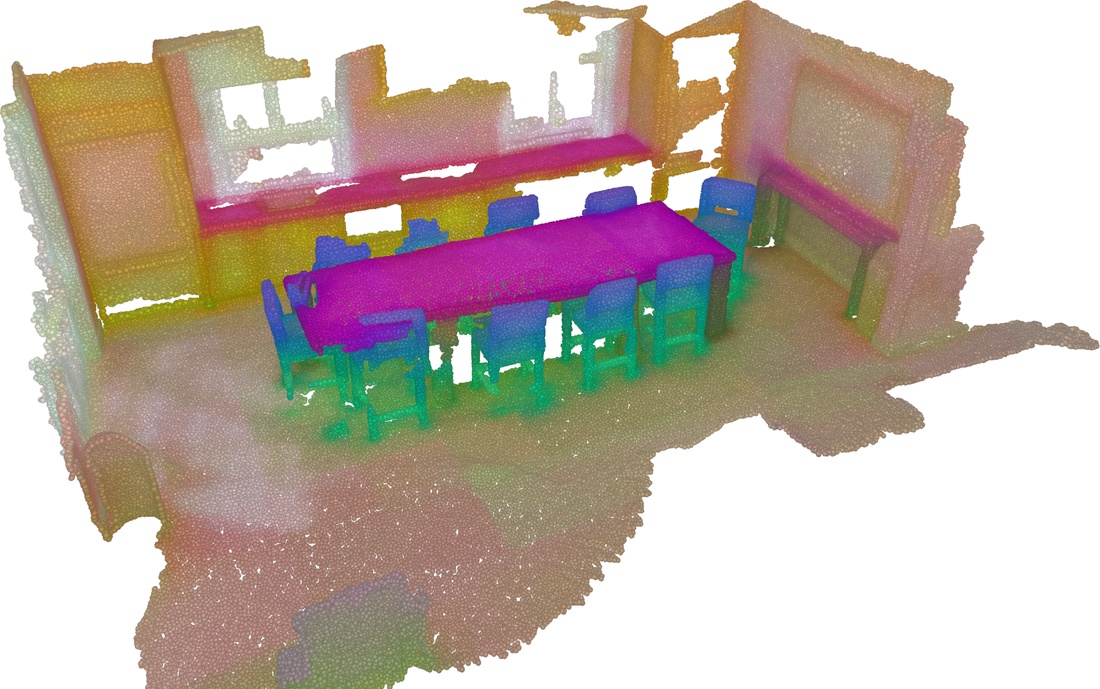} & \includegraphics[width=0.4\textwidth]{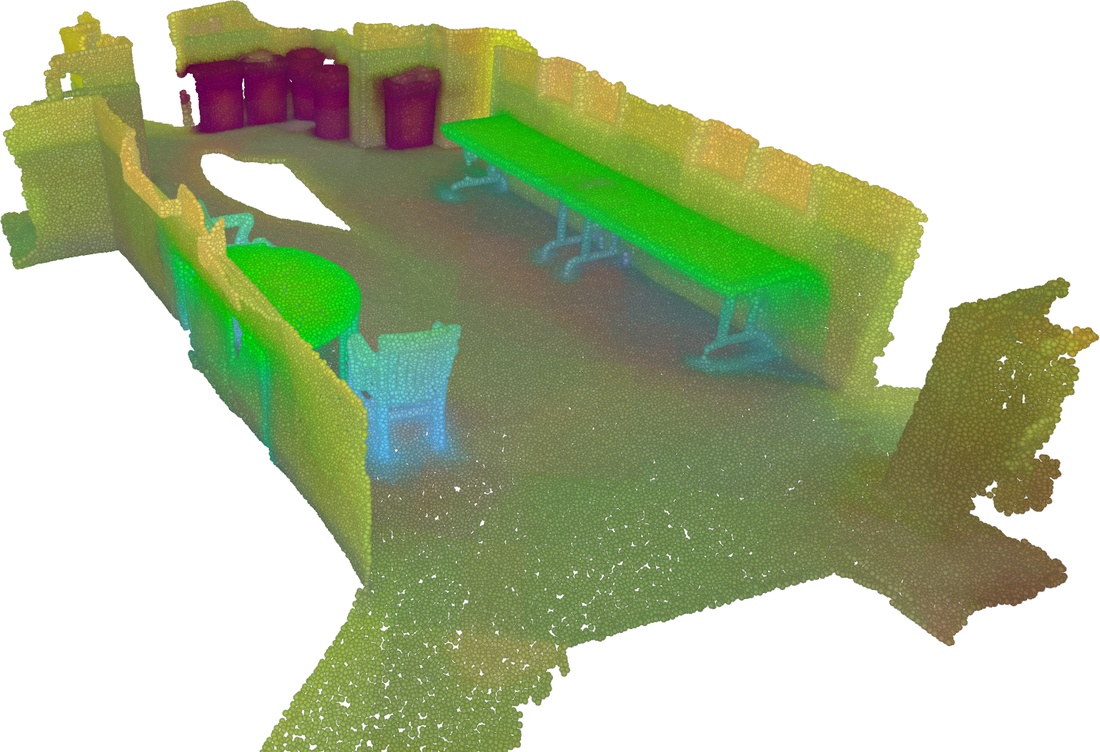} \\\includegraphics[width=0.4\textwidth]{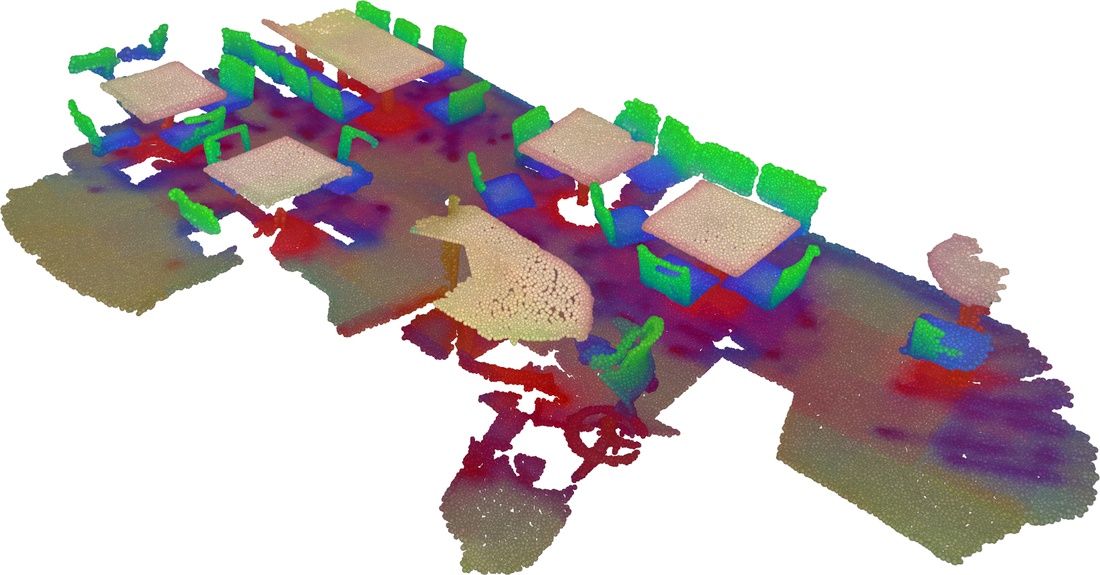} & \includegraphics[width=0.4\textwidth]{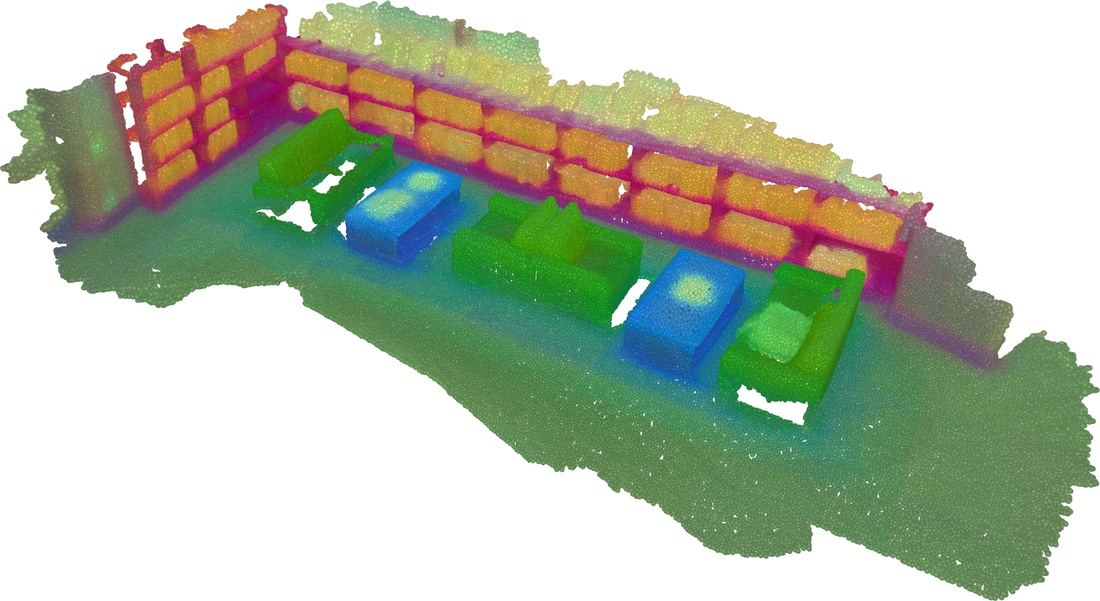} \\
        \includegraphics[width=0.4\textwidth]{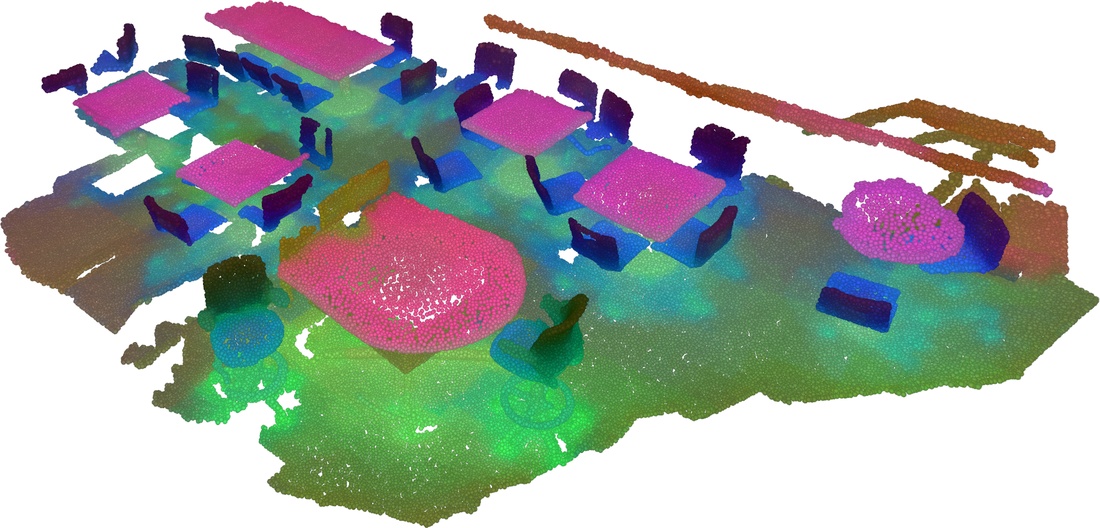} &
\includegraphics[width=0.4\textwidth]{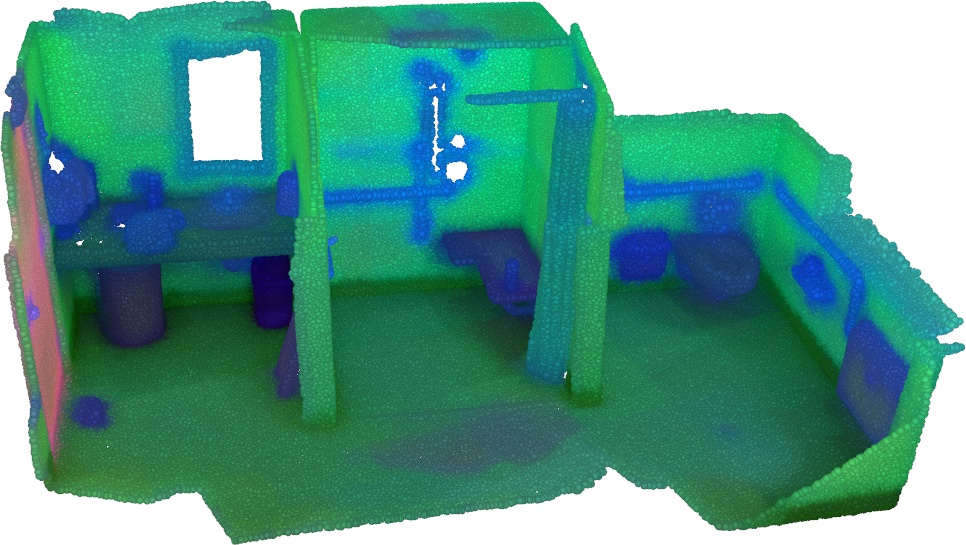} \\
        \includegraphics[width=0.4\textwidth]{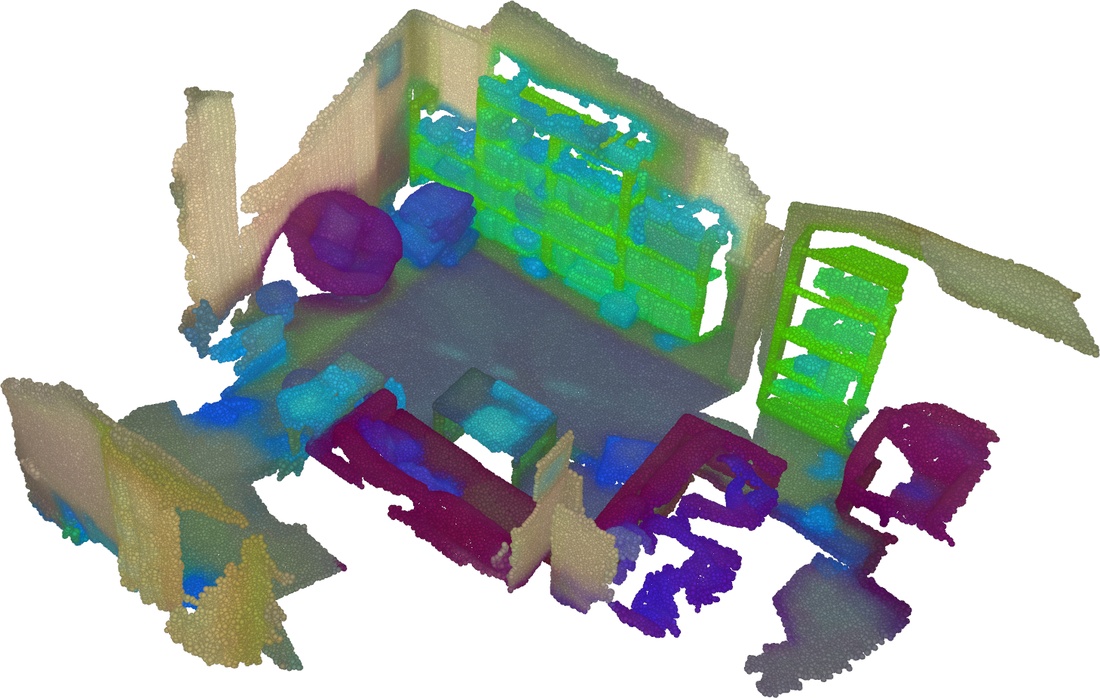} & 
        \includegraphics[width=0.4\textwidth]{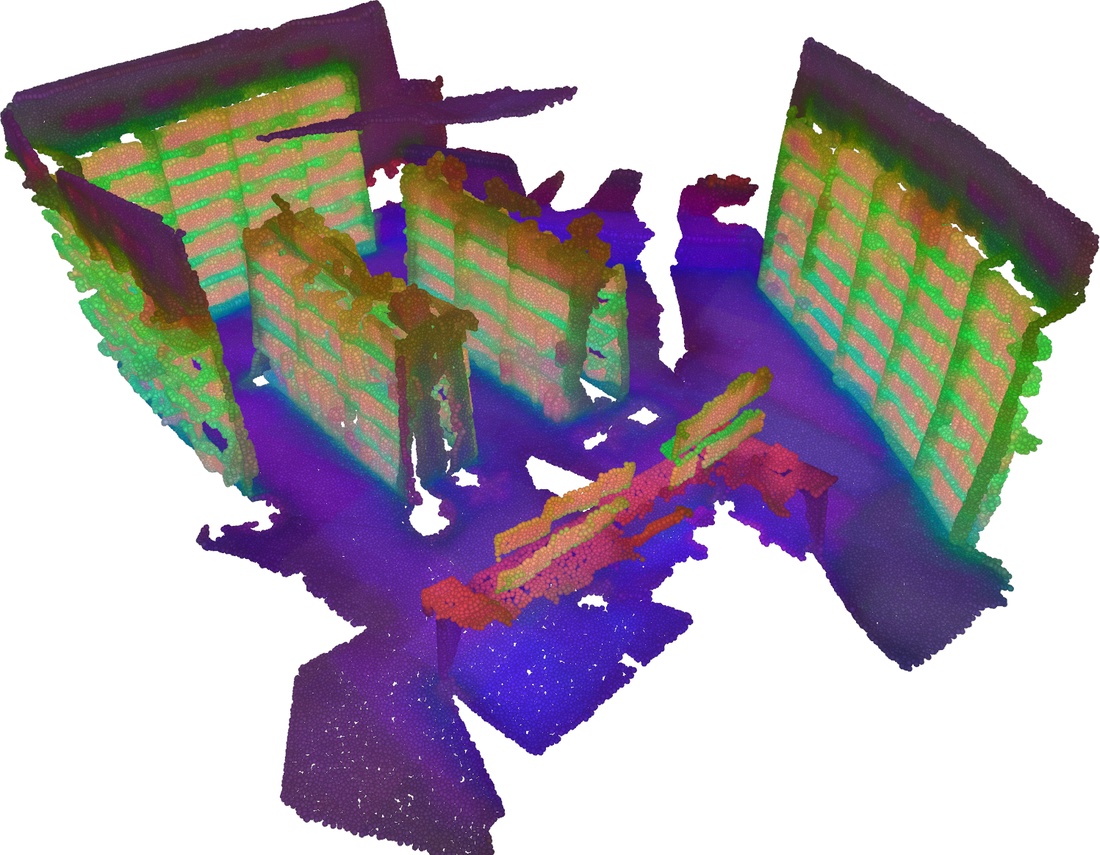} \\
    \end{tabular}
    \caption{
        \textbf{PCA visualization of distilled \oursdist{} features for ScanNet~\cite{dai2017scannet}.}
        We visualize the first three principal components of the point features learned by the \oursdist{} model.
        PCA projections are computed per scene and hence there is no correspondence of colors between figures.
    }
    \label{fig:supp:pca_scannet_1}
\end{figure*}

\begin{figure*}[ht]
    \centering
    \begin{tabular}{cc}
        \includegraphics[width=0.4\textwidth]{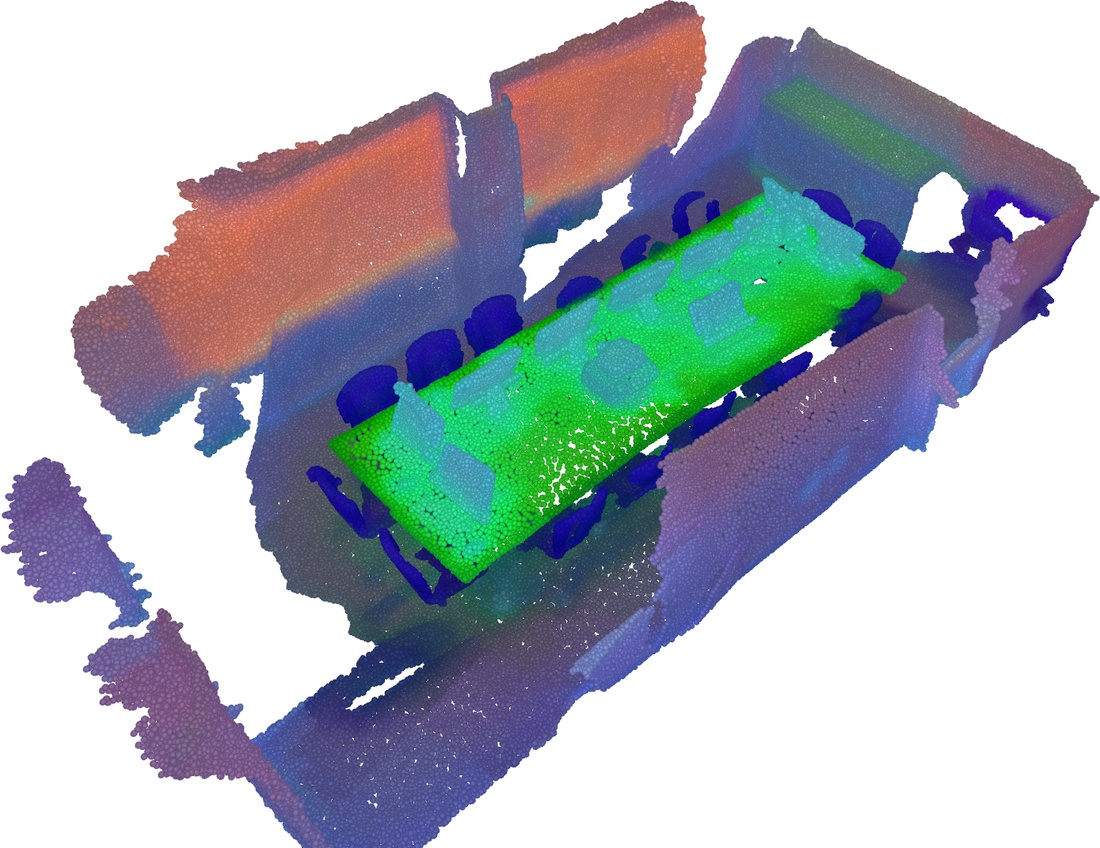} &
\includegraphics[width=0.4\textwidth]{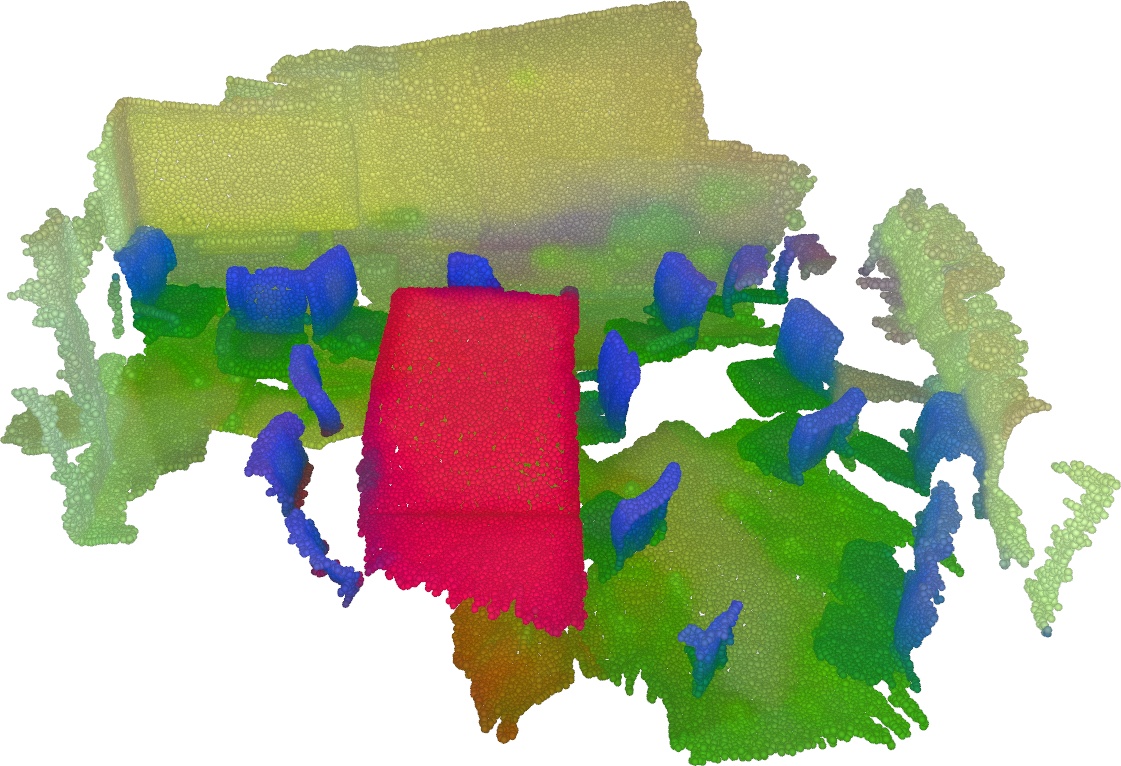} \\
        \includegraphics[width=0.4\textwidth]{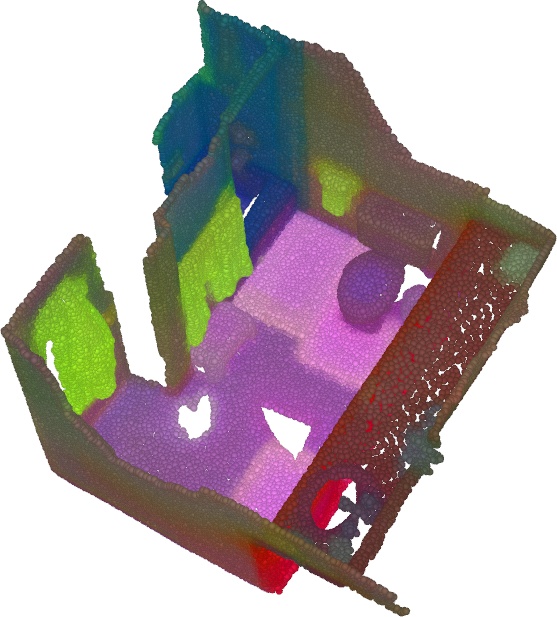} & 
        \includegraphics[width=0.4\textwidth]{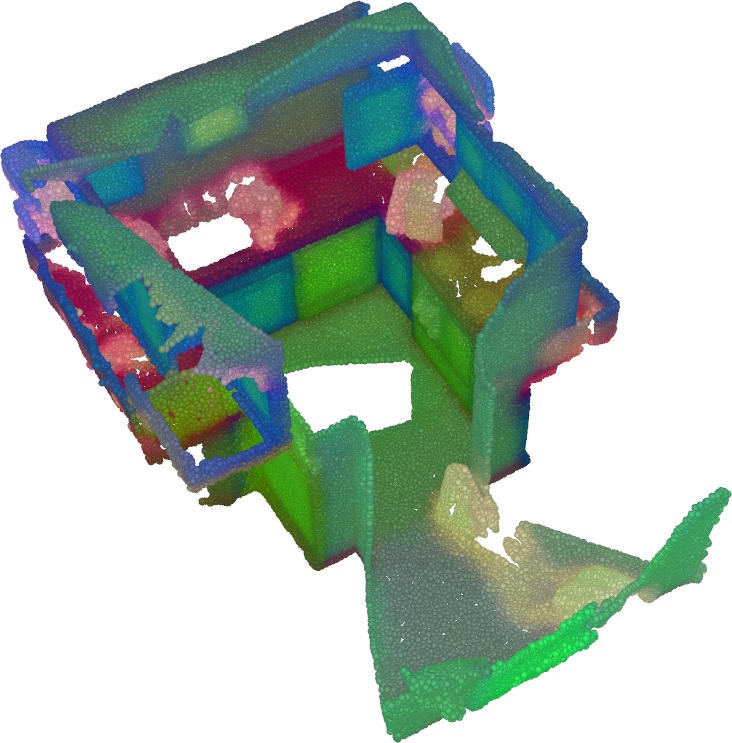} \\
\includegraphics[width=0.4\textwidth]{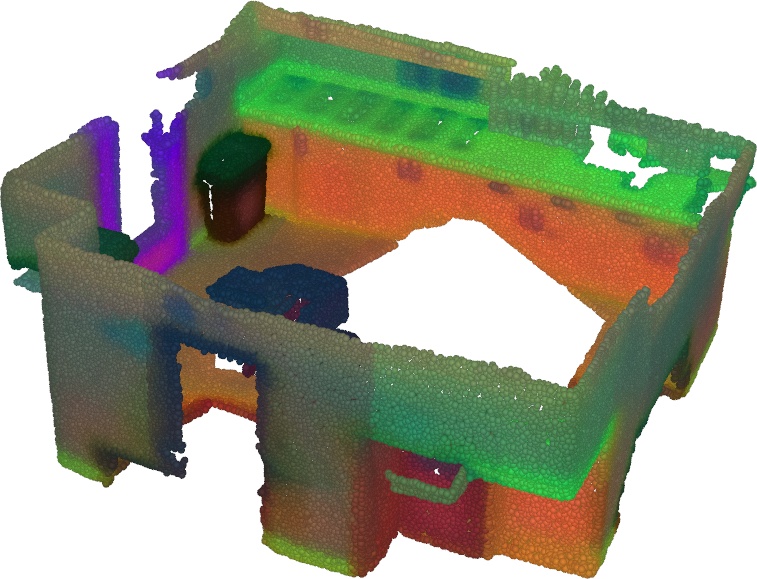} & 
        \includegraphics[width=0.4\textwidth]{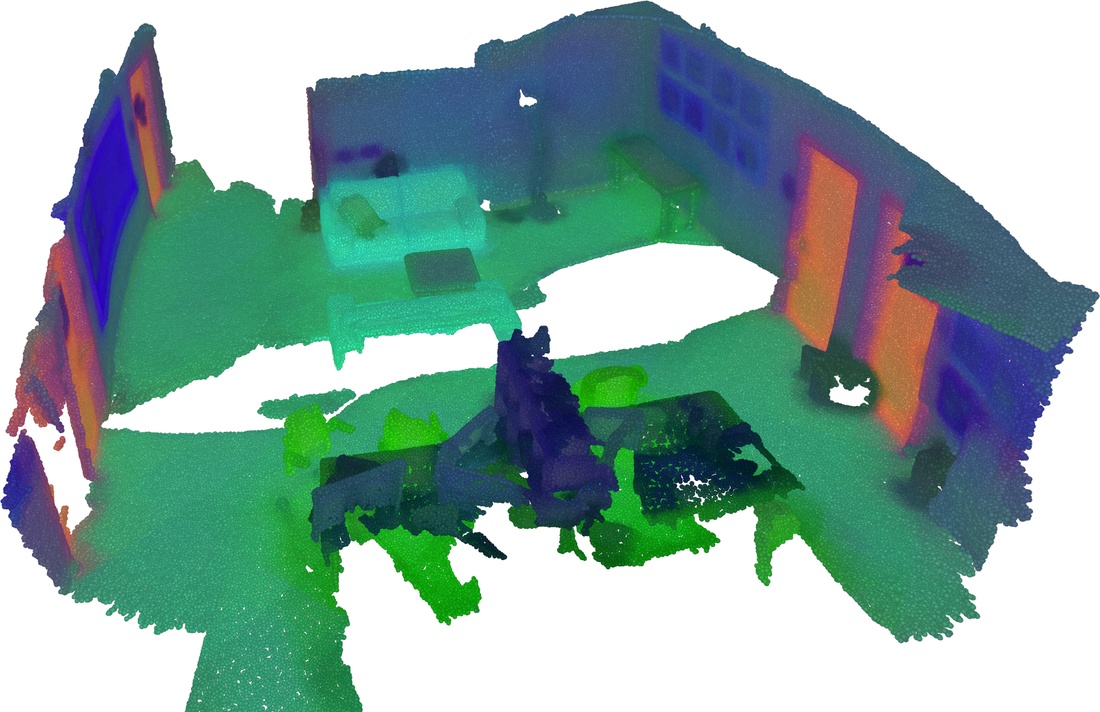} \\
    \end{tabular}
    \caption{
        \textbf{(Cont.) PCA visualization of distilled \oursdist{} features for ScanNet~\cite{dai2017scannet}.}
        We visualize the first three principal components of the point features learned by the \oursdist{} model.
        PCA projections are computed per scene and hence there is no correspondence of colors between figures.
    }
    \label{fig:supp:pca_scannet_2}
\end{figure*}

%% file: fig/supplementary/pca_kitti.tex
\begin{figure*}[ht]
    \centering
    \begin{tabular}{cc}
        \includegraphics[width=0.4\textwidth,trim={5px 5px 5px 5px},clip]{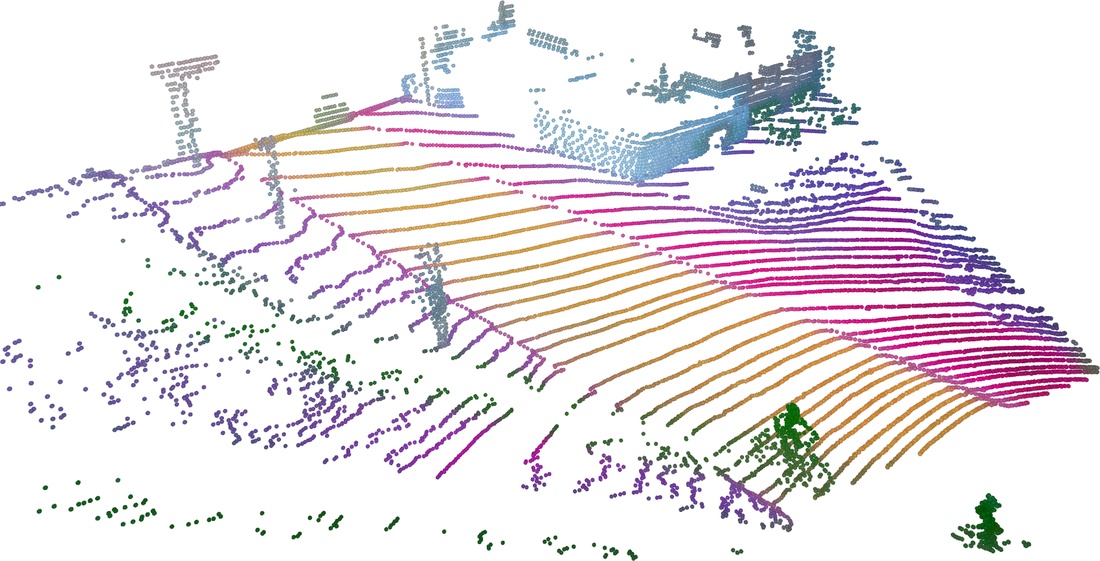} & 
        \includegraphics[width=0.4\textwidth]{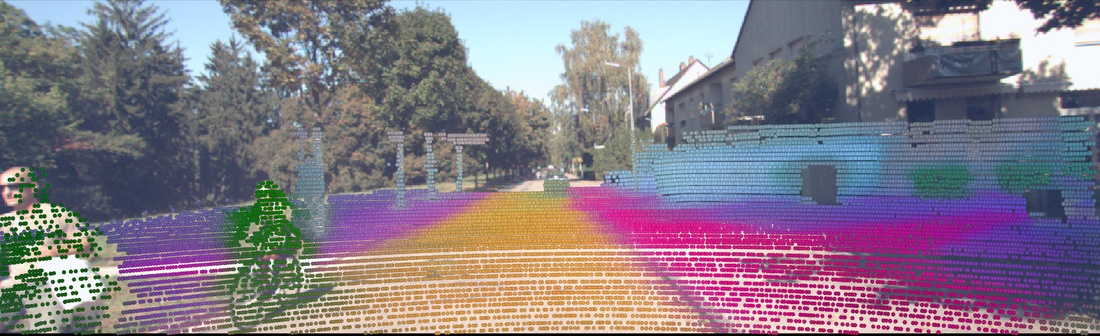} \\
        \includegraphics[width=0.4\textwidth,trim={5px 5px 5px 5px},clip]{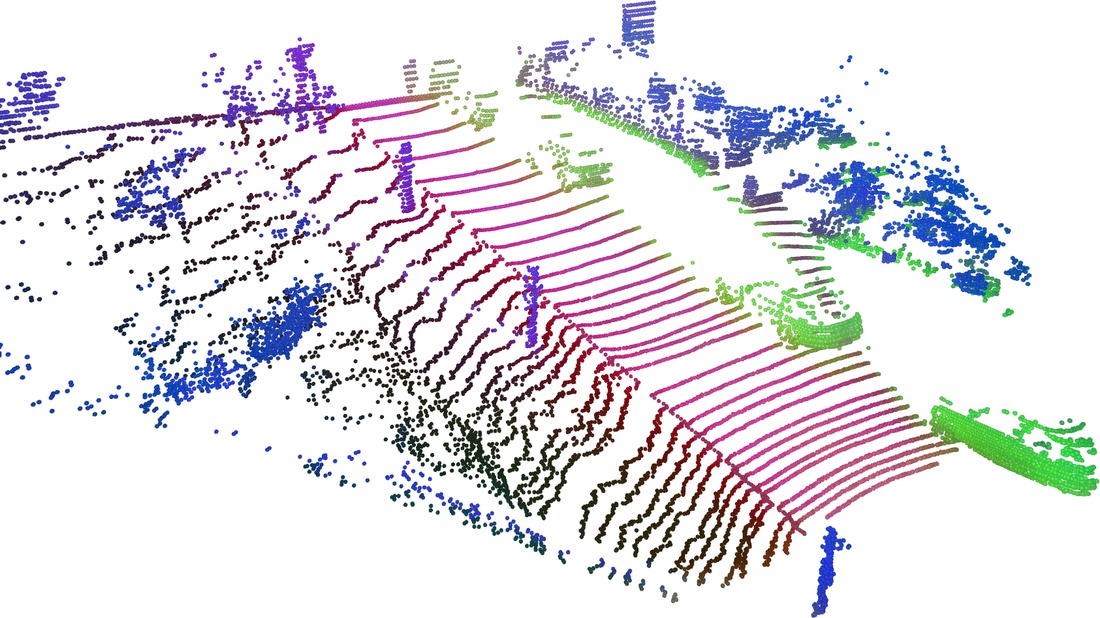} & 
        \includegraphics[width=0.4\textwidth]{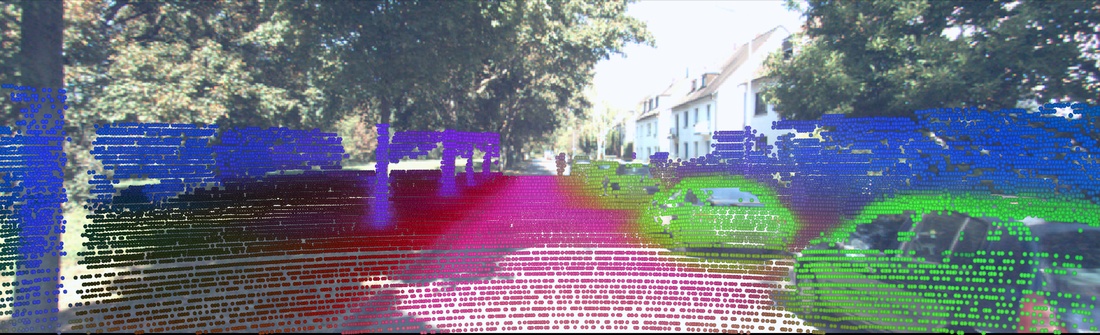} \\
        \includegraphics[width=0.4\textwidth,trim={5px 5px 5px 5px},clip]{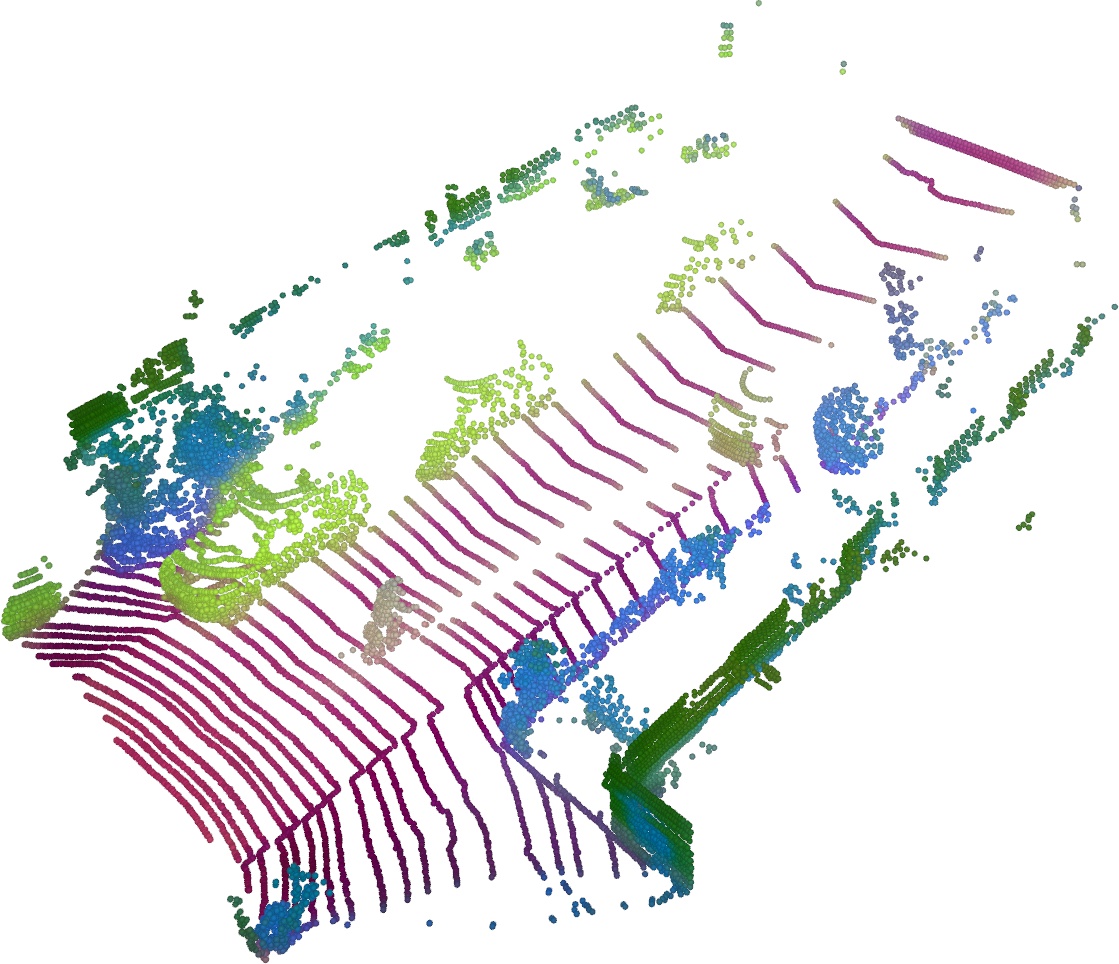} & 
        \includegraphics[width=0.4\textwidth]{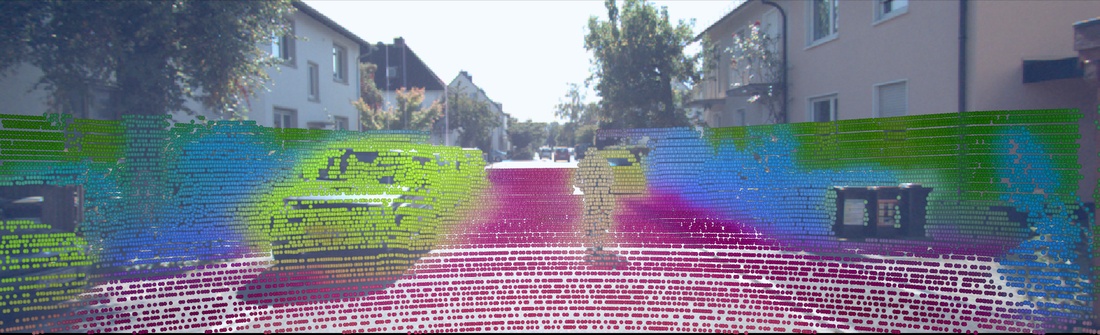} \\
        \includegraphics[width=0.4\textwidth,trim={5px 5px 5px 5px},clip]{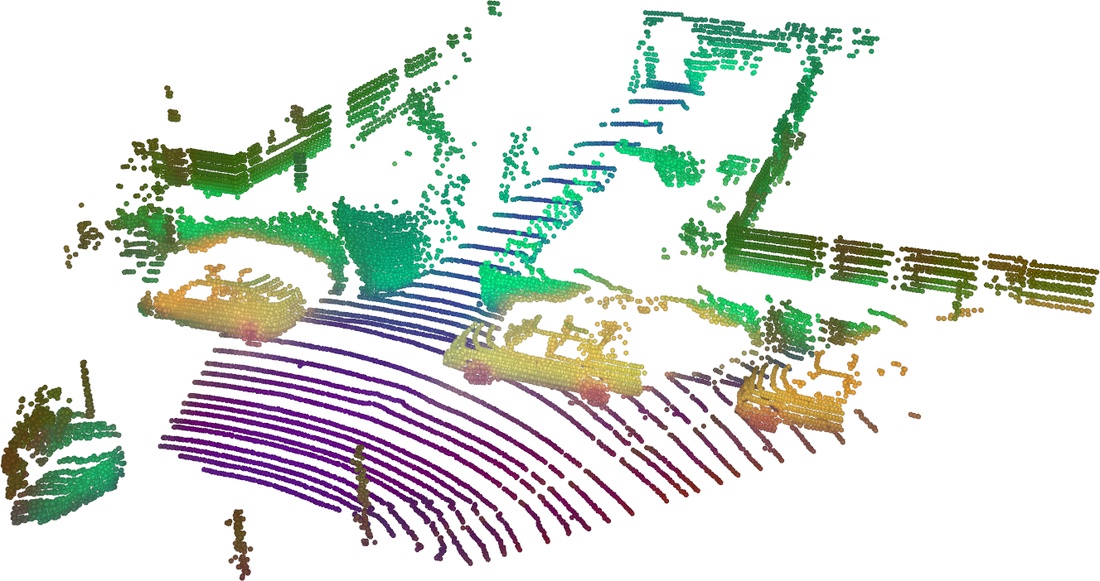} & 
        \includegraphics[width=0.4\textwidth]{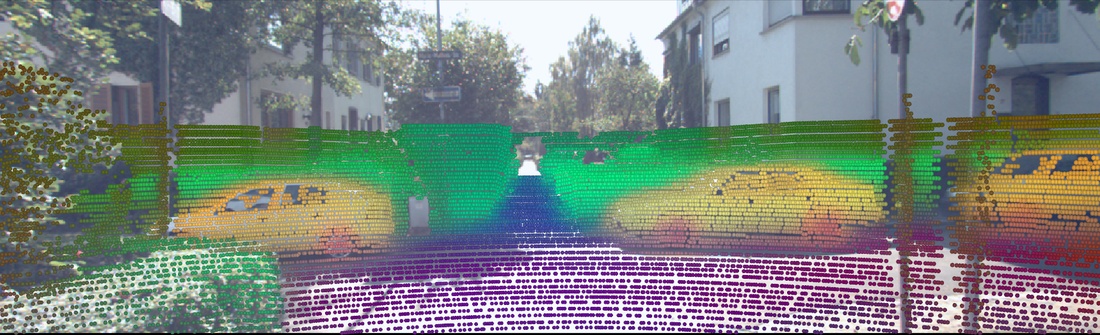} \\
    \end{tabular}
    \caption{
        \textbf{PCA visualization of distilled \oursdist{} features for SemanticKITTI~\cite{behley2019semantickitti}.}
        We visualize the first three principal components of the point features learned by the \oursdist{} model (left), as well as the corresponding 2D projections (right).
    }
    \label{fig:supp:pca_kitti}
\end{figure*}

%% file: fig/supplementary/pca_nuscenes.tex
\begin{figure*}[ht]
    \centering
    \begin{tabular}{ccc}
        \includegraphics[width=0.3\textwidth,trim={5px 5px 5px 5px},clip]{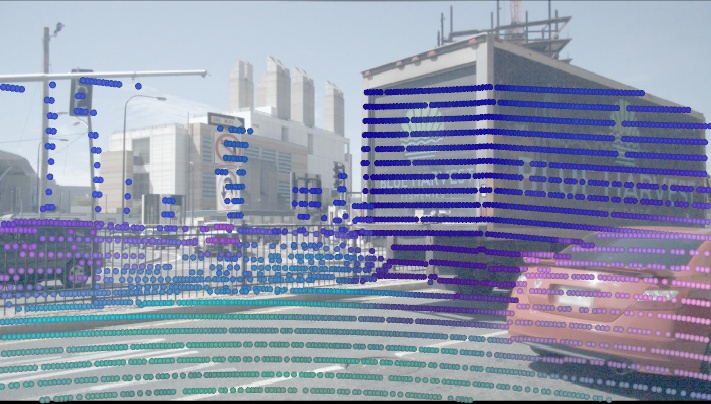} &
        \includegraphics[width=0.3\textwidth,trim={5px 5px 5px 5px},clip]{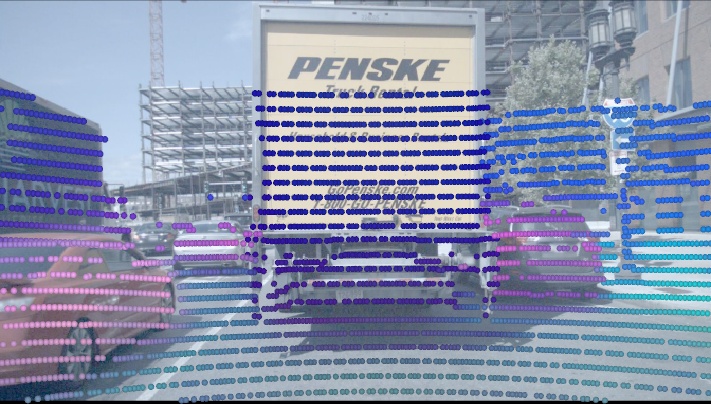} & 
        \includegraphics[width=0.3\textwidth,trim={5px 5px 5px 5px},clip]{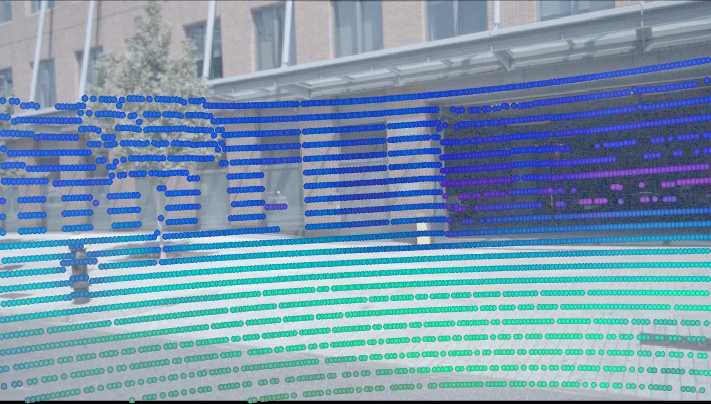} \\

        \multicolumn{3}{c}{\includegraphics[width=0.9\textwidth]{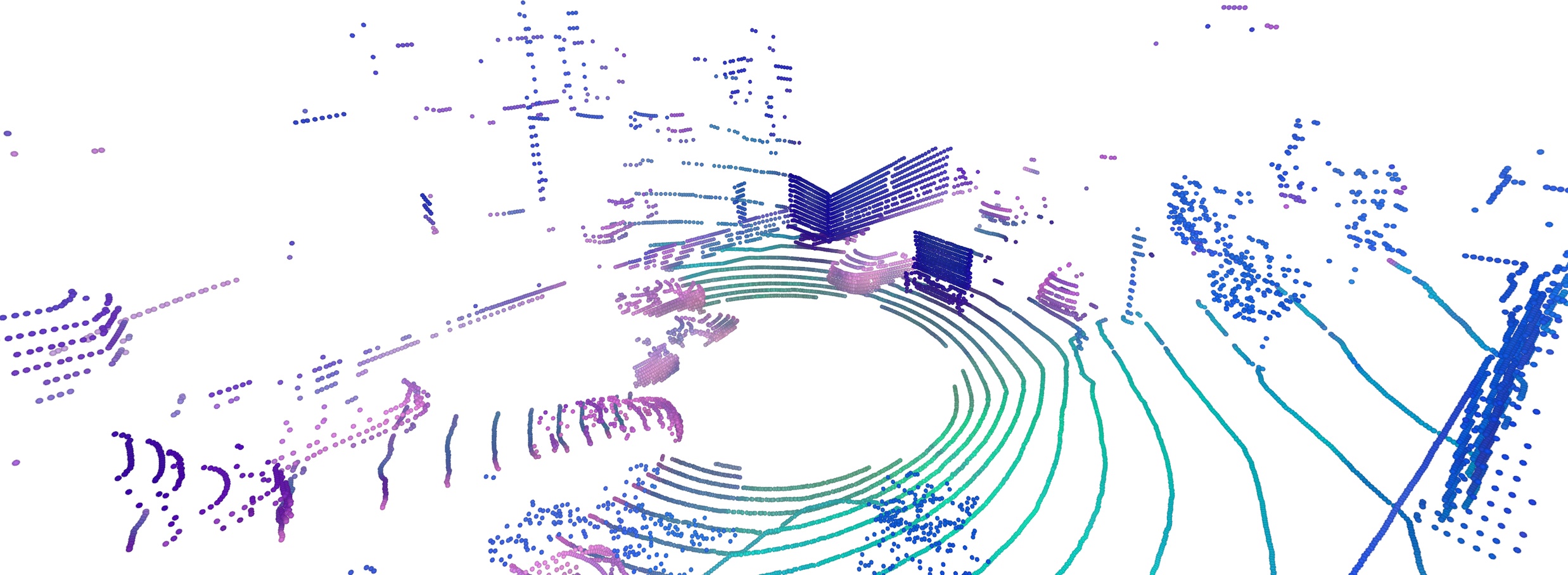}} \\

        \includegraphics[width=0.3\textwidth,trim={5px 5px 5px 5px},clip]{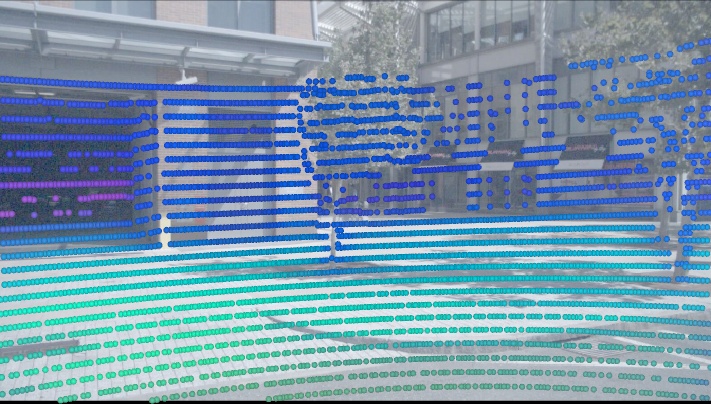} &
        \includegraphics[width=0.3\textwidth,trim={5px 5px 5px 5px},clip]{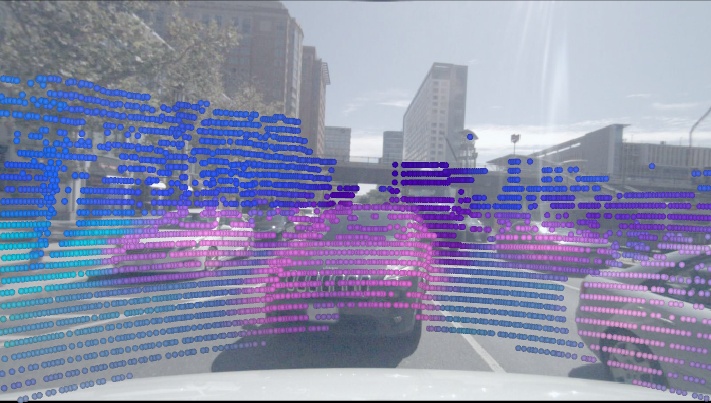} & 
        \includegraphics[width=0.3\textwidth,trim={5px 5px 5px 5px},clip]{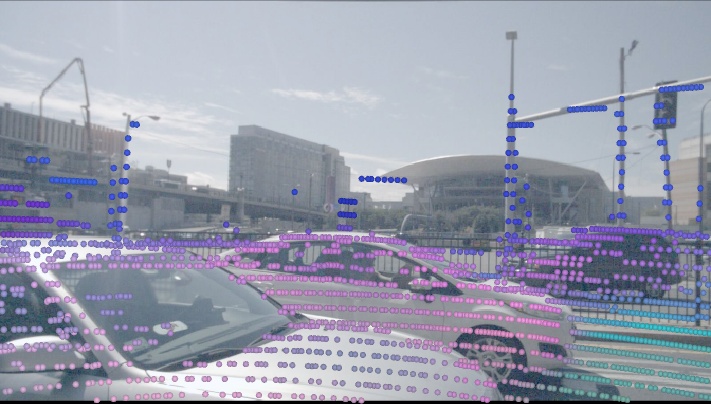} \\
    \end{tabular}
    \caption{
        \textbf{PCA visualization of distilled \oursdist{} features for nuScenes~\cite{caesar2020nuscenes}.}
        We visualize the first three principal components of the point features learned by the \oursdist{} model.
    }
    \label{fig:supp:pca_nuscenes_1}
\end{figure*}

\begin{figure*}[ht]
    \centering
    \begin{tabular}{ccc}
        \includegraphics[width=0.3\textwidth,trim={5px 5px 5px 5px},clip]{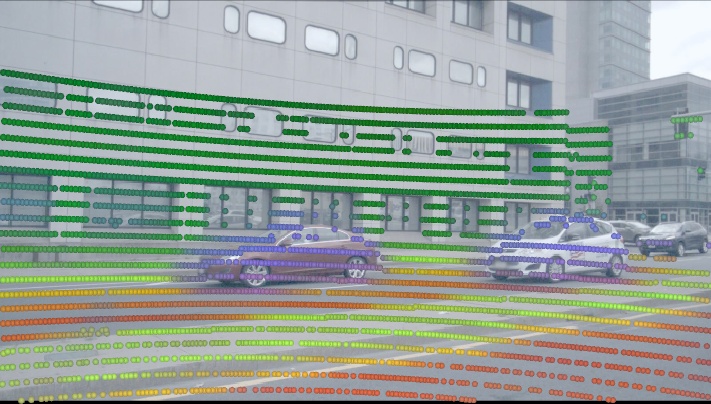} &
        \includegraphics[width=0.3\textwidth,trim={5px 5px 5px 5px},clip]{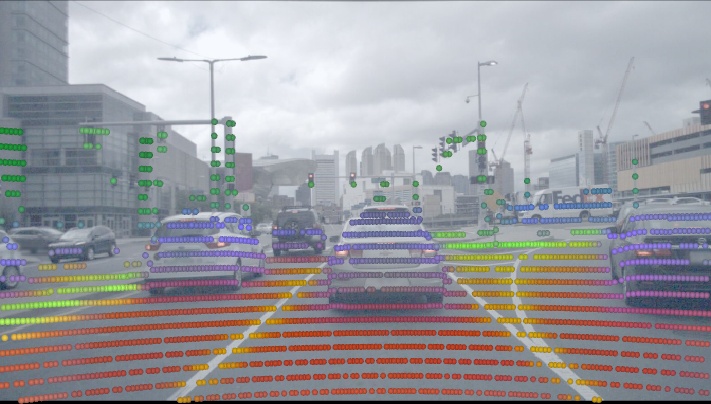} & 
        \includegraphics[width=0.3\textwidth,trim={5px 5px 5px 5px},clip]{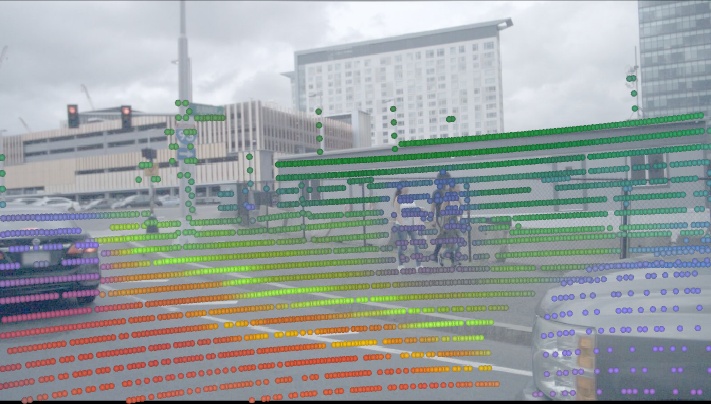} \\

        \multicolumn{3}{c}{\includegraphics[width=0.9\textwidth]{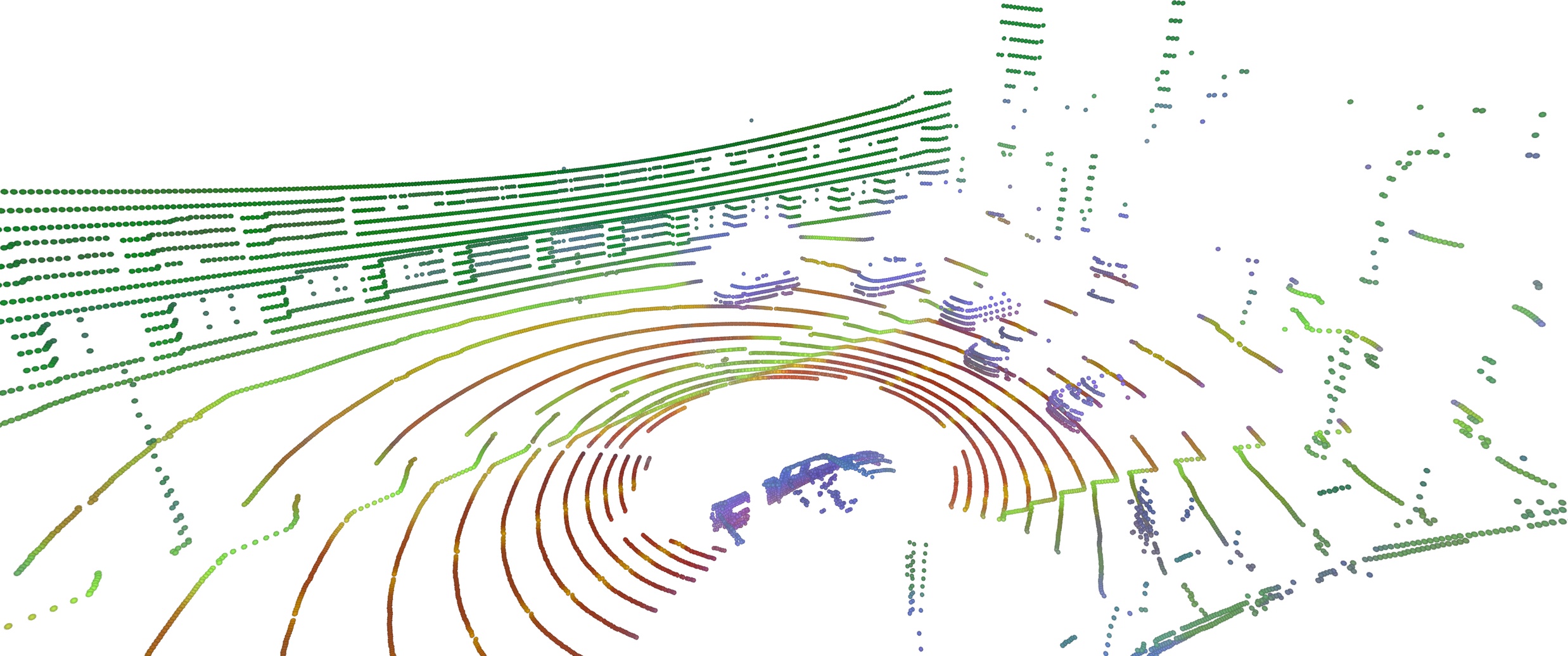}} \\

        \includegraphics[width=0.3\textwidth,trim={5px 5px 5px 5px},clip]{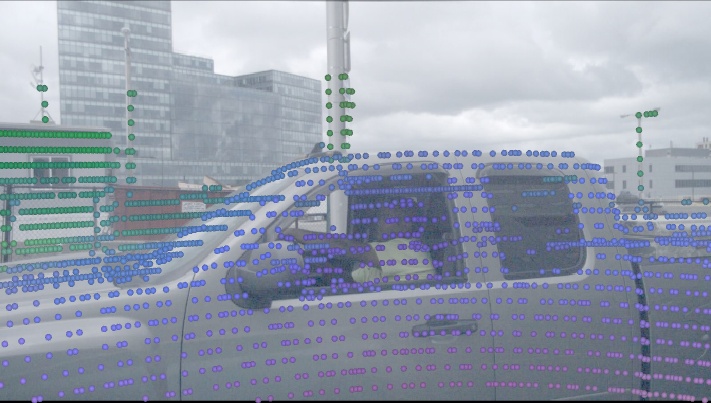} &
        \includegraphics[width=0.3\textwidth,trim={5px 5px 5px 5px},clip]{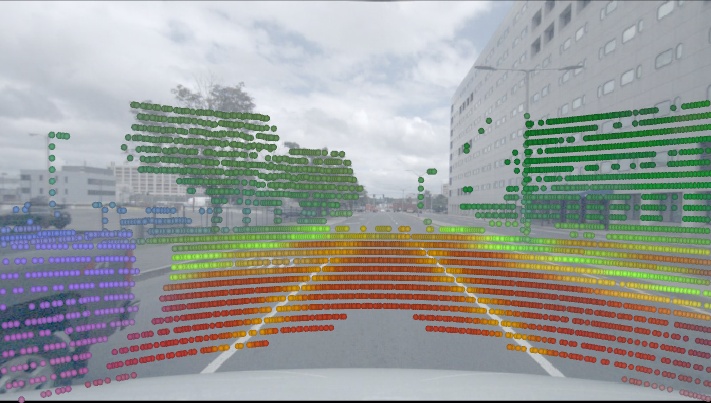} & 
        \includegraphics[width=0.3\textwidth,trim={5px 5px 5px 5px},clip]{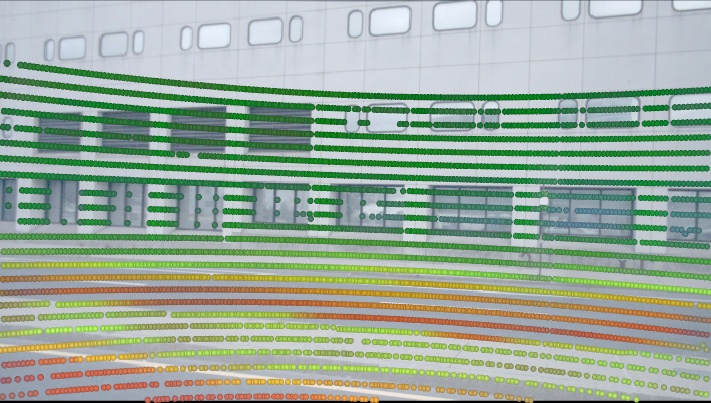} \\
    \end{tabular}
    \caption{
        \textbf{PCA visualization of distilled \oursdist{} features for nuScenes~\cite{caesar2020nuscenes}.}
        We visualize the first three principal components of the point features learned by the \oursdist{} model.
    }
    \label{fig:supp:pca_nuscenes_2}
\end{figure*}

%% file: fig/supplementary/seg_scannet.tex
\begin{figure*}[ht]
    \centering
    \newcommand{\specialfig}[1]{
        \begin{tikzpicture}
            \node[anchor=south west, inner sep=0] (image) at (0,0) {\includegraphics[width=0.4\textwidth]{#1}};
            \begin{scope}[x={(image.south east)}, y={(image.north west)}]
                \coordinate (A) at (0.35, 1.0);
                \coordinate (B) at (0.7, 1.0);
                \coordinate (C) at (0.98, 0.9);
                \coordinate (D) at (0.98, 0.3);
                \coordinate (E) at (0.4, 0.7);
                \draw[very thick, red] plot [smooth cycle, tension=0.05] coordinates {(A) (B) (C) (D) (E)};
            \end{scope}
        \end{tikzpicture}}
    \begin{tabular}{cc}
        \includegraphics[width=0.33\textwidth,trim={5px 5px 5px 5px},clip]{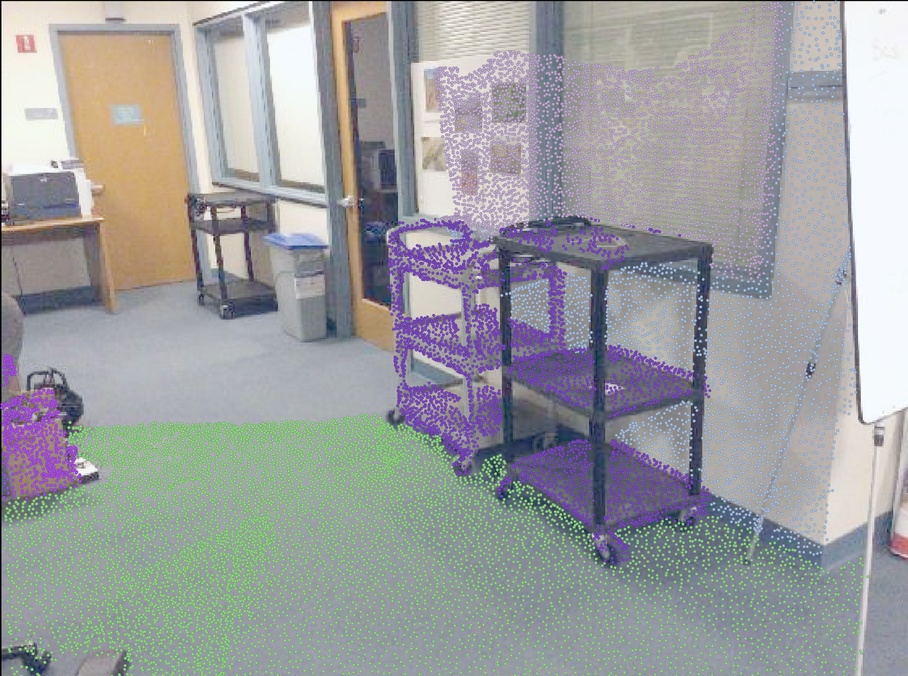}   &
\specialfig{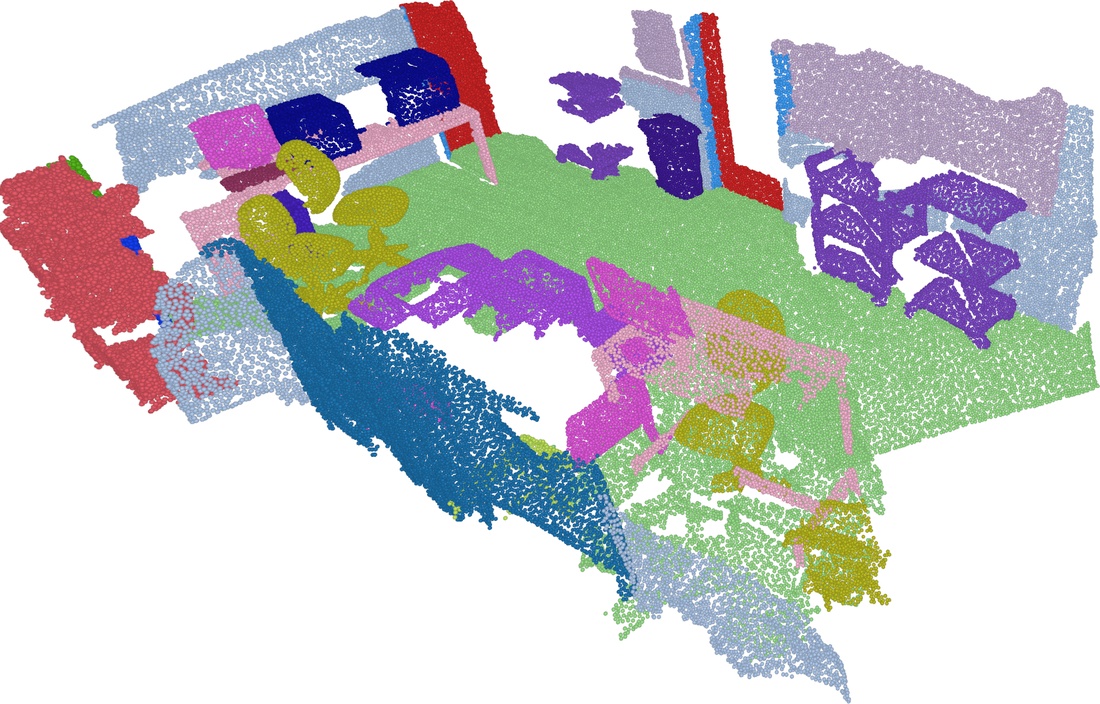}
        \\
        2D projected ground truth                                                                  & Full 3D ground truth                         \\[10pt]
        \includegraphics[width=0.33\textwidth,trim={5px 5px 5px 5px},clip]{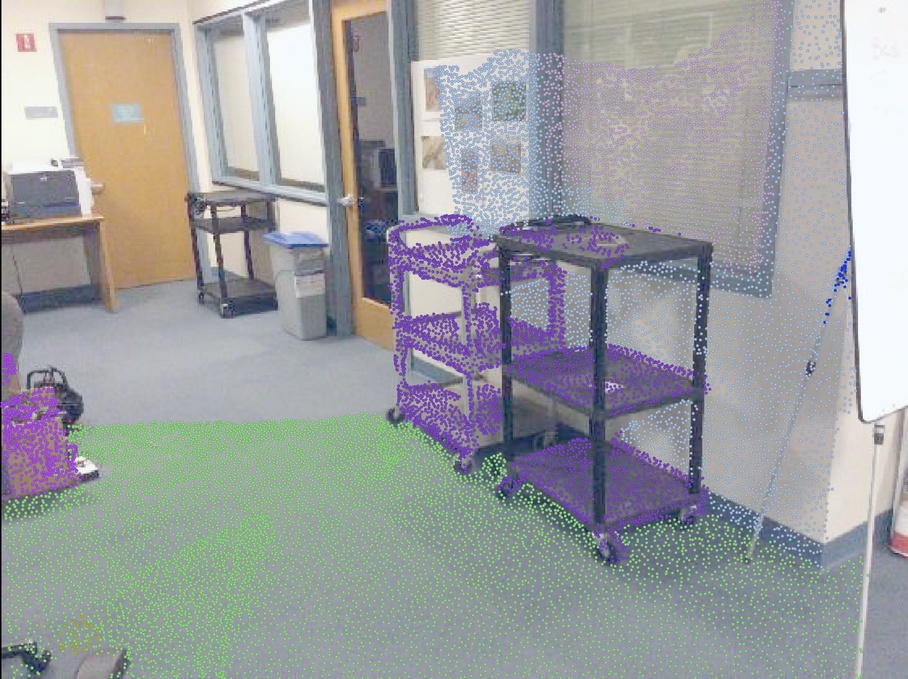} &
        \specialfig{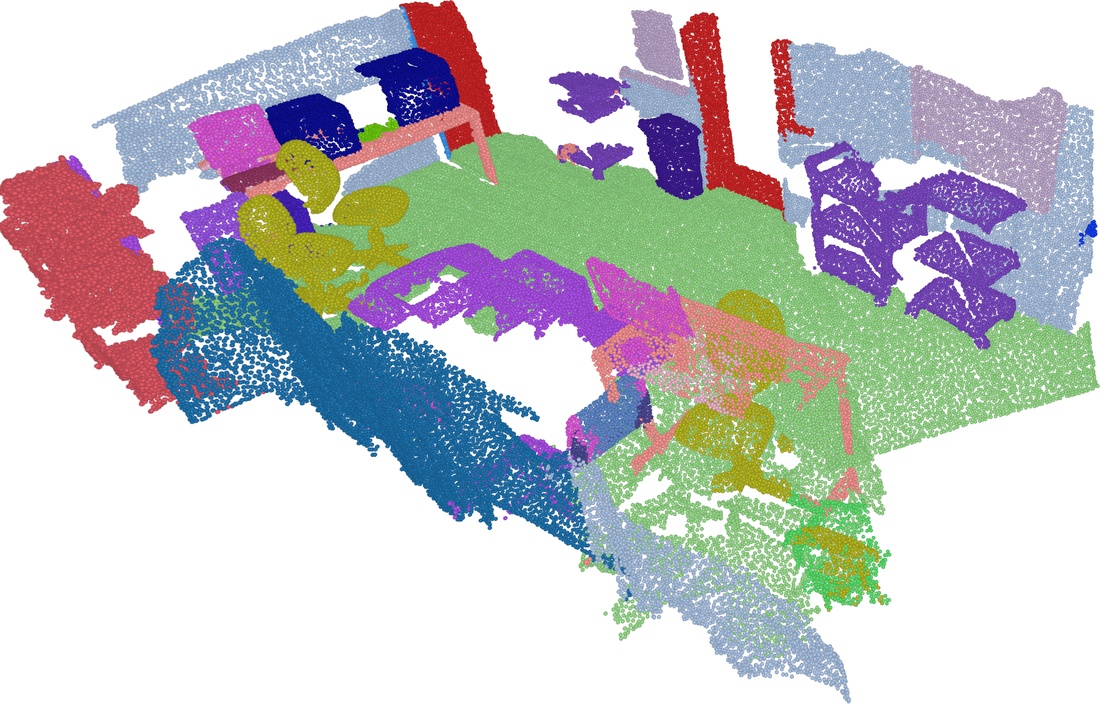}                                                                           \\
        2D projected DITR predictions                                                              & Full 3D DITR predictions                     \\[10pt]
        \includegraphics[width=0.33\textwidth,trim={5px 5px 5px 5px},clip]{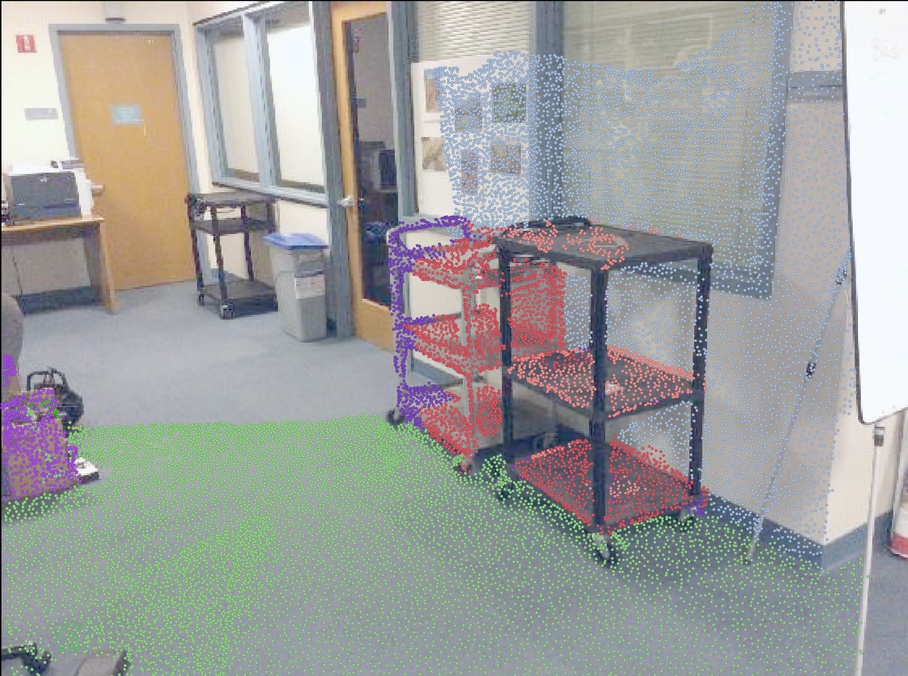} &
        \specialfig{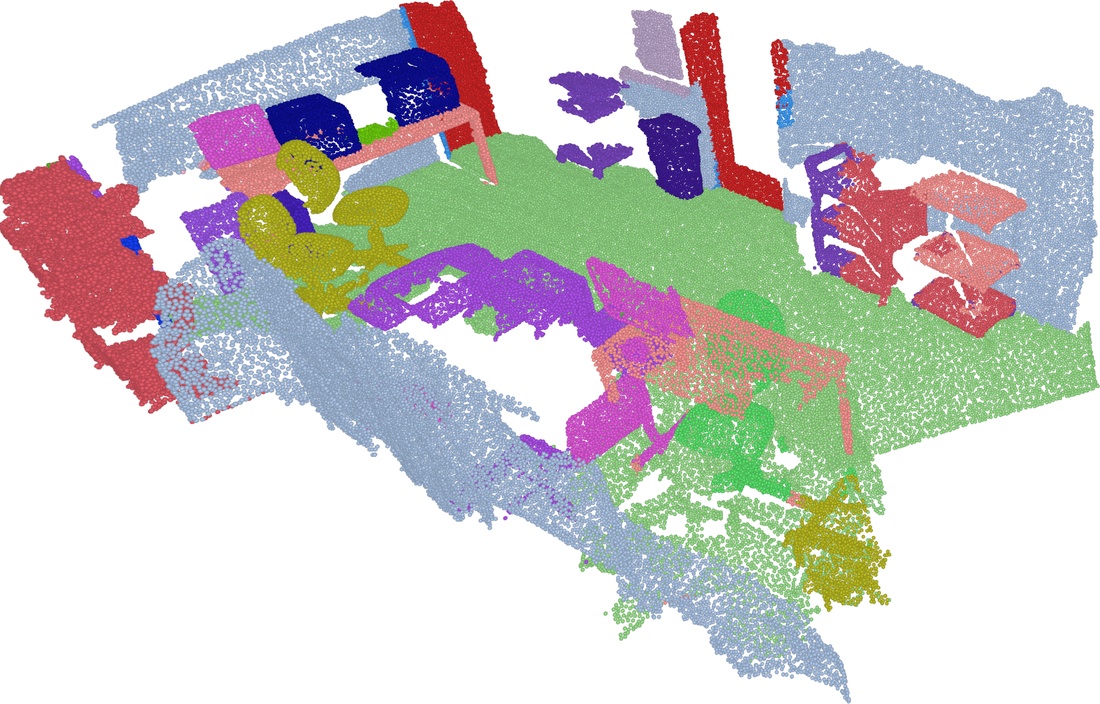}                                                                           \\
        2D projected PTv3 predictions                                                              & Full 3D PTv3 predictions                     \\[10pt]
        \includegraphics[width=0.4\textwidth]{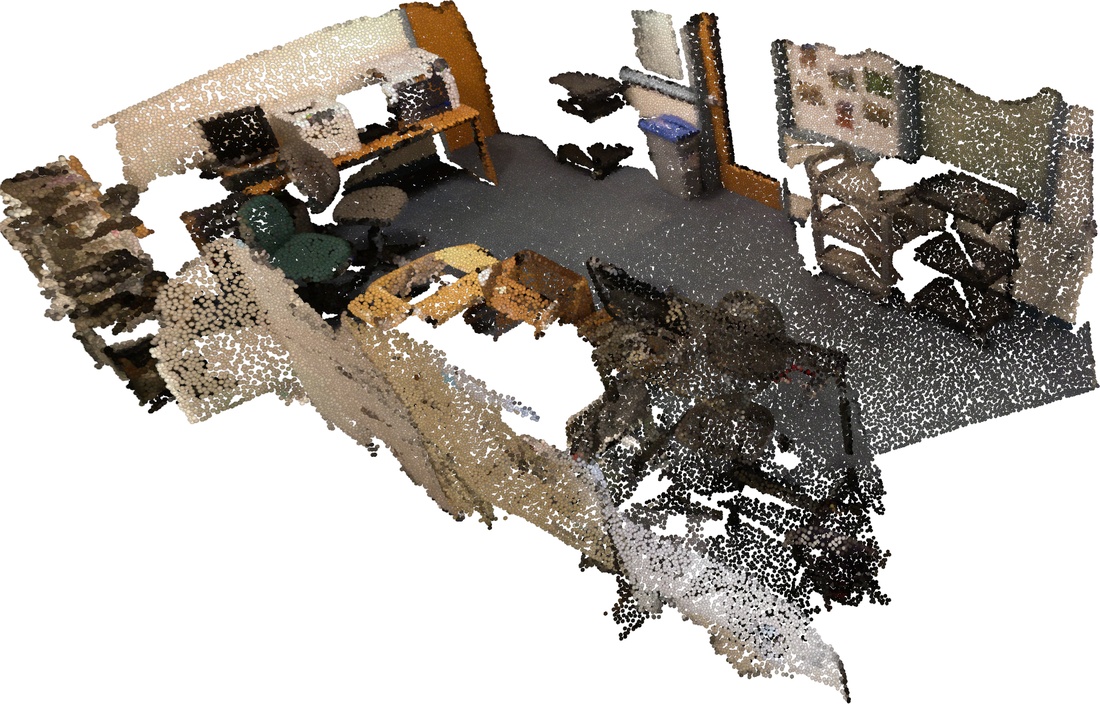} &
        \includegraphics[width=0.4\textwidth]{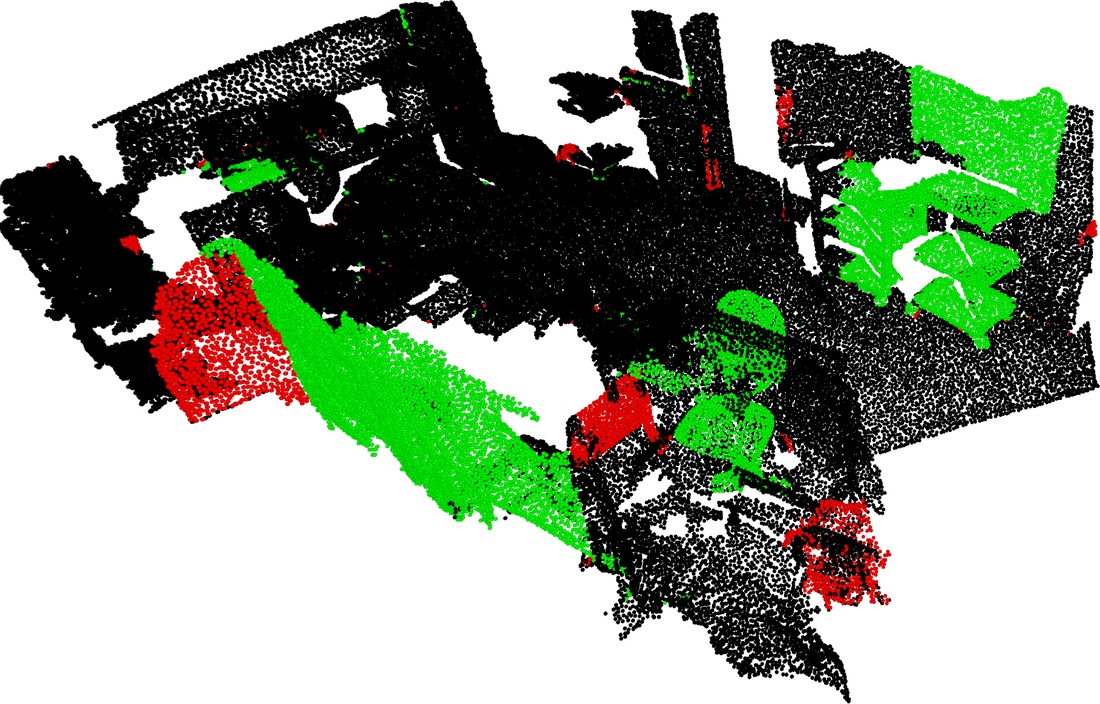}                                                 \\
        3D point cloud input                                                                       & Difference between PTv3 and DITR predictions \\
    \end{tabular}
    \caption{
        \textbf{DITR and PTv3~\cite{wu2024ptv3} segmentation for ScanNet~\cite{dai2017scannet}.}
        The wheels of the cart are barely visible in the 3D point cloud, likely causing PTv3 to misclassify it as a shelf. By contrast, \ours{} can leverage semantically rich image features to correctly identify this cart. The bottom-right visualization illustrates the differences between PTv3 and DITR predictions. Points where both methods are correct or both are incorrect are shown in black. Points correctly predicted by only one method are color-coded: green for DITR and red for PTv3.
    }
    \label{fig:supp:seg_scannet_1}
\end{figure*}

\begin{figure*}[ht]
    \newcommand{\specialfig}[1]{\begin{tikzpicture}
            \node[anchor=south west, inner sep=0] (image) at (0,0) {\includegraphics[width=0.41\textwidth]{#1}};
            \begin{scope}[x={(image.south east)}, y={(image.north west)}]
                \coordinate (A) at (0.12, 0.85);
                \coordinate (B) at (0.17, 0.33);
                \coordinate (C) at (0.5, 0.5);
                \coordinate (D) at (0.5, 0.95);
                \draw[very thick, red] plot [smooth cycle, tension=0.05] coordinates {(A) (B) (C) (D)};

            \end{scope}
        \end{tikzpicture}}
    \centering
    \begin{tabular}{cc}
        \includegraphics[width=0.35\textwidth,trim={5px 5px 5px 5px},clip]{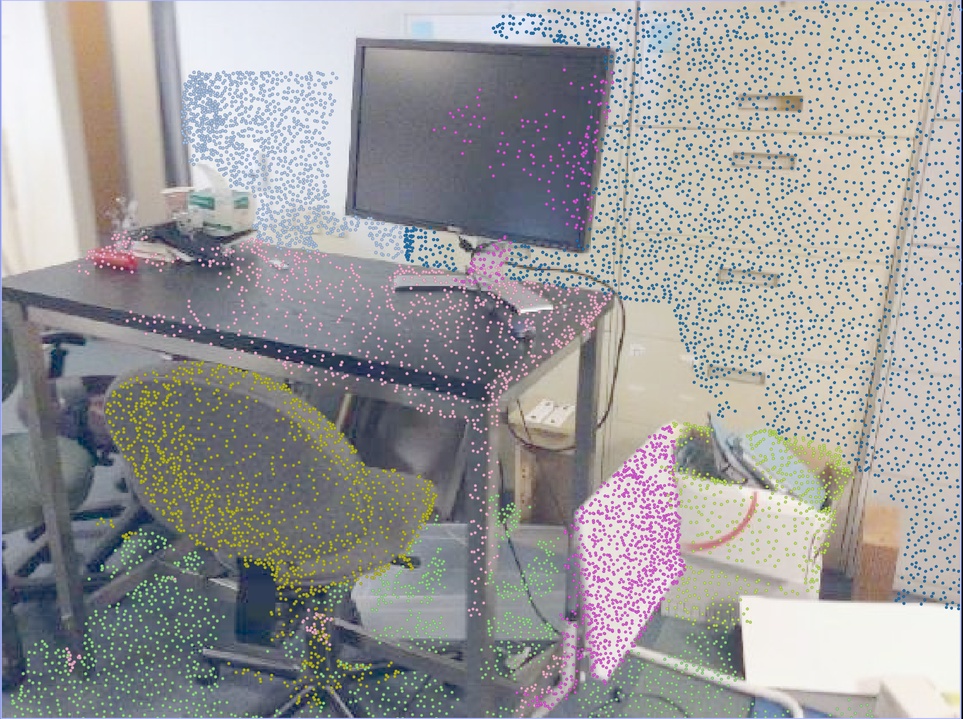}    &
\specialfig{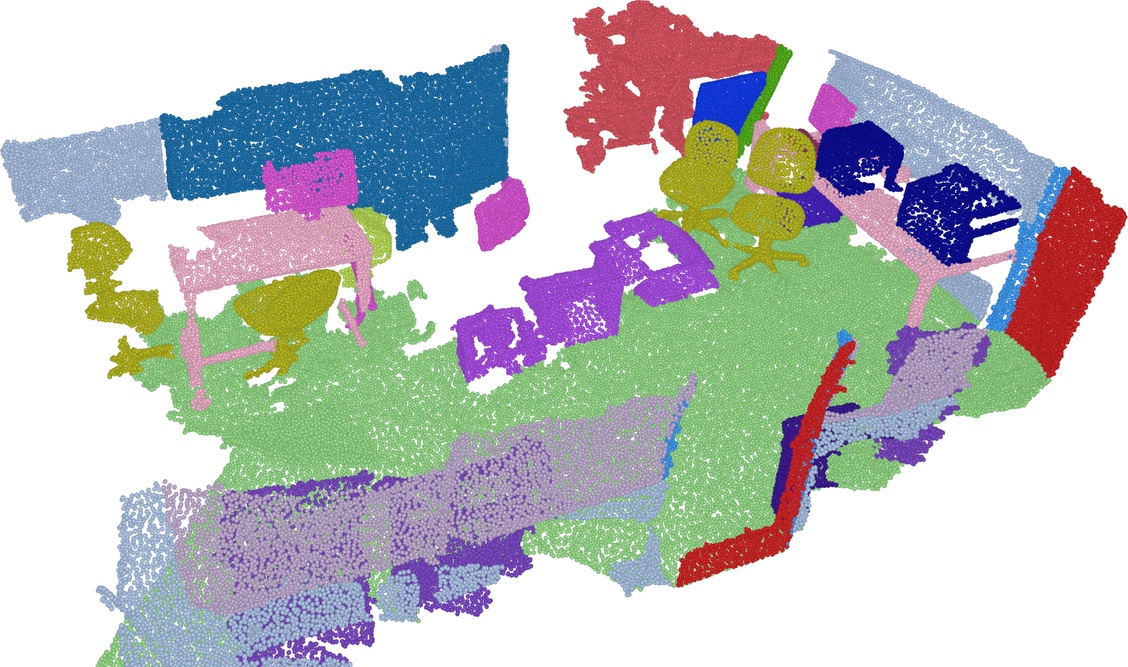}                                                                              \\
        2D projected ground truth                                                                   & Full 3D ground truth                         \\[10pt]
        \includegraphics[width=0.35\textwidth,trim={5px 5px 5px 5px},clip]{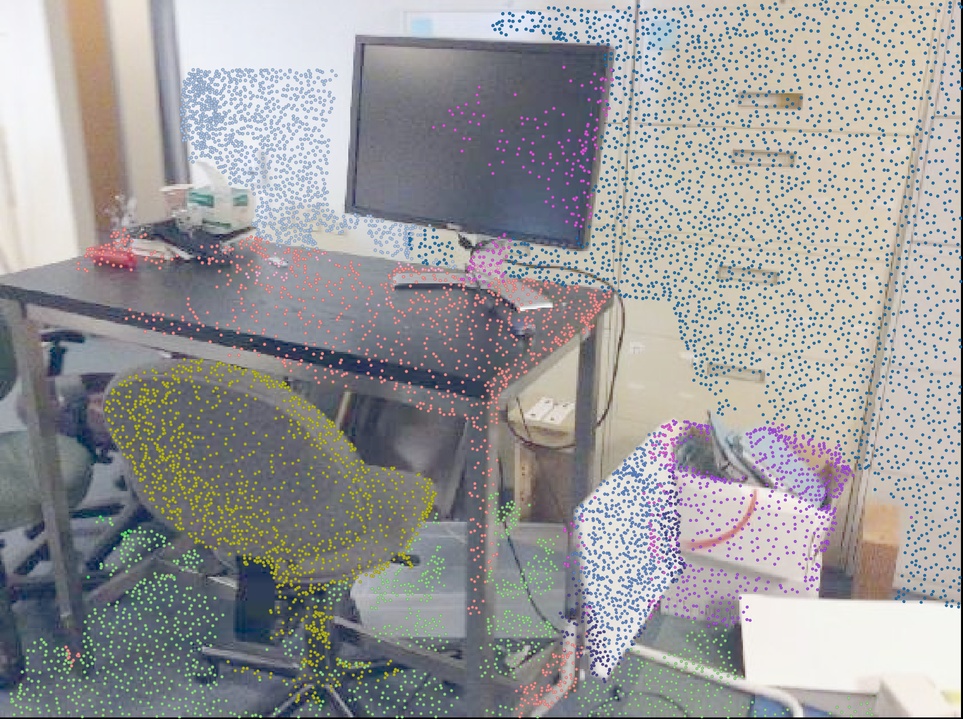}  &
        \specialfig{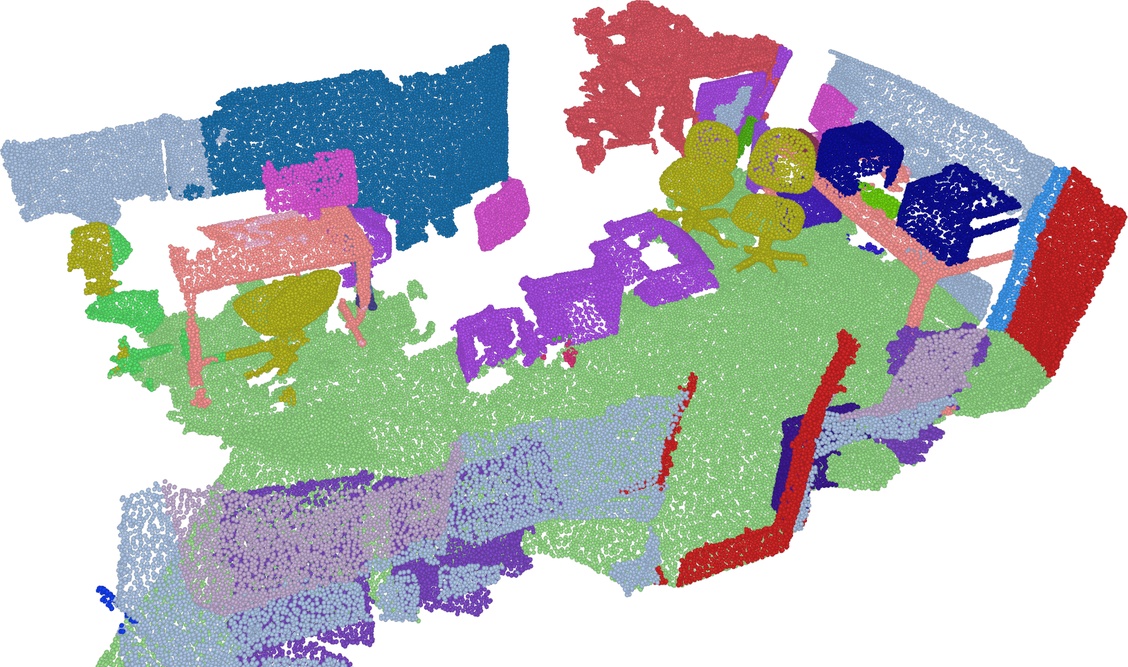}                                                                            \\
        2D projected DITR predictions                                                               & Full 3D DITR predictions                     \\[10pt]
        \includegraphics[width=0.35\textwidth,trim={5px 5px 5px 5px},clip]{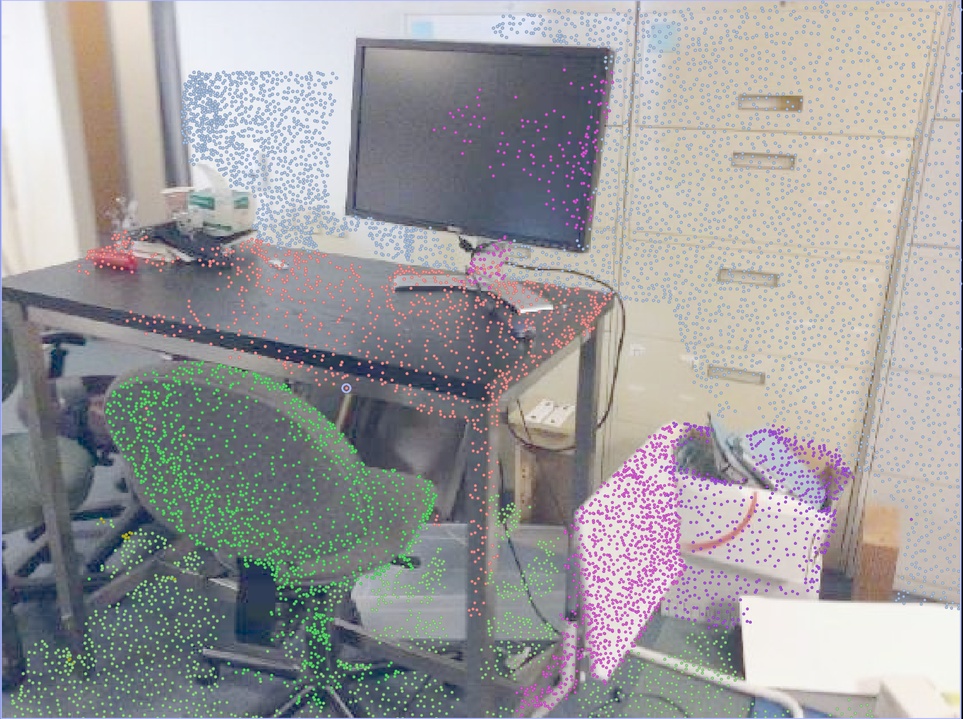}  &
        \specialfig{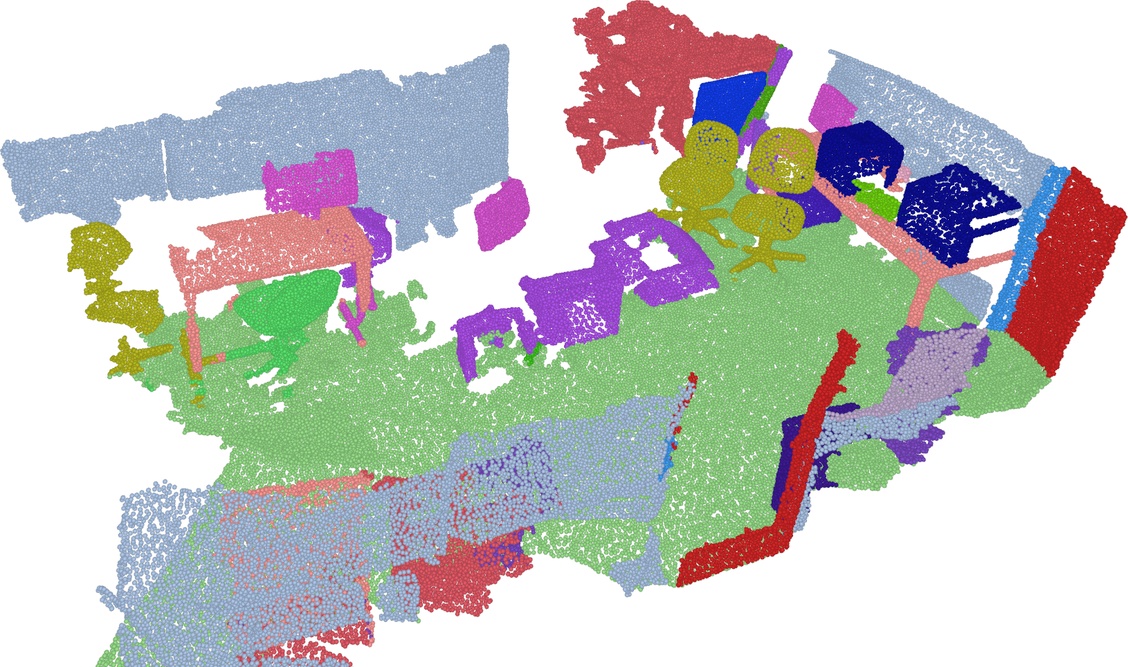}                                                                            \\
        2D projected PTv3 predictions                                                               & Full 3D PTv3 predictions                     \\[10pt]
        \includegraphics[width=0.42\textwidth]{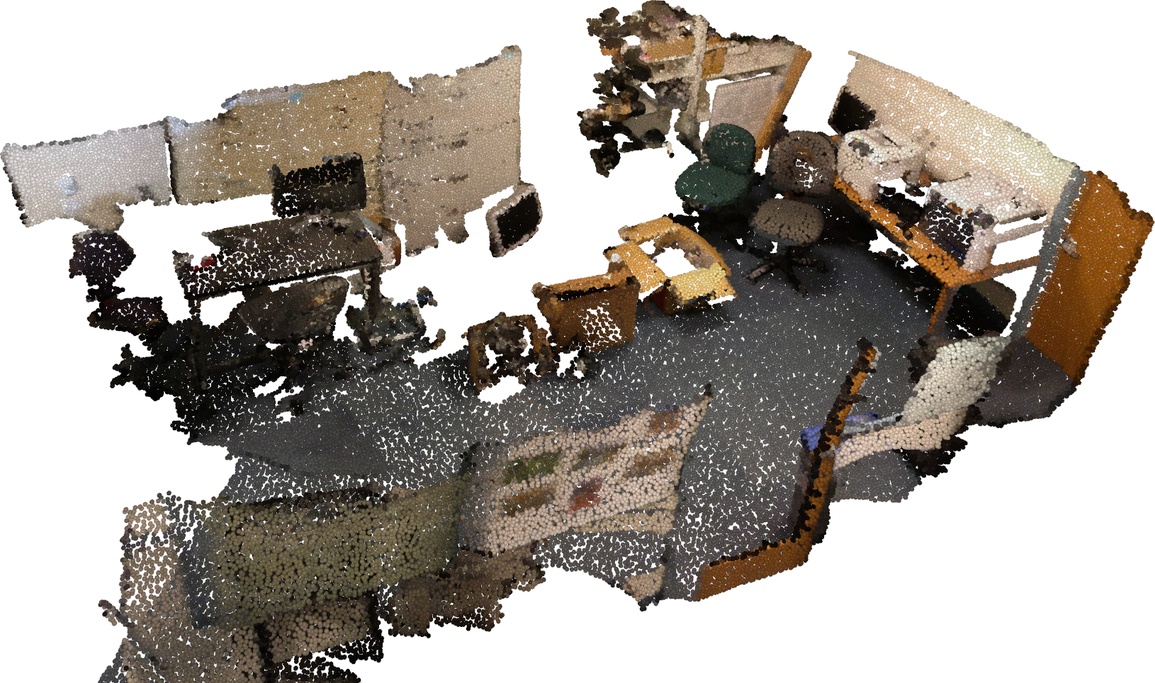} &
        \includegraphics[width=0.42\textwidth]{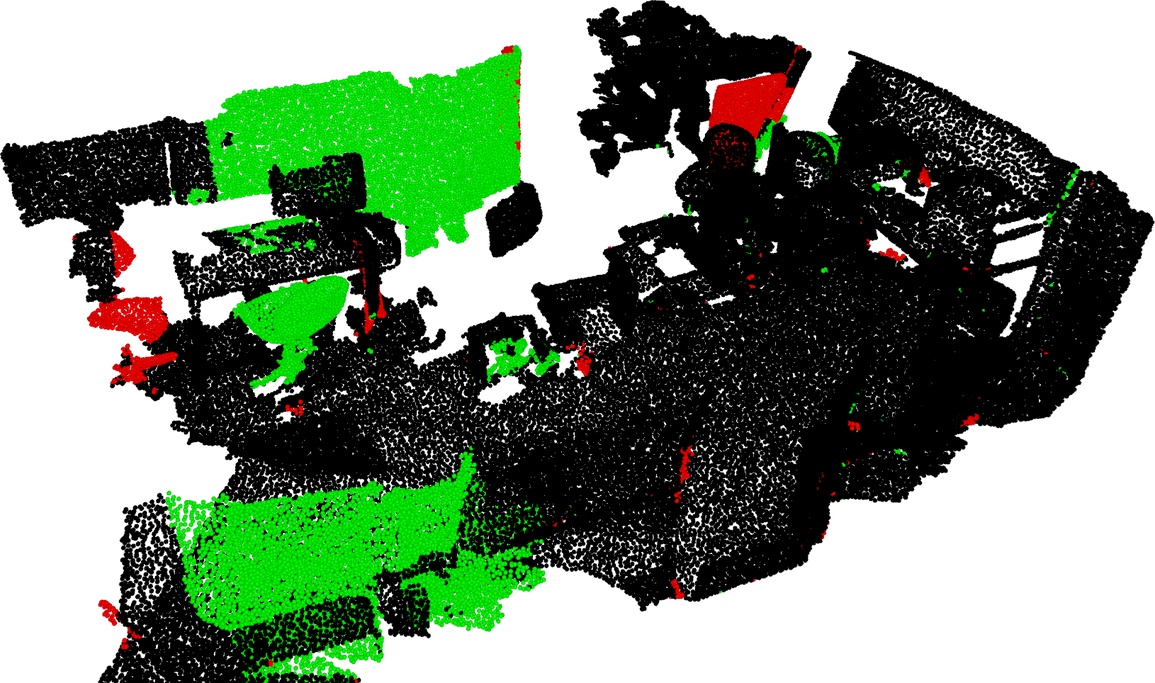}                                                 \\
        3D point cloud input                                                                        & Difference between PTv3 and DITR predictions \\
    \end{tabular}
    \caption{
        \textbf{DITR and PTv3~\cite{wu2024ptv3} semantic segmentation for ScanNet~\cite{dai2017scannet}.}
        In the 3D point cloud, the cabinet at the back appears as a flat surface flush with the wall, leading PTv3 to misclassify it as part of the wall. However, in the corresponding image, this region is clearly identifiable as a cabinet.
        The bottom-right visualization illustrates the differences between PTv3 and DITR predictions. Points where both methods are correct or both are incorrect are shown in black. Points correctly predicted by only one method are color-coded: green for DITR and red for PTv3.
    }
    \label{fig:supp:seg_scannet_2}
\end{figure*}

%% file: fig/supplementary/seg_nuscenes.tex
\begin{figure*}[ht]
    \centering
    \begin{tabular}{cc}
        \includegraphics[width=0.34\textwidth,trim={5px 5px 5px 5px},clip]{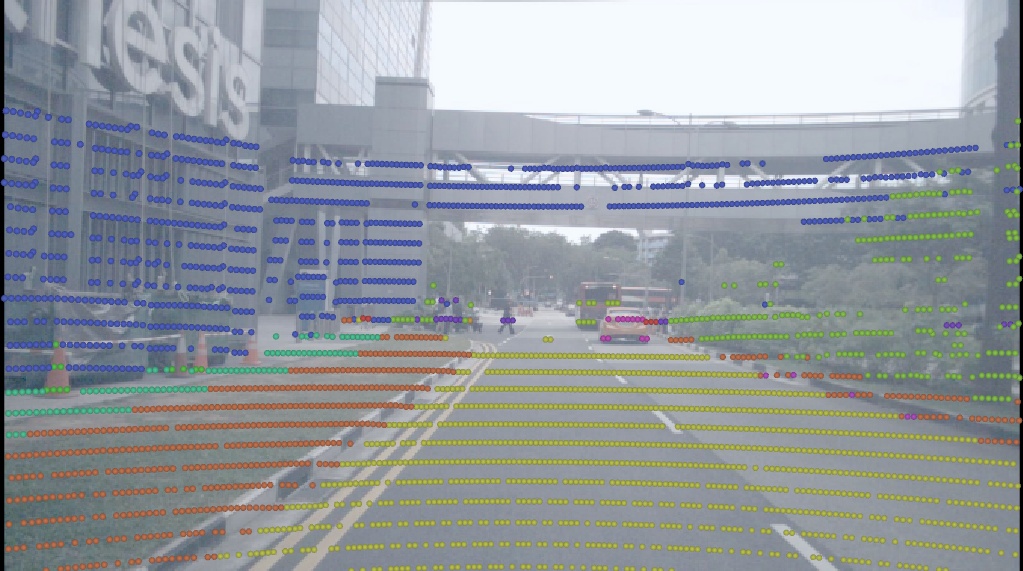} & 
        \includegraphics[width=0.41\textwidth,trim={12cm 14cm 12cm 6cm},clip]{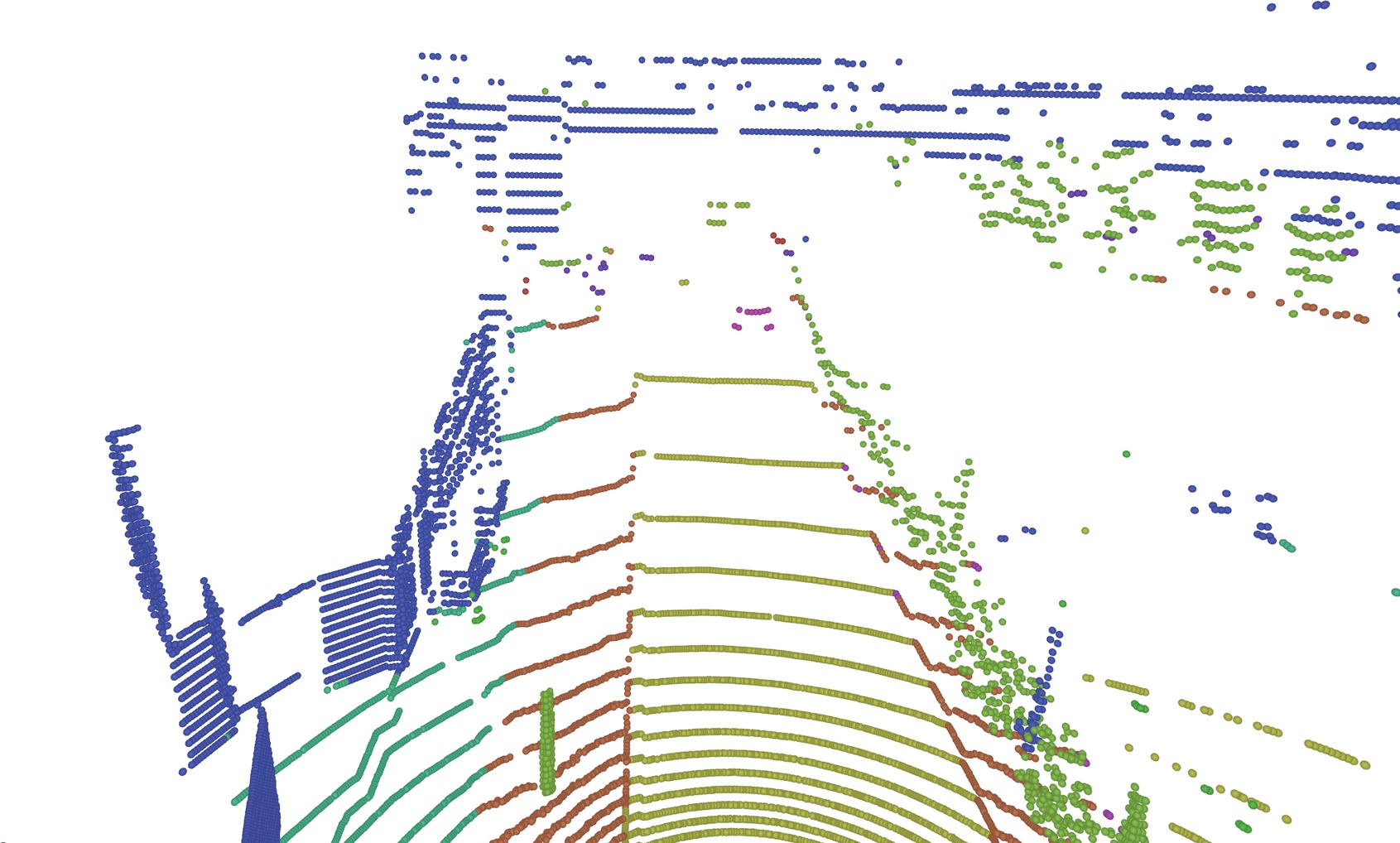} \\
        2D projected ground truth & Front view 3D ground truth \\[10pt]
        \includegraphics[width=0.34\textwidth,trim={5px 5px 5px 5px},clip]{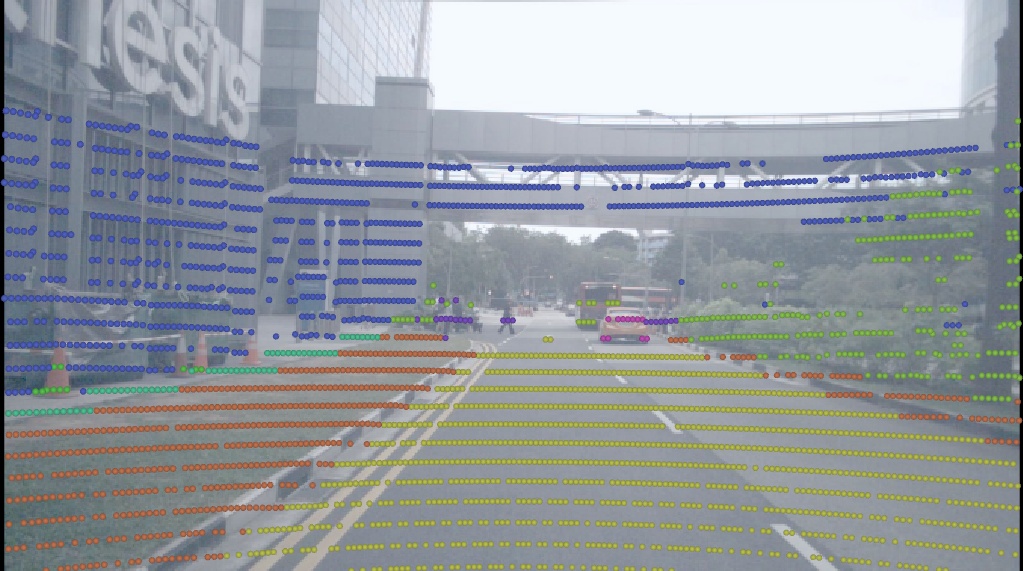} & 
        \includegraphics[width=0.41\textwidth,trim={12cm 14cm 12cm 6cm},clip]{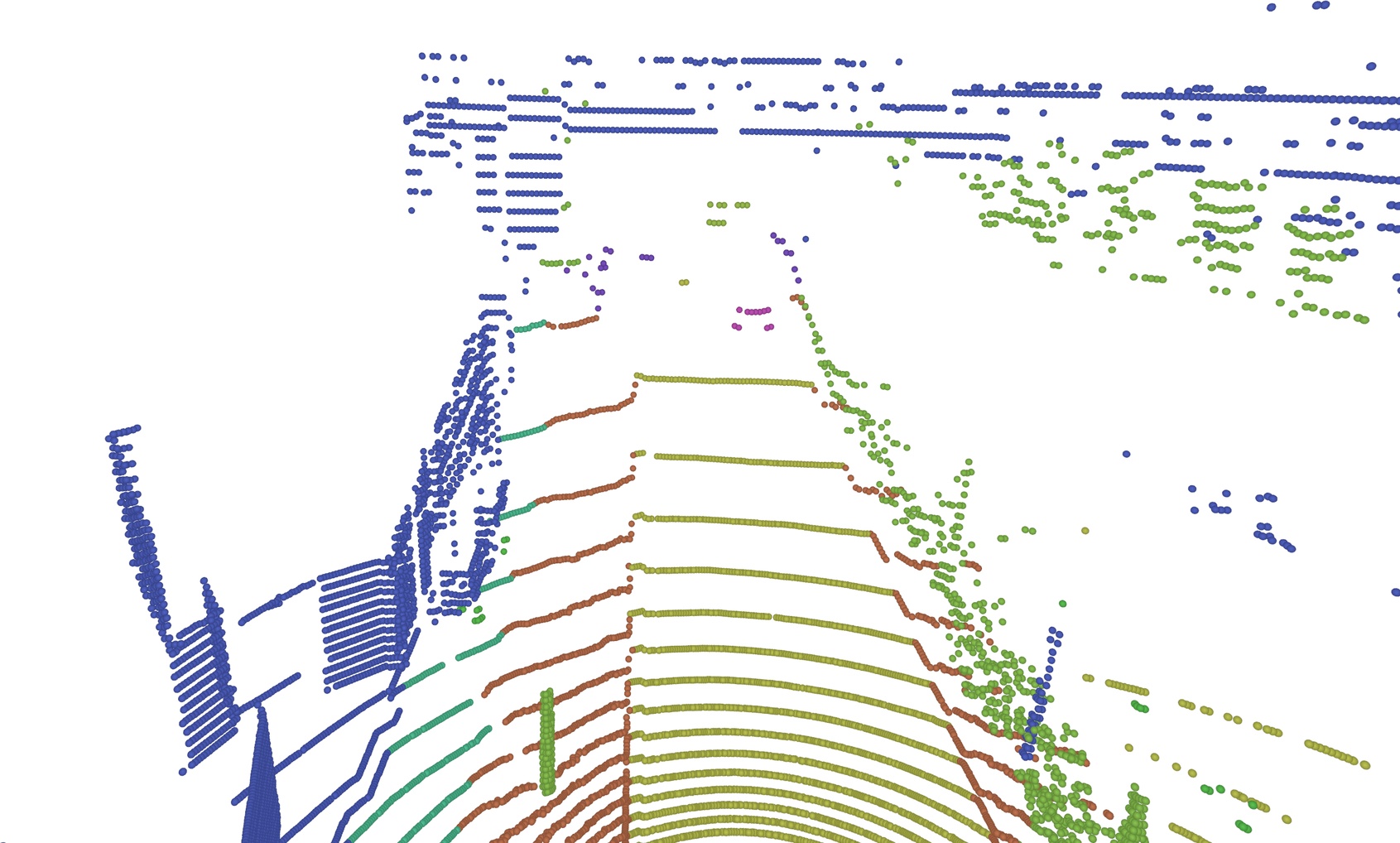} \\
        2D projected DITR predictions & Front view 3D DITR predictions \\[10pt]
        \includegraphics[width=0.34\textwidth,trim={5px 5px 5px 5px},clip]{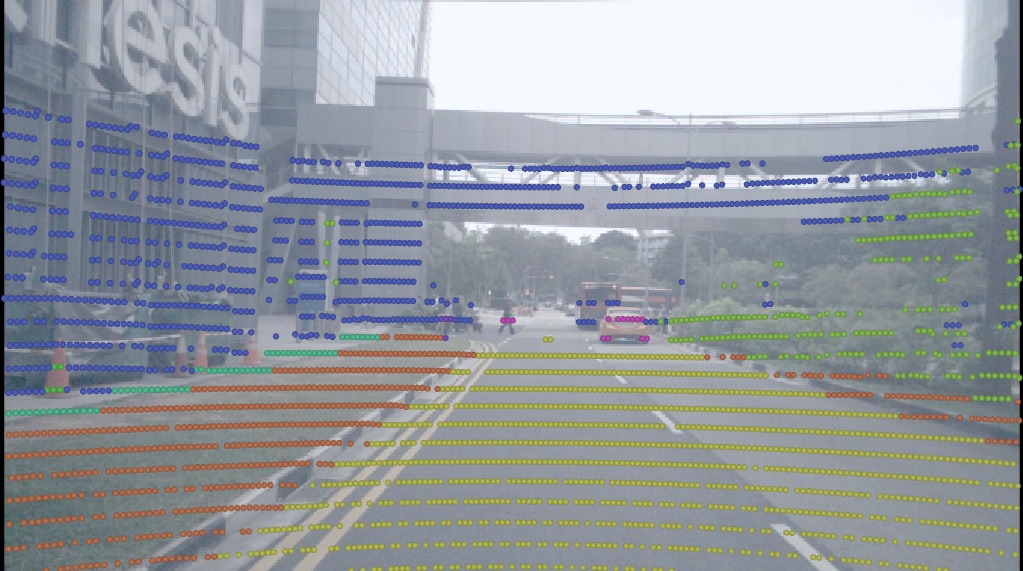} & 
        \includegraphics[width=0.41\textwidth,trim={12cm 14cm 12cm 6cm},clip]{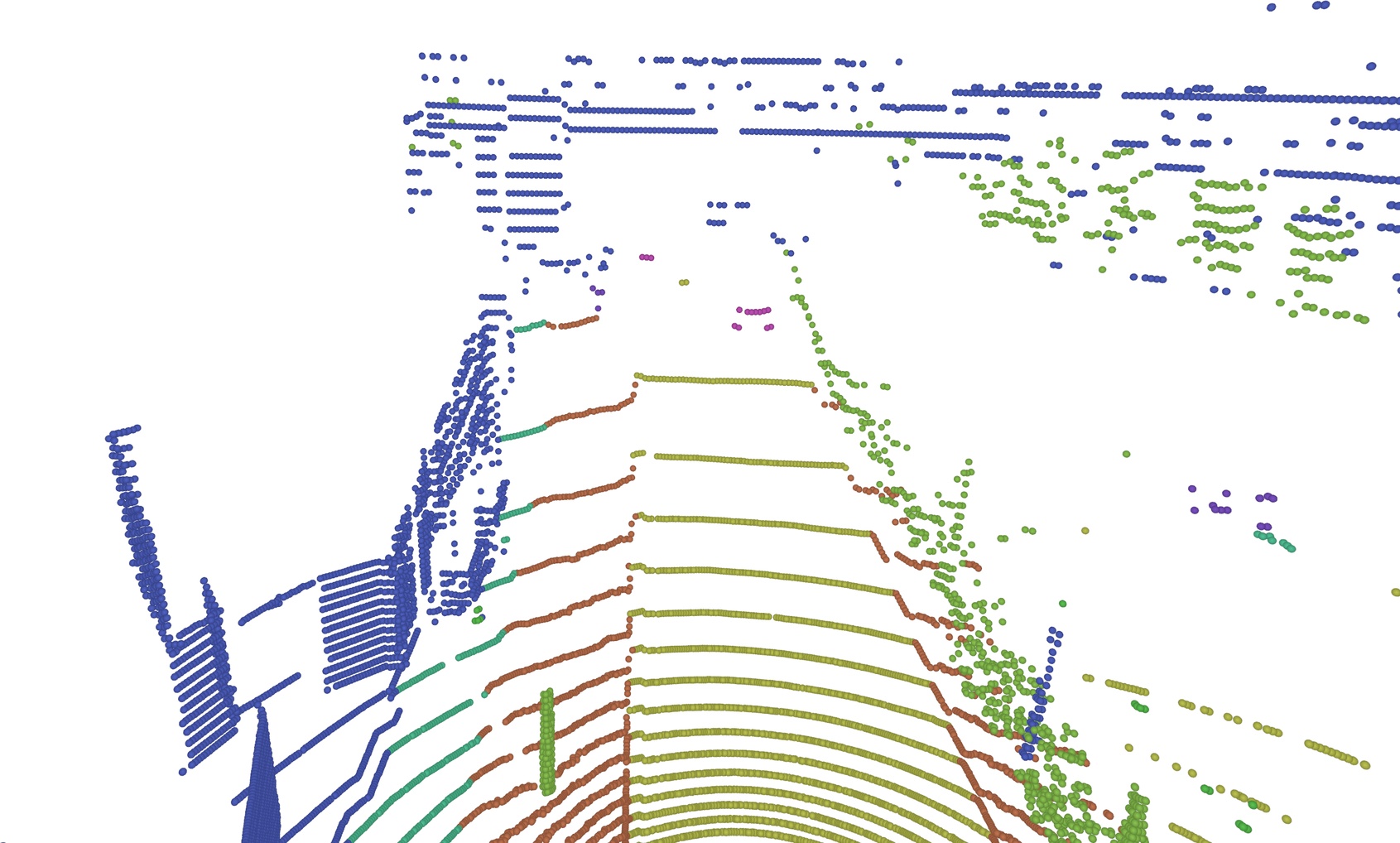} \\
        2D projected PTv3 predictions & Front view 3D PTv3 predictions \\[10pt]
        & 
        \includegraphics[width=0.41\textwidth,trim={12cm 14cm 12cm 6cm},clip]{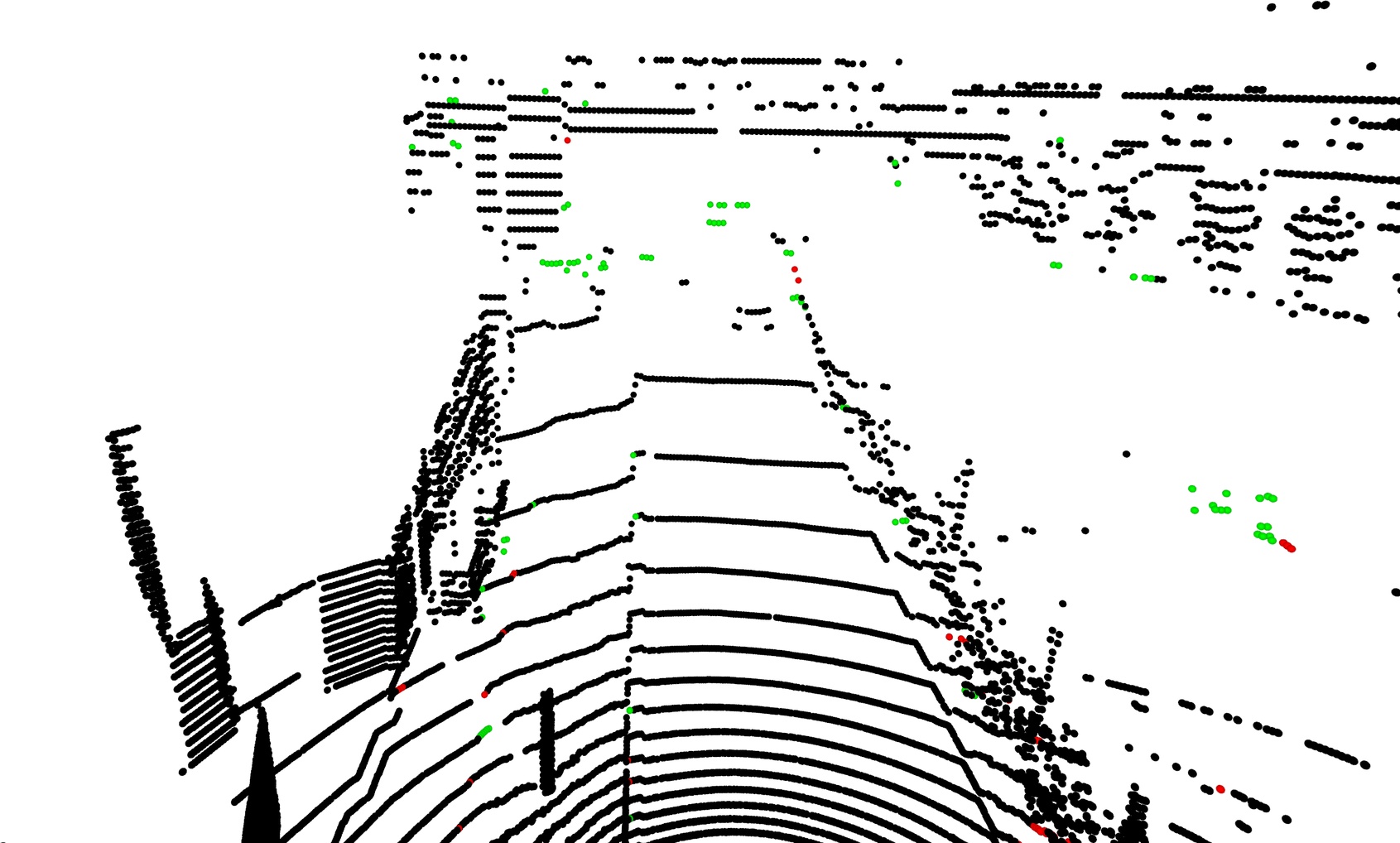} \\
        & Difference between PTv3 and DITR predictions \\
    \end{tabular}
    \caption{
        \textbf{DITR and PTv3~\cite{wu2024ptv3} semantic segmentation for nuScenes~\cite{caesar2020nuscenes}.}
        PTv3 misclassifies the pedestrian crossing the street as a car, overlooks the pedestrians on the left sidewalk and the pedestrian on the right, and mistakes the bus in front of the car for a building.
        These errors occur in areas distant from the ego vehicle, where LiDAR data is sparse, and objects are represented by only a few points.
        In contrast, DITR can leverage the corresponding image, where the bus and pedestrian on the street are clearly visible, allowing it to make the correct predictions despite the objects being represented by only a few points.
        The bottom-right visualization illustrates the differences between PTv3 and DITR predictions. Points where both methods are correct or both are incorrect are shown in black. Points correctly predicted by only one method are color-coded: green for DITR and red for PTv3.
    }
    \label{fig:supp:seg_nuscenes}
\end{figure*}